\documentclass[runningheads]{llncs}

\usepackage{eccv}

\usepackage{eccvabbrv}

\usepackage{graphicx}
\usepackage{booktabs}

\usepackage[toc,page]{appendix}
\usepackage{titletoc}

\usepackage[accsupp]{axessibility}  

\usepackage{hyperref}

\usepackage{orcidlink}

\usepackage{color}
\usepackage{colortbl}
\usepackage{xcolor}
\usepackage{tikz}
\definecolor{gold}{HTML}{BD820B}
\definecolor{silver}{HTML}{909090}
\definecolor{bronze}{HTML}{9A5F26}
\definecolor{lgray}{gray}{0.95}
\definecolor{gaincolor}{RGB}{230, 245, 255}
\definecolor{Gray}{gray}{0.91}
\definecolor{LightCyan}{rgb}{0.82,0.82,1}
\newcolumntype{a}{>{\columncolor{Gray}}c}
\newcolumntype{B}{>{\columncolor{LightCyan}}c}
\usepackage{multirow}
\usepackage{arydshln}
\usepackage{algorithmicx}
\usepackage{algpseudocode}
\usepackage{stackengine}
\usepackage{adjustbox}
\usepackage{hhline}
\usepackage{threeparttable}
\usepackage{soul}
\usepackage{fix-cm}
\usepackage[linesnumbered,ruled,vlined]{algorithm2e}
\usepackage{siunitx}
\usepackage{subcaption}
\usepackage{tabularx}

\begin{document}

\title{SENTRY \includegraphics[height=1.5em]{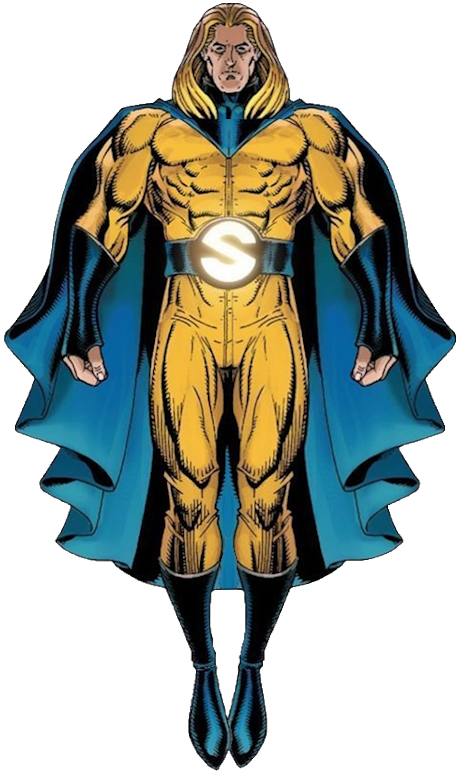}\hspace{0.1em}: SAM2-Enhanced Neighbor-Aware and Temporally Reasoned Memory for Visual Tracking} 

\titlerunning{SENTRY}


\author{
Mohamad Alansari$^{1}$\thanks{Equal contribution.}\orcidlink{0000-0003-2960-2972} 
\and
Yonathan Michael$^{1*}$\orcidlink{0009-0005-8457-294X} 
\and
Hasan AlMarzouqi$^{1}$\orcidlink{0000-0002-2826-1515}
\and
Muzammal Naseer$^{1,2}$\orcidlink{0000-0001-7663-7161}
\and
Naoufel Werghi$^{1}$\orcidlink{0000-0002-5542-448X}
\and
Sajid Javed$^{1}$\orcidlink{0000-0002-0036-2875}
}

\authorrunning{M.~Alansari, Y. Michael et al.}


\institute{
$^1$Khalifa University, Abu Dhabi, UAE \quad
$^2$University of Western Australia, Australia\\
\email{\{100061914,100053679,hasan.almarzouqi,muhammadmuzammal.naseer,\\naoufel.werghi,sajid.javed\}@ku.ac.ae} \\
\href{https://hamadya.github.io/SENTRY/page/}{https://hamadya.github.io/SENTRY/}
}


\maketitle

\begin{abstract}
\noindent We revisit the memory update mechanism in SAM2-based visual object tracking and identify confidence-only mask selection as the dominant cause of drift under occlusion, rapid motion, and distractors.
We introduce SENTRY, a training-free, plug-and-play, refine-before-write module that validates each memory update for short-horizon temporal consistency before committing it.
SENTRY aggregates diverse segmentation hypotheses per frame, backtracks them into short tracklets, and uses neighbor-aware cycle-consistent matching against recent trajectories to favor temporally and geometrically consistent masks.
It leaves the base architecture untouched, replacing confidence-driven writes with consistency-validated ones.
For fair evaluation, we re-evaluate major open-source SAM2-based trackers across all available scales and datasets, filling gaps in prior reports.
Integrated into five strong baselines, SENTRY delivers consistent gains across nine benchmarks, achieving new zero-shot SOTA on LaSOT, LaSOT\textsubscript{ext}, GOT-10k, VOT20, VOT22, and DiDi.
Despite these checks, the SAM2-L version runs at 32.8 FPS on an A100, and across compatible hosts adds only about 0.4--0.6 GB VRAM.
Our results provide the first unified all-scale evaluation of SAM2-based trackers and show that enforcing temporal validity at write time stabilizes memory-augmented tracking without retraining.
\vspace{-1em}
{  \keywords{Segment Anything \and Temporal Verification \and Visual Object Tracking} }
\end{abstract}
\vspace{-1em}

\begin{figure}[t!]
\centering
\begin{subfigure}[b]{0.49\textwidth}
\includegraphics[width=\linewidth]{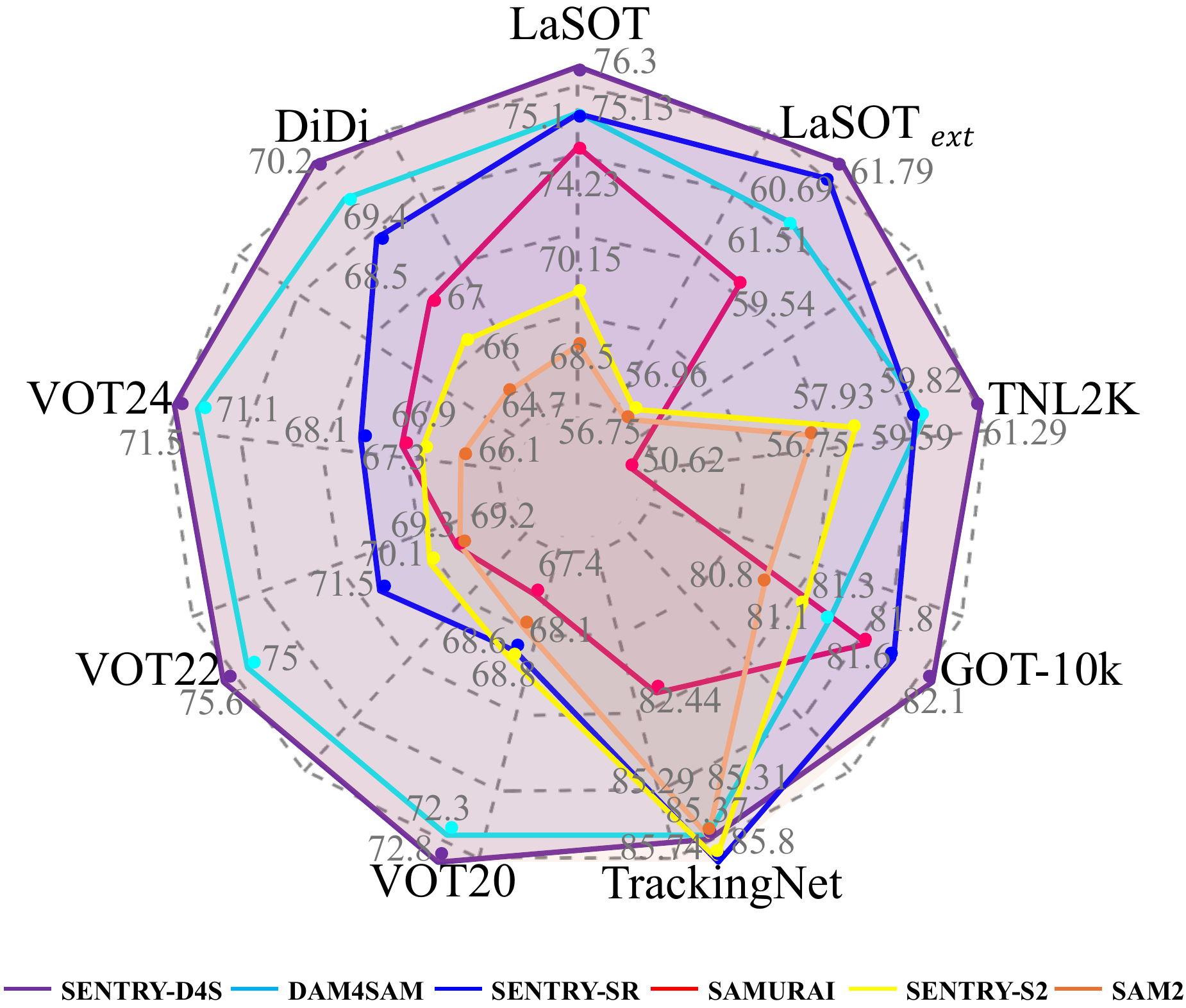}
\caption{}
\end{subfigure}
\begin{subfigure}[b]{0.49\textwidth}
\includegraphics[width=\linewidth]{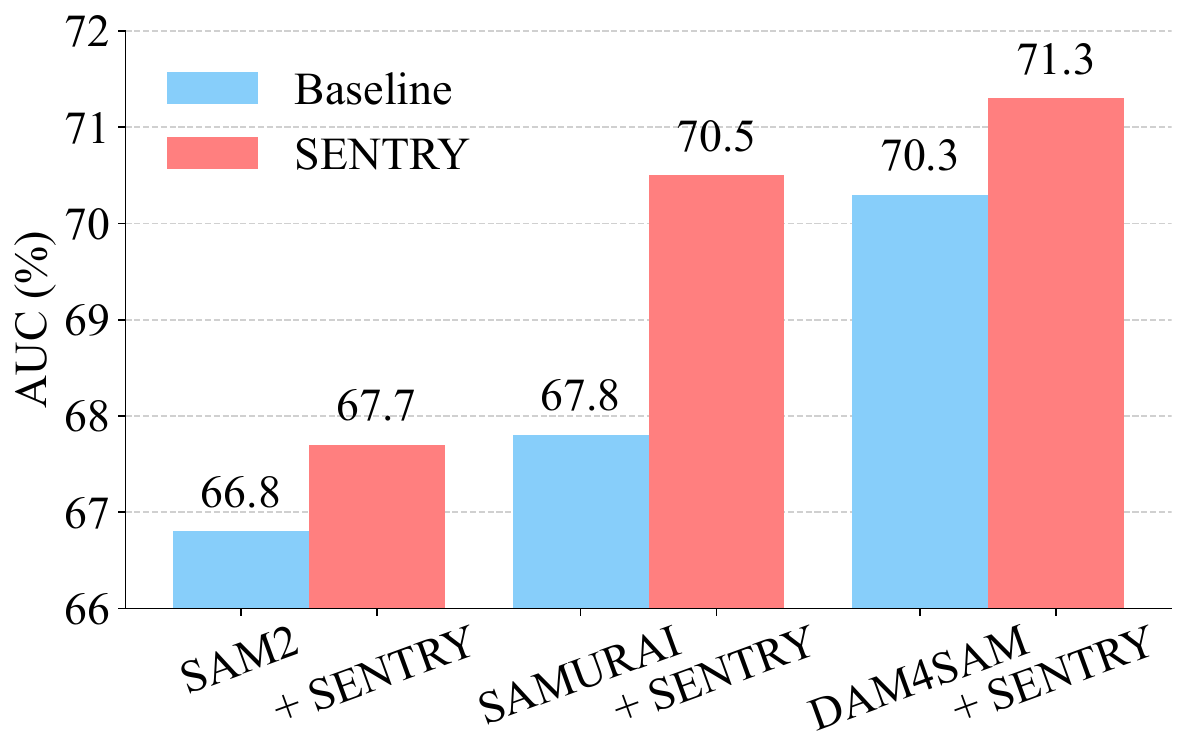}
\caption{}
\end{subfigure}
\vspace{-1em}
\caption{
SENTRY performs favorably against SAM2-based variants across major tracking benchmarks.
We denote SENTRY applied to SAM2, SAMURAI, and DAM4SAM as SENTRY-S2, SENTRY-SR, and SENTRY-D4S.
(a) AUC per benchmark; (b) average AUC across \cite{lasot,lasotext,tnl2k,trackingnet}.
}
\label{fig:1}
\vspace{-2em}
\end{figure}

\vspace{-2em}
\section{Introduction} \label{sec:intro}
\noindent Memory-based architectures have become central to visual object tracking (VOT) by maintaining feature histories that help recover from occlusion, motion blur, and appearance changes. Instead of processing frames independently \cite{stark,ostrack,cldtracker,dutrack,artrack,spmtrack,untrack,mambavlt,sglatrack}, recent trackers build temporal associations by querying a memory bank of past representations, improving robustness and identity preservation \cite{aot,xmem,rmem,vot2020,vot2022,vots2023,vots2024}.

\begin{figure*}[t!]
\vspace{-1em}
  \centering \small
  \begin{subfigure}[b]{0.325\textwidth}
    \includegraphics[width=\textwidth]{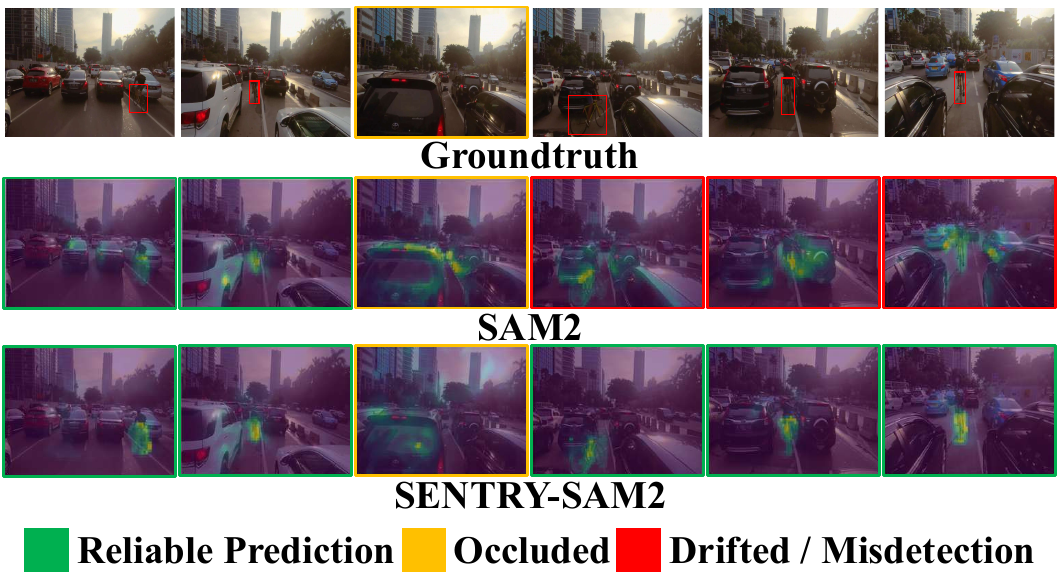}
    \caption{}
  \end{subfigure}
  \begin{subfigure}[b]{0.325\textwidth}
    \includegraphics[width=\textwidth]{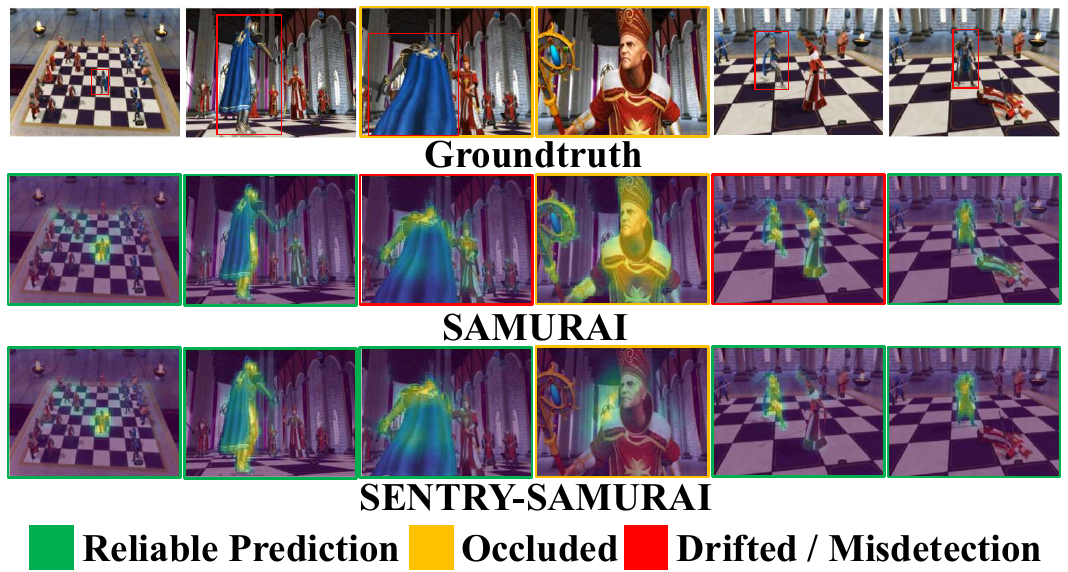}
    \caption{}
  \end{subfigure}
  \begin{subfigure}[b]{0.325\textwidth}
    \includegraphics[width=\textwidth]{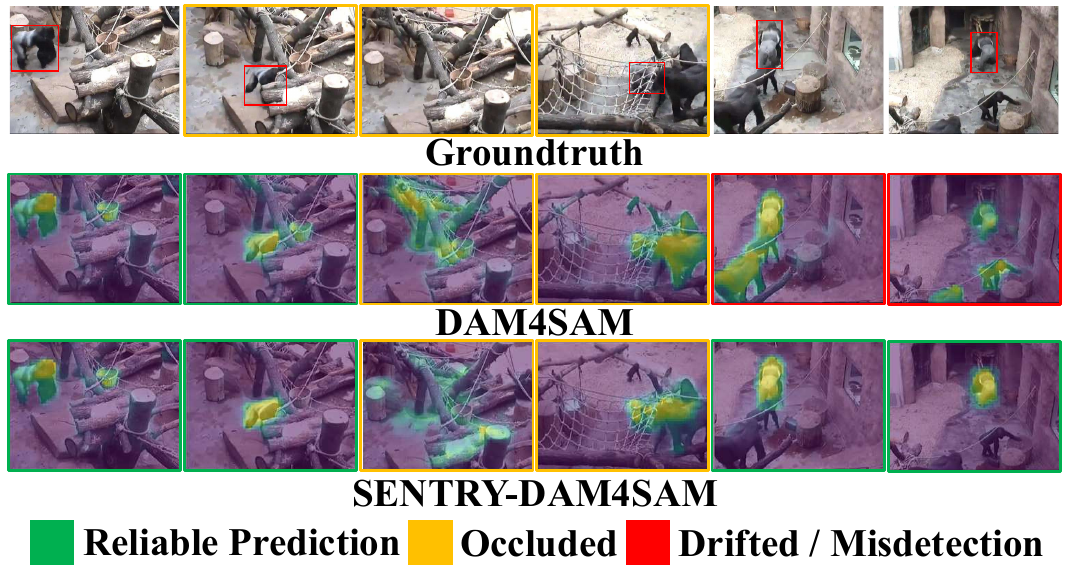}
    \caption{}
  \end{subfigure}
  \vspace{-1em}
\caption{
Qualitative comparison under occlusion, abrupt motion, and distractor interference. 
We visualize decoder attention to highlight spatial focus during tracking.
(a) SAM2 \cite{sam2} loses target identity after occlusion, (b) SAMURAI \cite{samurai} drifts under rapid motion, and (c) DAM4SAM \cite{dam4sam} misidentifies distractors due to heuristic filtering.
The last row shows SENTRY maintaining accurate localization and consistent masks.
}
\label{fig:2}
\vspace{-2em}
\end{figure*}

SAM2 \cite{sam2} follows this trend with streaming memory and multiple mask hypotheses per frame. Its main weakness is confidence-driven mask selection: the highest-confidence mask is chosen without verifying short-term temporal consistency. Under occlusion, fast motion, or distractors, this cue is often unreliable, causing incorrect masks to be written into memory and triggering drift (Fig. \ref{fig:2}(a)).
Existing extensions attempt to mitigate this with motion filtering \cite{samurai} or distractor checks \cite{dam4sam}, but they still rely heavily on SAM2’s confidence-based mask selection. Motion filters can break under nonlinear trajectories, and heuristic thresholds often fail in cluttered scenes, leading to similar error accumulation (Figs. \ref{fig:2}b,c). To our knowledge, \emph{current SAM2-based trackers lack a systematic, trajectory-level mechanism to verify short-term temporal consistency before a memory write}.

We address this gap with \textbf{S}AM2-\textbf{E}nhanced \textbf{N}eighbor-aware and \textbf{T}emporally \textbf{R}easoned memor\textbf{Y} (SENTRY), a training-free inference-time memory-admission module that reduces incorrect memory writes.
SENTRY targets a different interface from prior motion- or distractor-aware extensions: the memory-write decision. Rather than only improving current-frame prediction or score, it verifies which candidate mask should become future evidence in the autoregressive memory stream.
SENTRY replaces confidence-driven selection with a refine-before-write mechanism that explicitly tests each hypothesis for short-horizon temporal plausibility. 
For every candidate, SENTRY (i) propagates it backward over a fixed window, (ii) applies neighbor-aware checks that penalize identity swaps with nearby instances or distractors, and (iii) evaluates cycle-consistency by requiring that forward–backward trajectories remain aligned with the target’s recent motion. 
This trajectory-level verification, absent in existing SAM2-style architectures, adds a lightweight temporal reasoning layer without modifying the backbone or memory.
Because SENTRY operates entirely outside the backbone and memory module, it integrates cleanly with existing SAM2-based variants and substantially improves robustness when confidence cues are unreliable (see the bottom row in Fig. \ref{fig:2}).
Our contributions are:
\vspace{-0.5em}
\begin{enumerate}
    
    \item We propose SENTRY, a training-free refine-before-write module that enforces short-horizon temporal consistency before any memory update in SAM2-based trackers. It replaces confidence-driven selection with a unified mechanism that evaluates multi-source mask candidates through backward re-tracking and neighbor-aware trajectory matching.

    \item To the best of our knowledge, we conduct the first unified, all-scale re-evaluation of five open-source SAM2-based trackers across nine benchmarks and four model sizes, establishing consistent and reproducible baselines.

\end{enumerate}
\vspace{-0.5em}
Experiments across nine benchmarks including LaSOT \cite{lasot}, LaSOT\textsubscript{ext} \cite{lasotext}, TNL2K \cite{tnl2k}, GOT-10k \cite{got10k}, TrackingNet \cite{trackingnet}, VOT20/22 \cite{vot2020,vot2022}, VOTS24 \cite{vots2024}, and DiDi \cite{dam4sam}, demonstrate that integrating SENTRY into strong baselines yields consistent improvements and new zero-shot SOTA performance (Fig. \ref{fig:1}), while preserving real-time operation (Sec.~\ref{sec:ablation_study}, App.~\ref{supplement:runtime_memory_overhead}).

\vspace{-1em}
\section{Related Work}
\label{sec:related_work}

\subsection{Memory-Based Visual Object Tracking}
\noindent Memory networks are widely used in VOT and video object segmentation (VOS) to maintain temporal coherence under appearance change, occlusion, and motion. Key–value memory designs such as STM \cite{stm} and FEELVOS \cite{feelvos} established long-term feature propagation, later refined through kernelized or decomposed attention \cite{kmn,stcn,deaot} and transformer-based memory structures \cite{aot,xmem,rmem}. These ideas were further explored in segmentation-based tracking frameworks such as STMTrack \cite{stmtrack}, and inspired memory-augmented single- and multi-object trackers including RTracker \cite{rtracker}, MeMOT \cite{memot}, and multimodal variants like MemVLT \cite{memvlt}.
Despite architectural diversity, memory-based VOS trackers typically update memory using frame-level confidence, similarity metrics, or simple schedules. When predictions are unreliable due to occlusion, motion blur, or distractors, incorrect masks or features may enter memory and cause drift. Existing works explore restricted or hierarchical updates \cite{xmem,rmem}, but they still assume the selected representation is temporally reliable. Explicit short-horizon verification of a mask’s temporal consistency before committing it to memory has received little attention.

\vspace{-1em}
\subsection{SAM2 and Its Extensions}
\noindent SAM2 \cite{sam2} generalizes memory-based reasoning to a promptable segmentation-tracking framework but retains a confidence-driven memory update, making it vulnerable to high-confidence but temporally inconsistent masks. Extensions stabilize this process from different directions: motion-based priors in SAMURAI \cite{samurai}, distractor filtering in DAM4SAM \cite{dam4sam}, long-term memory tree in SAM2Long \cite{sam2long}, and improved scoring or hierarchical features in SAMITE \cite{samite} and HiM2SAM \cite{him2sam}. However, they do not explicitly validate competing candidate masks through short-horizon backtracking before the selected prediction becomes future tracking state.
Across SAM2-based variants \cite{sam2,samurai,sam2long,dam4sam,samite,him2sam}, temporal consistency is treated mainly as a current-frame prediction, scoring, or trajectory-selection problem rather than as an explicit memory-admission decision at the write interface.
We address this with a training-free mechanism that checks short-term consistency before storage, complementing existing SAM2 extensions without modifying their architectures or memory rules.
For plug-in evaluation, we focus on five open-source SAM2-based trackers that operate in an online, frame-by-frame setting \cite{sam2,samurai,dam4sam,samite,him2sam}. We report SAM2Long \cite{sam2long} as a standalone baseline rather than applying SENTRY to it, because SAM2Long performs constrained memory-tree inference over multiple branches, whereas SENTRY is a pre-write admission rule for a single online memory stream. Combining them would require redesigning branch expansion, pruning, or trajectory scoring, making it a new hybrid method rather than a controlled plug-in ablation.

\vspace{-1em}
\section{Method}
\label{sec:method}

\subsection{Preliminaries}
\noindent SAM2 \cite{sam2} consists of an image encoder, a prompt encoder, a memory module, and a mask decoder.
\noindent \textbf{Image Encoder.} A hierarchical ViT backbone (Hiera \cite{hiera}) produces patch tokens for each frame, which serve as visual features queried by the memory-augmented decoder.
\noindent \textbf{Prompt Encoder.} User prompts (points, boxes, or masks) are encoded into tokens. In tracking, only the first frame is prompted; subsequent frames rely solely on memory-guided decoding.
\noindent \textbf{Memory Module.} The memory bank $\mathbf{Mem}^t$ stores embeddings of the initialization mask and a fixed-length queue of recent frame-mask pairs. Temporal cues are added to the recent frames to capture their order, while the initialization frame is left unaltered to serve as a fixed supervised anchor. Each memory token encodes frame features modulated by the predicted mask, and the queue is updated in FIFO order. Incorrect writes to $\mathbf{Mem}^t$ propagate errors forward, causing drift, one of the key failure modes that SENTRY addresses.
\noindent \textbf{Mask Decoder.}
The decoder attends to $\mathbf{Mem}^t$ and the current frame features to produce three candidate hypotheses $\tilde{\mathcal{M}}^t_{\text{dec}} = \{ \tilde{\mathcal{M}}^t_1, \tilde{\mathcal{M}}^t_2, \tilde{\mathcal{M}}^t_3 \}$, each with a predicted IoU score $\hat{\text{IoU}}^t_i$ and occlusion score $\hat{o}^t_i$.
SAM2 selects the mask with the highest $\hat{\text{IoU}}^t_i$ and writes it to memory; however, this selection relies solely on instantaneous confidence and often yields temporally inconsistent memory updates.

\begin{figure*}[!t]
\centering
\includegraphics[width=0.99\linewidth]{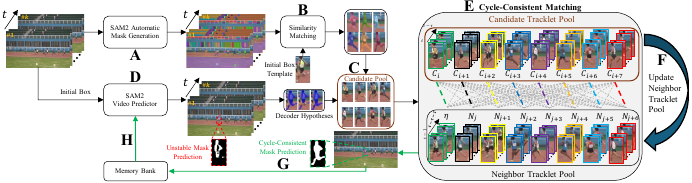}
\vspace{-1em}
\caption{
Overview of the SENTRY framework. Per-frame candidates combine Automatic Mask Generation (AMG) proposals (A–B) and decoder hypotheses (D) into a joint set (C). Each candidate is evaluated through cycle-consistent, neighbor-aware bipartite matching in trajectory space (E–F). The most consistent mask is written to memory (G–H) following the baseline update schedule.
}
\label{fig:3}
\vspace{-2em}
\end{figure*}

\vspace{-1em}
\subsection{SENTRY}
\noindent SENTRY is a training-free, architecture-agnostic module adding a refine-before-write step to SAM2-based trackers.
Unlike prior works that impose temporal consistency through training objectives \cite{aot,xmem}, SENTRY performs explicit consistency validation at inference time, preventing erroneous memory updates online.
Instead of relying on the highest-confidence decoder mask, SENTRY selects the mask with the highest short-horizon temporal consistency before it is written to memory. 
This makes temporal reasoning part of the tracker state transition: the admitted mask becomes the memory state used for future predictions, rather than a post-hoc correction of the current frame.
As shown in Fig. \ref{fig:3}, SENTRY consists of three stages: (i) candidate generation from Automatic Mask Generation (AMG) (A–B) and decoder (D) to form $\mathcal{C}^t$ (C); (ii) cycle-consistent, neighbor-aware reasoning (E–F); and (iii) memory update (G–H).
Algorithm 1 summarizes the refine-before-write procedure.
Intuitively, SENTRY first collects multiple plausible masks for the target, then re-tracks each candidate backward over a short time window to test which one aligns best with the target’s recent motion before memory writing.

\vspace{-2em}
\subsubsection{Candidate Mask Generation.} \label{sec:method_candidate_gen}
\noindent SAM2 generates two sources of per-frame hypotheses: (1) three decoder masks with predicted IoU and occlusion scores, and (2) a set of target-agnostic proposals from AMG. SENTRY filters and regularizes these hypotheses into a compact, diverse candidate set that preserves plausible target states.

\noindent \emph{Decoder hypotheses.}
The decoder outputs masks $\tilde{\mathcal{M}}^t_{\text{dec}}=\{\tilde{\mathcal{M}}^t_i\}_{i=1}^3$ with corresponding IoU scores $\hat{\text{IoU}}^t_i$ and occlusion scores $\hat{o}^t_i$. 
SENTRY retains hypotheses that are (i) close to the best decoder prediction and (ii) not flagged as occluded:
\begin{equation}
\mathcal{C}^t_{\text{dec}} =
\left\{ \tilde{\mathcal{M}}^t_i \mid 
\hat{\text{IoU}}^t_{i} > \alpha \max_j \hat{\text{IoU}}^t_{j} \land \hat{o}^t_i > 0 \right\},
\end{equation}
where $\alpha\!\in\![0,1]$ controls tolerance to decoder uncertainty.

\noindent \emph{AMG proposals.} AMG generates \(N^t\) target-agnostic proposals $\{ \tilde{\mathcal{M}}^t_{\text{AMG},j}\}$. Let the image regions covered by these masks be
\(\mathcal{R}^t_j\).
To identify proposals relevant to the target, SENTRY computes cosine similarity between proposal features and a first-frame template $\mathcal{T}^0$, defined as the image region covered by the initialization mask $\bar{\mathcal{M}}^0$.
Let $f(\cdot)$ denote image encoder features averaged over the region. The proposal template similarity is:
\begin{equation}
s_j = \frac{f(\mathcal{R}^t_{j}) \cdot f(\mathcal{T}^0)}
{\| f(\mathcal{R}^t_{j}) \|_2 \, \| f(\mathcal{T}^0) \|_2}.
\end{equation}
We keep proposals that match the target relatively well:
\begin{equation}
    \mathcal{C}^t_{\mathrm{AMG}}
    =
    \left\{
        \tilde{\mathcal{M}}^{t}_{\mathrm{AMG},j}
        \,\middle|\,
        s_j \ge \beta \max_k s_k
    \right\},
\end{equation}
with $\beta\!\in\![0,1]$ controlling tolerance to appearance variation.
Both thresholds (\(\alpha\) and \(\beta\)) are fixed globally across all datasets (see App. \ref{supplement: Additional Ablation Study}).

\noindent \emph{Candidate filtration.} The union $\bar{\mathcal{C}}^t=\mathcal{C}^t_{\text{dec}}\cup\mathcal{C}^t_{\mathrm{AMG}}$ may contain redundant or overlapping hypotheses. Each mask is projected to its minimum-enclosing bounding box, and Soft-NMS \cite{softnms} is applied in box space to suppress near duplicates while preserving spatial diversity. The remaining boxes are mapped back to their masks, yielding the refined candidate set $\hat{\mathcal{C}}^t$.

\noindent \emph{Kalman motion prior.}
To maintain robustness under heavy occlusion, SENTRY includes a motion-only fallback derived from a constant-acceleration Kalman filter. Using the center and scale of the previous refined mask $\bar{\mathcal{M}}^{t-1}$, the filter predicts a new box center $b_{\kappa}^t$. A motion-derived mask $\mathcal{M}^t_{\kappa}$ is produced by translating $\bar{\mathcal{M}}^{t-1}$ so its bounding box aligns with $b_{\kappa}^t$. Its shape is fixed since it is used only as a fallback. This prior is appended to the candidate pool, \(\mathcal{C}^t = \hat{\mathcal{C}}^t \cup \{\mathcal{M}^t_{\kappa}\}\), but excluded from Soft-NMS, since its role is to provide a motion-only fallback rather than compete with appearance-derived proposals.

\vspace{-1em}
\subsubsection{Neighbor Reasoning and Matching.}
Each candidate is evaluated through short-horizon temporal reasoning. SENTRY builds backward-propagated candidate tracklets, forward-propagated neighbor tracklets for recent distractors, and performs cycle-consistent matching to select the mask with the strongest temporal support.
Neighbor-aware means that candidates whose trajectories align strongly with distractor motions are penalized during matching, reducing the chance of drift in cluttered scenes.

\noindent \emph{Candidate tracklet pool.} A mask that appears plausible at frame \(t\) can still be inconsistent with the target’s recent motion. To evaluate temporal reliability, SENTRY builds a short backward tracklet for each candidate $\mathcal{M}^t_i \in \mathcal{C}^t$. The mask $\mathcal{M}^t_i$ is used as a dense prompt in SAM2’s promptable segmentation, and the previous $\tau$ frames $(I^{t-1}, \ldots, I^{t-\tau})$ are re-segmented to obtain \(\xi^t_i = (\mathcal{M}^t_i,\, \mathcal{M}^{t-1}_i,\, \ldots,\, \mathcal{M}^{t-\tau}_i)\). These newly generated backward segmentations form the candidate tracklet pool, $\mathbf{P}_c^t = \{\xi^t_i \mid \mathcal{M}^t_i \in \mathcal{C}^t\}$. 
For consistency evaluation, SENTRY maintains a reference trajectory $\eta^t = (\bar{\mathcal{M}}^{t}, \ldots, \bar{\mathcal{M}}^{t-\tau})$ constructed from previously selected refined masks. This trajectory serves as a temporal anchor against which all candidate tracklets are evaluated.
The temporal consistency length $\tau$ is fixed to 10 for all datasets, with runtime analysis in Sec. \ref{sec:ablation_study}.

\noindent \emph{Neighbor tracklet pool.}
To provide context during ambiguity, SENTRY maintains a small set of distractor tracklets.  This pool encodes \textit{what the target is not}, giving contextual cues when candidates become ambiguous or visually similar.
After selecting $\bar{\mathcal{M}}^t$, the remaining candidate tracklets from frame $t$ are retained as neighbors (i.e., masks not chosen for update). Each neighbor is updated one step forward by re-segmenting $I^{t+1}$ using its last mask as a prompt, keeping all sequences aligned within the $\tau$-frame window. 
The resulting pool $\mathbf{P}_n^{t+1}$ serves as a negative motion context: candidates whose trajectories align with these distractors are penalized during cycle-consistent matching. The pool requires no explicit pruning; neighbor tracklets naturally expire once they fall outside the $\tau$-frame window. This keeps the pool small, prevents unbounded growth, and avoids long-term drift.

\noindent \emph{Cycle-consistent matching.}
Given candidate and neighbor tracklets, SENTRY evaluates each hypothesis by comparing its backward-propagated trajectory against the forward-aligned target and neighbor tracklets over the same temporal window. Following the notion of cycle consistency used in \cite{neighbourtrack}, a candidate is considered reliable if tracking it backward produces a trajectory that overlaps well with the target’s recent forward track; conversely, candidates whose trajectories align more strongly with neighbor motion are treated as distractors.
For every candidate tracklet $\xi^t_i \in \mathbf{P}_c^t$ and neighbor or target tracklet $\zeta^t_j \in \mathbf{P}_n^t \cup \{\eta^t\}$, we compute a trajectory similarity score
\begin{equation} \label{eq:sim}
\mathrm{Sim}(\xi_i,\zeta_j)
= \frac{1}{\tau} \sum_{k=t-\tau}^{t-1} 
\mathrm{IoU}(b_{i}^k,\, b_{j}^k),
\end{equation}
where $b_{i}^k$ and $b_{j}^k$ are the corresponding boxes at frame $k$.
These similarities form a bipartite matrix over candidates and neighbors/target. A complete bipartite graph is constructed with weights $\mathrm{Sim}(\xi_i,\zeta_j)$, and the Hungarian algorithm \cite{hungarian_algorithm} is applied to obtain a one-to-one assignment that maximizes total trajectory agreement. This matching acts as a motion-consistency filter: candidates that are temporally aligned with the target are favored, while those that match neighbor trajectories are suppressed. 
Although final selection uses candidate–target similarity, the Hungarian assignment implicitly penalizes candidates aligned with distractor motion. This assignment serves purely as a consistency filter; the final mask is determined solely by candidate–target similarity $s_i$ under the threshold rules.

Let $s_i = \mathrm{Sim}(\xi^t_i,\eta^t)$ denote each candidate’s similarity to the target trajectory. 
We select the candidate with the highest $s_i$ if it exceeds a reliability threshold $\theta_{\mathrm{rel}}$ ($\max_i s_i \ge \theta_{\mathrm{rel}}$). 
When no strong match exists but at least one candidate is temporally plausible ($\max_i s_i \ge \theta_{\mathrm{min}}$), the highest-scoring candidate is used. 
If all candidates exhibit weak or inconsistent motion ($\max_i s_i < \theta_{\mathrm{min}}$), SENTRY falls back to the Kalman prior $\mathcal{M}^t_{\kappa}$, which provides a motion-only estimate during severe occlusion or appearance failure. 
The selected mask $\bar{\mathcal{M}}^t$ is then written to memory under SAM2’s default update schedule.

The thresholds $\theta_{\mathrm{rel}}$ and $\theta_{\mathrm{min}}$ are fixed for all datasets (see App. \ref{supplement: Additional Ablation Study}). 
IoU-based trajectory similarity is used instead of appearance distance because appearance cues become unreliable under occlusion, clutter, or lighting changes. 
Backward propagation uses each candidate mask as a segmentation prompt, so appearance information is already implicitly embedded in the resulting tracklet; using geometric motion as the comparison metric yields a more stable and robust signal \cite{neighbourtrack}.
\vspace{-2em}
\begin{algorithm}[!h] \label{algorithm1}
\caption{\small SENTRY: Refine-Before-Write Tracking}
\begin{algorithmic}[1]
\Require \small Video frames $\{I_t\}_{t=0}^{T}$; initial mask $\bar{\mathcal{M}}^0$; memory bank $\mathbf{Mem}^0$; image encoder $f(\cdot)$; sampled-point grid $\mathcal{G}$; backtrack horizon $\tau$

\State \textbf{Init}:
    \State \quad $\mathbf{Mem}^0 \gets \mathbf{Mem}^0 \cup \{\bar{\mathcal{M}}^0\}$
    \State \quad $\mathbf{P}_{\,n}^0 \gets \emptyset$
    
\State \textbf{for} $(I^t) \in (I^T)$ \textbf{do}:

\State \quad $F^t \gets f(I^t)$
\State \quad $\mathcal{C}^{\,\text{dec}}_t \gets \text{DecodeCandidates}(F^t, \mathbf{Mem}^{t-1})$
\State \quad $\mathcal{C}_{\,\text{AMG}}^t \gets \text{DecodeFromGrid}(F^t, \mathcal{G})$
\State \quad $\hat{\mathcal{C}}^t \gets \text{Soft-NMS}(\mathcal{C}_{\,\text{dec}}^t \cup \mathcal{C}_{\,\text{AMG}}^t)$

\State \quad $\mathcal{M}_{\,\kappa}^t \gets \text{MaskFromKF}(\text{BBox}(\bar{\mathcal{M}}^{t-1}))$
\State \quad $\mathcal{C}^t \gets \hat{\mathcal{C}}^t \cup \{\mathcal{M}_{\,\kappa}^t\}$

\State \quad $\mathbf{P}_{\,c}^t \gets \{\text{BacktrackPoints}(\mathcal{M}_i^t, \tau) \mid \mathcal{M}_i^t \in \mathcal{C}^t\}$
\State \quad $\eta^t \gets \text{CycleConsistentMatch}(\mathbf{P}_{\,c}^t, \mathbf{P}_{\,n}^{t-1})$
\State \quad $\mathbf{P}_{\,n}^t \gets \mathbf{P}_{\,c}^t \setminus \eta^t$
\State \quad $\mathbf{Mem}^t \gets \text{UpdateMemory}(\mathbf{Mem}^{t-1}, \eta^t)$
\State \quad $\bar{\mathcal{M}}^t \gets \text{SelectOutputMask}(\eta^t, \mathcal{M}_{\,\kappa}^t)$

\State \textbf{end for}

\end{algorithmic}
\end{algorithm}
%
\vspace{-3.5em}
\section{Experiments}
\label{sec:results}

\noindent We integrate SENTRY into five SAM2-based trackers: SAM2 \cite{sam2}, SAMURAI \cite{samurai}, DAM4SAM \cite{dam4sam}, SAMITE \cite{samite}, and HiM2SAM \cite{him2sam} (the latter two are evaluated in App. \ref{supplement: Extended evaluations on additional SAM2-based frameworks}).
We denote SENTRY applied to SAM2, SAMURAI, and DAM4SAM as SENTRY-S2, SENTRY-SR, and SENTRY-D4S.
In App. \ref{supplement:generality_beyond_sam2}, we further report a small generality study on non-SAM memory trackers and other SAM-family trackers, suggesting that the benefit stems from verified memory writes rather than SAM2-specific internals.
We evaluate all models across nine standard tracking benchmarks covering long-term and short-term bounding box tracking (LaSOT, LaSOT\textsubscript{ext}, TNL2K, GOT-10k, TrackingNet) and the VOT/DiDi protocols. Each dataset’s official metrics are used. Comprehensive tables and per-scale results are provided in App. \ref{supplement: Additional quantitative results on bounding box benchmarks} and App. \ref{supplement: Additional quantitative results on VOT benchmarks}, and ablation studies on scalability, candidate generation, and the temporal consistency window $\tau$ appear in Sec. \ref{sec:ablation_study}.
We additionally evaluate VOS benchmarks under the first-frame-mask protocol in App. \ref{supplement:additional_vos_results}, where SENTRY improves each corresponding host tracker.
For fair comparison, we reproduce any missing baseline results using each method’s official implementation under the same evaluation protocol. In the quantitative tables, the top three entries in each column are marked using \textcolor{red}{\textbf{red}} for best, \textcolor{green}{\textbf{green}} for second, and \textcolor{blue}{\textbf{blue}} for third. Relative improvements of each SENTRY variant over its baseline are reported as percentage-point gains, shown in \textcolor{gray}{\textbf{($\uparrow$)}}.

\begin{table*}[!t]
\centering

\caption{SENTRY provides consistent improvements across baselines and datasets.}
\vspace{-1em}
\setlength{\tabcolsep}{1pt}
		\scalebox{0.4}[0.5]{

\begin{tabular}{ll|ccc|ccc|ccc|ccc|ccc}

\hline

\rowcolor{Gray} 
& Method
& \multicolumn{3}{c|}{LaSOT \cite{lasot}}
& \multicolumn{3}{c|}{LaSOT\textsubscript{ext} \cite{lasotext}}
& \multicolumn{3}{c|}{TNL2K \cite{tnl2k}}
& \multicolumn{3}{c|}{GOT-10k \cite{got10k}}
& \multicolumn{3}{c}{TrackingNet \cite{trackingnet}}
\\

\cline{3-17}
\rowcolor{Gray} 
& 
& S & NP & P 
& S & NP & P 
& S & NP & P 
& AO & SR\textsubscript{0.50} & SR\textsubscript{0.75} 
& S & NP & P 
\\

\hline
 
\parbox[t]{0.25cm}{\multirow{5}{*}{\rotatebox[origin=c]{90}{\begin{tabular}[l]{@{}l@{}}\textbf{Vision-Based}\end{tabular}}}}

& DiffusionTrack \cite{diffusiontrack} 
& 70.8 & 79.8 & 76.7 
& -- & -- & -- 
& 56.4 & 72.5 & 57.3 
& 74.8 & 85.4 & 72.0 
& 83.8 & 88.2 & 82.1 
\\

& HIPTrack \cite{hiptrack} 
& 72.7 & 82.9 & 79.5 
& 53.0 & 64.3 & 60.6 
& -- & -- & -- 
& 77.4 & 88.0 & 74.5 
& 84.5 & 89.1 & 83.8 
\\

& AQATrack\textsubscript{256} \cite{aqatrack} 
& 71.4 & 81.9 & 78.6 
& 51.2 & 62.2 & 58.9 
& 57.8 & 59.4 & -- 
& 73.8 & 83.2 & 72.1 
& 83.8 & 88.6 & 83.1 
\\

& ARPTrack\textsubscript{256} \cite{arptrack} 
& 72.6 & 81.4 & 78.5 
& 52.0 & 62.9 & 58.7 
& -- & -- & -- 
& 77.7 & 87.3 & 74.3 
& 85.5 & 90.0 & 85.3 
\\

& SPMTrack-B \cite{spmtrack} 
& \textcolor{blue}{\textbf{74.9}} & 84.0 & \textcolor{blue}{\textbf{81.7}} 
& -- & -- & -- 
& 62.0 & 79.7 & 66.7 
& 76.5 & 85.9 & 76.3 
& \textcolor{red}{\textbf{86.1}} & 90.2 & 85.6 
\\

\hline

\parbox[t]{0.25cm}{\multirow{5}{*}{\rotatebox[origin=c]{90}{\textbf{VLM}}}}

& UVLTrack-B \cite{uvltrack} 
& 69.4 & -- & 74.9 
& 49.2 & -- & 55.8 
& 62.7 & -- & 65.4 
& -- & -- & -- 
& 83.4 & -- & 82.1 
\\

& QueryNLT \cite{querynlt} 
& 59.9 & 69.6 & 63.5 
& -- & -- & -- 
& 57.8 & 75.6 & 58.7 
& -- & -- & -- 
& -- & -- & -- 
\\

& DUTrack\textsubscript{384} \cite{dutrack} 
& 74.1 & \textcolor{green}{\textbf{84.9}} & \textcolor{red}{\textbf{82.9}} 
& 52.5 & 63.6 & 60.5 
& \textcolor{green}{\textbf{65.6}} & \textcolor{green}{\textbf{83.2}} & \textcolor{red}{\textbf{71.9}} 
& 77.8 & -- & -- 
& -- & -- & -- 
\\

& MambaVLT \cite{mambavlt} 
& 66.6 & 77.3 & 71.0 
& -- & -- & -- 
& \textcolor{red}{\textbf{66.5}} & \textcolor{red}{\textbf{90.9}} & \textcolor{green}{\textbf{69.9}} 
& -- & -- & -- 
& -- & -- & -- 
\\

& CLDTracker \cite{cldtracker} 
& 74.0 & 83.9 & 81.1 
& 53.1 & 64.8 & 60.6 
& 61.5 & \textcolor{blue}{\textbf{82.2}} & 64.3 
& 77.5 & 85.4 & 75.6 
& 85.1 & 89.7 & 84.9 
\\

\hline

\parbox[t]{0.25cm}{\multirow{9}{*}{\rotatebox[origin=c]{90}{\textbf{Memory-Based}}}}

& MemVLT \cite{memvlt} 
& 72.9 & \textcolor{red}{\textbf{85.7}} & 80.5 
& 52.1 & 63.3 & 59.8 
& \textcolor{blue}{\textbf{63.3}} & 80.9 & 67.4 
& -- & -- & -- 
& -- & -- & -- 
\\

& RTracker-L \cite{rtracker} 
& 74.7 & 84.5 & -- 
& 54.9 & 65.5 & 62.7 
& 60.6 & -- & 63.7 
& 77.9 & 87.0 & 76.9 
& -- & -- & -- 
\\
\cline{2-17}
& \textit{\textbf{Zero-shot Method}} 
&  &  & 
&  &  & 
&  &  & 
&  &  & 
\\

& SAM2-L \cite{sam2} 
& 68.5 & 76.1 & 73.6 
& 56.8 & 71.1 & 67.0 
& 56.7 & 75.4 & 62.5 
& 80.8 & 91.3 & 75.5 
& 85.3 & \textcolor{blue}{\textbf{91.3}} & \textcolor{green}{\textbf{88.2}} 
\\

& SAMURAI-L \cite{samurai} 
& 74.2 & 82.7 & 80.2 
& \textcolor{blue}{\textbf{61.0}} & \textcolor{blue}{\textbf{73.9}} & \textcolor{blue}{\textbf{72.2}} 
& 50.6 & 67.5 & 54.2 
& \textcolor{blue}{\textbf{81.7}} & \textcolor{blue}{\textbf{92.2}} & 76.9 
& 85.3 & 88.2 & 85.0 
\\

& DAM4SAM-L \cite{dam4sam} 
& \textcolor{green}{\textbf{75.1}} & 83.3 & 81.1 
& 60.9 & \textcolor{green}{\textbf{75.3}} & \textcolor{blue}{\textbf{72.2}} 
& 59.8 & 79.8 & 66.8 
& 81.1 & 91.4 & \textcolor{green}{\textbf{77.2}} 
& 85.3 & 90.9 & 87.4 
\\

\cline{2-17}

& SENTRY-S2-L 
& 70.2 \footnotesize{(\textcolor{gray}{\textbf{1.7$\uparrow$}})} & 77.2 \footnotesize{(\textcolor{gray}{\textbf{1.1$\uparrow$}})} & 74.5 \footnotesize{(\textcolor{gray}{\textbf{0.9$\uparrow$}})} 
& 57.0 \footnotesize{(\textcolor{gray}{\textbf{0.2$\uparrow$}})} & 71.7 \footnotesize{(\textcolor{gray}{\textbf{0.6$\uparrow$}})} & 67.1 \footnotesize{(\textcolor{gray}{\textbf{0.1$\uparrow$}})} 
& 57.9 \footnotesize{(\textcolor{gray}{\textbf{1.2$\uparrow$}})} & 76.9 \footnotesize{(\textcolor{gray}{\textbf{1.5$\uparrow$}})} & 64.1 \footnotesize{(\textcolor{gray}{\textbf{1.6$\uparrow$}})} 
& 81.1 \footnotesize{(\textcolor{gray}{\textbf{0.3$\uparrow$}})} & 91.4 \footnotesize{(\textcolor{gray}{\textbf{0.1$\uparrow$}})} & 76.5 \footnotesize{(\textcolor{gray}{\textbf{1.0$\uparrow$}})} 
& 85.7 \footnotesize{(\textcolor{gray}{\textbf{0.4$\uparrow$}})} & \textcolor{red}{\textbf{91.9}} \footnotesize{(\textcolor{gray}{\textbf{0.6$\uparrow$}})} & \textcolor{red}{\textbf{88.9}} \footnotesize{(\textcolor{gray}{\textbf{0.7$\uparrow$}})} 
\\ 

& SENTRY-SR-L 
& \textcolor{green}{\textbf{75.1}} \footnotesize{(\textcolor{gray}{\textbf{0.9$\uparrow$}})} & 82.7 & 80.4 \footnotesize{(\textcolor{gray}{\textbf{0.2$\uparrow$}})} 
& \textcolor{green}{\textbf{61.5}} \footnotesize{(\textcolor{gray}{\textbf{0.5$\uparrow$}})} & 75.0 \footnotesize{(\textcolor{gray}{\textbf{1.1$\uparrow$}})} & \textcolor{green}{\textbf{72.9}} \footnotesize{(\textcolor{gray}{\textbf{0.7$\uparrow$}})} 
& 59.6 \footnotesize{(\textcolor{gray}{\textbf{9.0$\uparrow$}})} & 78.8 \footnotesize{(\textcolor{gray}{\textbf{11.3$\uparrow$}})} & 66.4 \footnotesize{(\textcolor{gray}{\textbf{12.2$\uparrow$}})} 
& \textcolor{green}{\textbf{81.8}} \footnotesize{(\textcolor{gray}{\textbf{0.1$\uparrow$}})} & \textcolor{green}{\textbf{92.3}} \footnotesize{(\textcolor{gray}{\textbf{0.1$\uparrow$}})} & \textcolor{blue}{\textbf{77.1}} \footnotesize{(\textcolor{gray}{\textbf{0.2$\uparrow$}})} 
& \textcolor{blue}{\textbf{85.8}} \footnotesize{(\textcolor{gray}{\textbf{0.5$\uparrow$}})} & 91.1 \footnotesize{(\textcolor{gray}{\textbf{2.9$\uparrow$}})} & \textcolor{blue}{\textbf{88.1}} \footnotesize{(\textcolor{gray}{\textbf{3.1$\uparrow$}})} 
\\ 

& SENTRY-D4S-L 
& \textcolor{red}{\textbf{76.3}} \footnotesize{(\textcolor{gray}{\textbf{1.2$\uparrow$}})}  & \textcolor{blue}{\textbf{84.7}} \footnotesize{(\textcolor{gray}{\textbf{1.4$\uparrow$}})} & \textcolor{green}{\textbf{82.4}} \footnotesize{(\textcolor{gray}{\textbf{1.3$\uparrow$}})} 
& \textcolor{red}{\textbf{61.8}} \footnotesize{(\textcolor{gray}{\textbf{0.9$\uparrow$}})} & \textcolor{red}{\textbf{76.6}} \footnotesize{(\textcolor{gray}{\textbf{1.3$\uparrow$}})} & \textcolor{red}{\textbf{73.8}} \footnotesize{(\textcolor{gray}{\textbf{1.6$\uparrow$}})} 
& 61.3 \footnotesize{(\textcolor{gray}{\textbf{1.5$\uparrow$}})} & 81.3 \footnotesize{(\textcolor{gray}{\textbf{1.5$\uparrow$}})} & \textcolor{blue}{\textbf{68.3}} \footnotesize{(\textcolor{gray}{\textbf{1.5$\uparrow$}})} 
& \textcolor{red}{\textbf{82.1}} \footnotesize{(\textcolor{gray}{\textbf{1.0$\uparrow$}})} & \textcolor{red}{\textbf{92.6}} \footnotesize{(\textcolor{gray}{\textbf{1.2$\uparrow$}})} & \textcolor{red}{\textbf{78.2}} \footnotesize{(\textcolor{gray}{\textbf{1.0$\uparrow$}})} 
& \textcolor{green}{\textbf{85.9}} \footnotesize{(\textcolor{gray}{\textbf{0.6$\uparrow$}})} & \textcolor{green}{\textbf{91.5}} \footnotesize{(\textcolor{gray}{\textbf{0.6$\uparrow$}})} & 87.9 \footnotesize{(\textcolor{gray}{\textbf{0.5$\uparrow$}})} 
\\

\hline

\end{tabular}
}
\label{tab:bbox_results}
\vspace{-1em}
\end{table*}

\subsection{SOTA comparison on box benchmarks} \label{sec:sota_bbox}
\begin{figure*}[!t]
\centering
\begin{subfigure}[b]{0.325\textwidth}
\includegraphics[width=\textwidth]{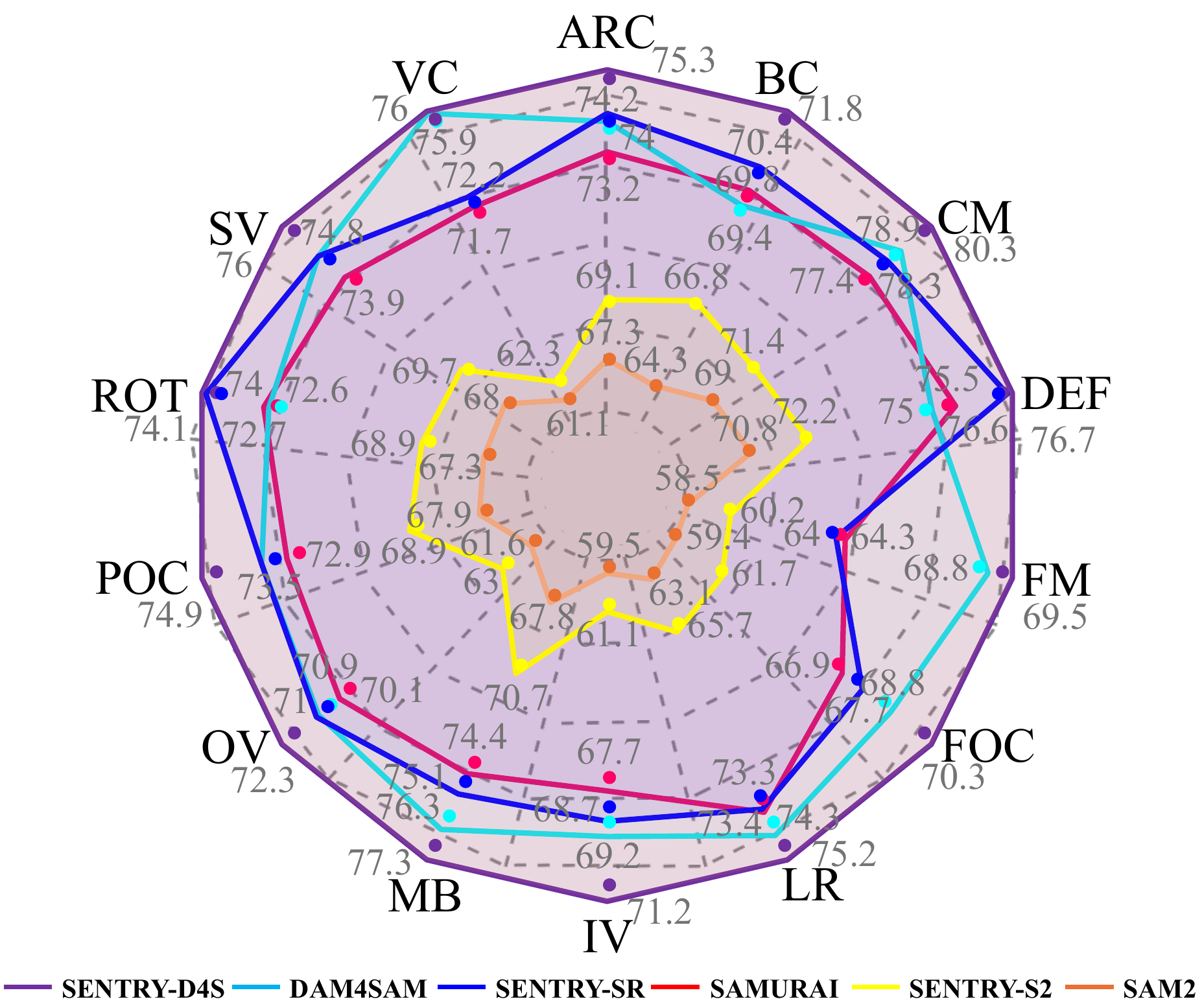}
\caption{}
\end{subfigure}
\begin{subfigure}[b]{0.325\textwidth}
\includegraphics[width=\textwidth]{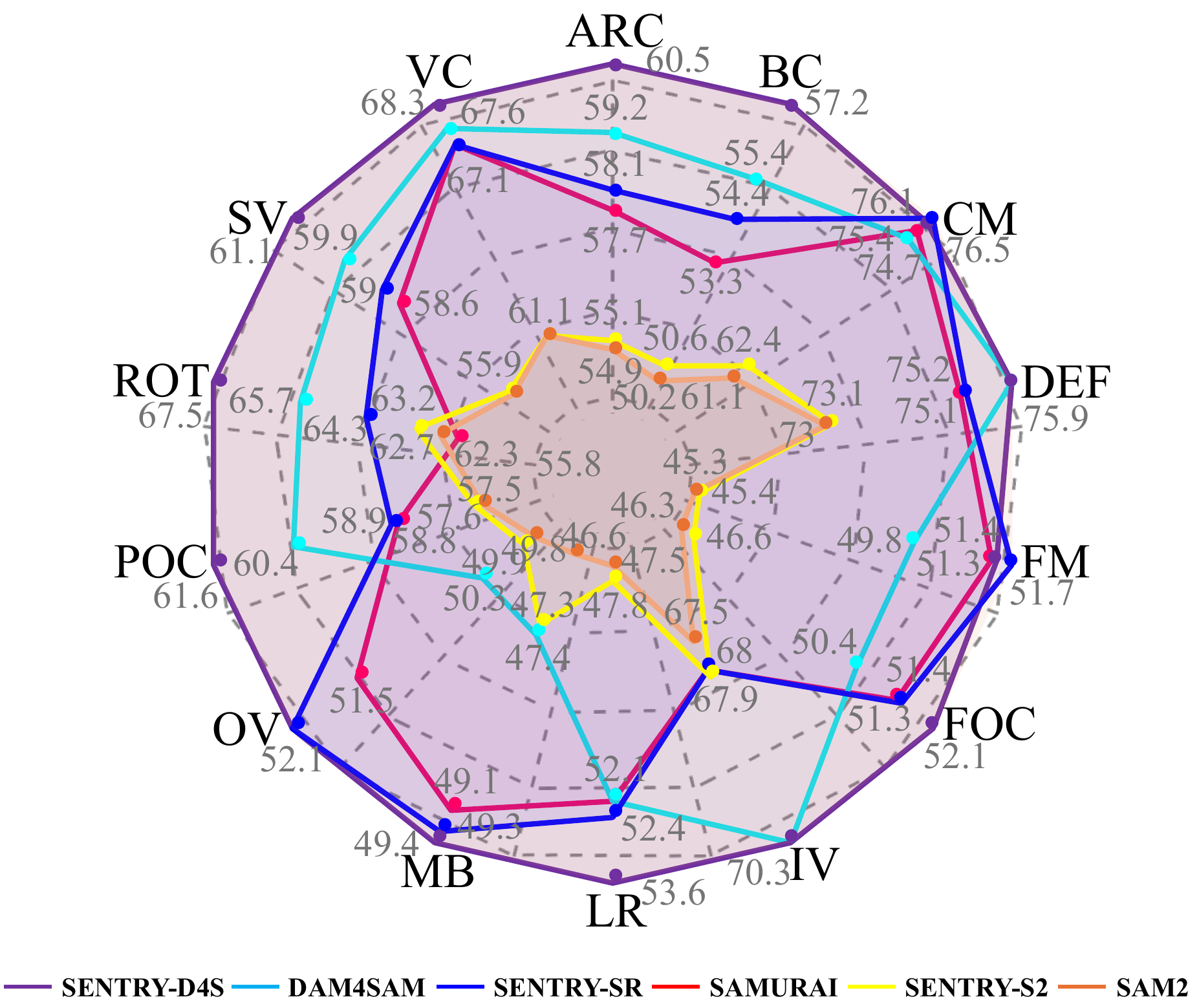}
\caption{}
\end{subfigure}
\begin{subfigure}[b]{0.325\textwidth}
\includegraphics[width=\textwidth]{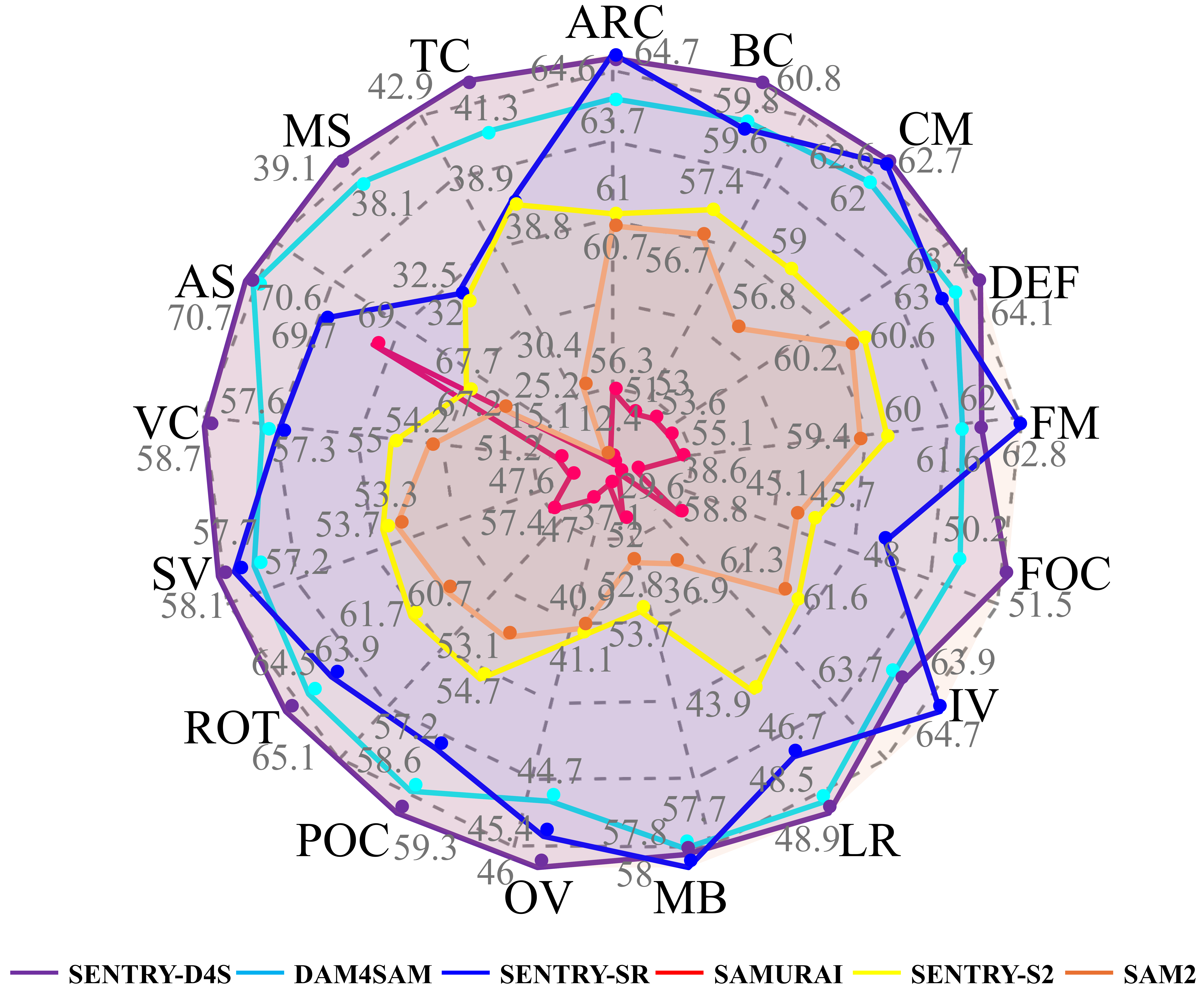}
\caption{}
\end{subfigure}
\vspace{-1em}
\caption{AUC of attributes on (a) LaSOT \cite{lasot}, (b) LaSOT\textsubscript{ext} \cite{lasotext}, and (c) TNL2K \cite{tnl2k}.}
\label{fig:vot_attributes}
\vspace{-2em}
\end{figure*}

\noindent Results on LaSOT, LaSOT\textsubscript{ext}, TNL2K, GOT-10k, and TrackingNet are summarized in Tab. \ref{tab:bbox_results}. Full results across all trackers and model scales appear in App. \ref{supplement: Additional quantitative results on bounding box benchmarks} (Tab. \ref{tab:sam2_comparison}). Attribute-level analyses are shown in Figs. \ref{fig:vot_attributes}(a)–(c) and in App. Tabs. \ref{tab:lasot_attribute}, \ref{tab:lasotext_attribute}, and \ref{tab:attribute_tnl2k}.
SENTRY’s gains follow each dataset’s difficulty profile: smaller improvements on short or clean sequences (TrackingNet), moderate improvements on heterogeneous or appearance-diverse benchmarks (GOT-10k, LaSOT\textsubscript{ext}), and larger improvements on long or rapidly changing sequences where baseline errors accumulate more (LaSOT, TNL2K).

\noindent \textbf{LaSOT \cite{lasot}.} 
All SENTRY variants outperform their SAM2-based baselines across all metrics. 
SENTRY-D4S gains 1.6\% S, 1.7\% NP, and 1.6\% P over DAM4SAM; SENTRY-SR improves SAMURAI by 1.2\% S and 0.2\% P; and SENTRY-S2 surpasses SAM2 by 2.5\% S, 1.4\% NP, and 1.2\% P. 
LaSOT’s long sequences amplify small mask errors into drift, making it a setting where SENTRY’s stability-focused updates yield substantial improvements.
\noindent \textbf{LaSOT\textsubscript{ext} \cite{lasotext}.} 
SENTRY-D4S sets a new SOTA on LaSOT\textsubscript{ext}, improving over DAM4SAM by 1.5\% S, 1.7\% NP, and 2.2\% P.
Other variants show consistent gains: SENTRY-S2 adds 0.4\% S, 0.8\% NP, and 0.1\% P, while SENTRY-SR improves SAMURAI by 0.8\% S, 1.5\% NP, and 1.0\% P. 
LaSOT\textsubscript{ext} contains many small objects and unseen categories, which frequently destabilize baseline predictions; SENTRY mitigates these cases, resulting in consistent gains.
\noindent \textbf{TNL2K \cite{tnl2k}.} 
All SENTRY variants outperform their SAM2-based baselines. 
SENTRY-D4S gains 2.5\% S, 1.9\% NP, and 2.2\% P over DAM4SAM; SENTRY-S2 adds 2.1\% S, 2.0\% NP, and 2.6\% P; and SENTRY-SR yields the largest boost of 17.8\% S, 16.7\% NP, and 22.5\% P over SAMURAI. 
TNL2K contains frequent appearance shifts, occlusions, and strongly non-linear motion, which disrupt SAMURAI’s Kalman filter-based linear update model; in contrast, SENTRY’s refine-before-write mechanism validates each memory update via temporal consistency rather than motion assumptions, yielding its largest gains specifically on this benchmark.
%
%
\noindent \textbf{GOT-10k \cite{got10k}.} 
SENTRY-D4S achieves the best overall performance and sets a new SOTA.
SENTRY-SR ranks second on AO and SR\textsubscript{0.5}, and third on SR\textsubscript{0.75}. 
Operating fully zero-shot, SENTRY-D4S even outperforms fine-tuned competitors, highlighting the scalability of reliability-aware refinement. 
Relative gains over baselines include 1.2\% AO, 1.3\% SR\textsubscript{0.5}, and 1.3\% SR\textsubscript{0.75} for DAM4SAM; 0.4\%, 0.1\%, and 1.3\% for SAM2; and 0.1\%, 0.1\%, and 0.3\% for SAMURAI.
GOT-10k’s category and appearance diversity induces mild inconsistencies that SENTRY stabilizes.
\noindent \textbf{TrackingNet \cite{trackingnet}.} 
All SENTRY variants improve over their respective baselines. 
Integrating SENTRY into DAM4SAM yields 0.7\% S, 0.7\% NP, and 0.6\% P; applying it to SAM2 adds 0.5\% S, 0.7\% NP, and 0.8\% P; and enhancing SAMURAI provides the largest gains of 0.6\% S, 3.3\% NP, and 3.6\% P. 
TrackingNet’s short, clean sequences leave limited room for correction, so SENTRY primarily smooths small prediction fluctuations, producing modest but consistent gains.
\noindent \textbf{Attribute-Wise Performance.} 
Across all datasets (Figs. \ref{fig:vot_attributes}(a)–(c)), SENTRY delivers its strongest gains in conditions that destabilize per-frame masks, such as BC, motion changes, LR, and occlusions. These patterns are consistent across SAM2, SAMURAI, and DAM4SAM, reflecting the shared sensitivity of SAM2-style memory updates to noisy frames.
Differences emerge at the dataset level. On LaSOT, the long sequences magnify issues like MB and CM, making drift suppression the dominant source of improvement.
On LaSOT\textsubscript{ext}, many targets are small or visually scarce, so gains are driven by better handling of LR views, weak cues, and occlusions.
On TNL2K, sudden appearance changes and non-linear motion are the main challenges, and SENTRY’s filtering stabilizes these rapid transitions.
Across benchmarks, SENTRY consistently removes unreliable updates, but the specific attributes that benefit most align with each dataset’s characteristic failure modes.
\begin{table}[t]
\centering
\setlength{\tabcolsep}{14pt}

\begin{minipage}[t]{0.49\columnwidth}
\centering
\caption{\small Comparison on the VOT20 \cite{vot2020}.}
\vspace{-1em}
\label{tab:results_vot20}
\scalebox{0.48}{
\begin{tabular}{l|ccc}
\hline
\rowcolor{Gray}
Method & \multicolumn{3}{c}{VOT20} \\
\rowcolor{Gray}
& Q & Acc & Rob \\
\hline
ODTrack \cite{odtrack}
& 0.605 & 0.761 & 0.902
\\

MixViT-L+AR \cite{mixvit}
& 0.584 & 0.755 & 0.890
\\

SeqTrack-L \cite{seqtrack}
& 0.561 & - & -
\\

MixFormer-L \cite{mixformer}
& 0.555 & 0.762 & 0.855
\\

RPT \cite{rpt}
& 0.530 & 0.700 & 0.869
\\

OceanPlus \cite{ocean}
& 0.491 & 0.685 & 0.842
\\

AlphaRef \cite{alpharef}
& 0.482 & 0.754 & 0.777
\\

AFOD \cite{afod}
& 0.472 & 0.713 & 0.795
\\

\specialrule{1pt}{0pt}{0pt}

\textit{\textbf{Zero-shot Method}} \\

SAM2-L \cite{sam2}
& 0.681 & 0.778 & 0.941
\\

SAMURAI-L \cite{samurai}
& 0.674 & 0.741 & 0.942
\\

DAM4SAM-L \cite{dam4sam}
& \textcolor{green}{\textbf{0.723}} & \textcolor{green}{\textbf{0.796}} & \textcolor{green}{\textbf{0.961}}
\\

\hline

SENTRY-S2-L 
& \textcolor{blue}{\textbf{0.690}} \footnotesize{(\textcolor{gray}{\textbf{0.9$\uparrow$}})} & \textcolor{blue}{\textbf{0.785}} \footnotesize{(\textcolor{gray}{\textbf{0.7$\uparrow$}})} & \textcolor{blue}{\textbf{0.949}} \footnotesize{(\textcolor{gray}{\textbf{0.8$\uparrow$}})}
\\

SENTRY-SR-L 
& 0.686 \footnotesize{(\textcolor{gray}{\textbf{1.2$\uparrow$}})} & 0.775 \footnotesize{(\textcolor{gray}{\textbf{3.4$\uparrow$}})} & 0.947 \footnotesize{(\textcolor{gray}{\textbf{0.5$\uparrow$}})}
\\
SENTRY-D4S-L 
& \textcolor{red}{\textbf{0.732}} \footnotesize{(\textcolor{gray}{\textbf{0.9$\uparrow$}})} & \textcolor{red}{\textbf{0.804}} \footnotesize{(\textcolor{gray}{\textbf{0.8$\uparrow$}})} & \textcolor{red}{\textbf{0.970}} \footnotesize{(\textcolor{gray}{\textbf{0.9$\uparrow$}})}
\\
\hline
\end{tabular}
}
\vspace{-1em}
\end{minipage}
\hfill
\begin{minipage}[t]{0.49\columnwidth}
\centering
\caption{\small Comparison on the VOT22 \cite{vot2022}.}
\vspace{-1em}
\label{tab:results_vot22}
\scalebox{0.48}{
\begin{tabular}{l|ccc}
\hline
\rowcolor{Gray}
Method & \multicolumn{3}{c}{VOT22} \\
\rowcolor{Gray}
& Q & Acc & Rob \\
\hline
MS\_AOT \cite{aot}
& 0.673 & 0.781 & 0.944
\\

DiffusionTrack \cite{diffusiontrack}
& 0.634 & - & -
\\

DAMTMask \cite{vot2022}
& 0.624 & 0.796 & 0.891
\\

MixFormer-M \cite{mixformer}
& 0.589 & \textcolor{green}{\textbf{0.799}} & 0.878
\\

OSTrackSTS \cite{ostrack}
& 0.581 & 0.775 & 0.867
\\

Linker \cite{linker}
& 0.559 & 0.772 & 0.861
\\

SRATransTS \cite{vot2022}
& 0.547 & 0.743 & 0.866
\\

TransT\_M \cite{transt}
& 0.542 & 0.743 & 0.865
\\

GDFormer \cite{vot2022}
& 0.538 & 0.744 & 0.861
\\

TransLL \cite{vot2022}
& 0.530 & 0.735 & 0.861
\\

LWL\_B2S \cite{lwl}
& 0.516 & 0.736 & 0.831
\\

D3Sv2 \cite{d3sv2}
& 0.497 & 0.713 & 0.827
\\

\hline

\textit{\textbf{Zero-shot Method}} \\

SAM2-L \cite{sam2}
& 0.692 & 0.779 & 0.946 
\\
SAMURAI-L \cite{samurai}
& 0.693 & 0.744 & 0.951
\\
DAM4SAM-L \cite{dam4sam}
& \textcolor{green}{\textbf{0.750}} & \textcolor{blue}{\textbf{0.797}} & \textcolor{green}{\textbf{0.971}}
\\

\hline

SENTRY-S2-L 
& 0.701 \footnotesize{(\textcolor{gray}{\textbf{0.9$\uparrow$}})} & 0.787 \footnotesize{(\textcolor{gray}{\textbf{0.8$\uparrow$}})} & 0.953 \footnotesize{(\textcolor{gray}{\textbf{0.7$\uparrow$}})}
\\
SENTRY-SR-L 
& \textcolor{blue}{\textbf{0.715}} \footnotesize{(\textcolor{gray}{\textbf{2.1$\uparrow$}})} & 0.775 \footnotesize{(\textcolor{gray}{\textbf{3.1$\uparrow$}})} & \textcolor{blue}{\textbf{0.957}} \footnotesize{(\textcolor{gray}{\textbf{0.6$\uparrow$}})}
\\
SENTRY-D4S-L 
& \textcolor{red}{\textbf{0.759}} \footnotesize{(\textcolor{gray}{\textbf{0.9$\uparrow$}})} & \textcolor{red}{\textbf{0.805}} \footnotesize{(\textcolor{gray}{\textbf{0.8$\uparrow$}})} & \textcolor{red}{\textbf{0.979}} \footnotesize{(\textcolor{gray}{\textbf{0.8$\uparrow$}})}
\\

\hline
\end{tabular}
}
\vspace{-1em}
\end{minipage}
\hfill
\begin{minipage}[t]{0.49\columnwidth}
\centering
\caption{\small Comparison on the VOTS24 \cite{vots2024}.}
\vspace{-1em}
\label{tab:results_vot24}
\scalebox{0.48}{
\begin{tabular}{l|ccc}
\hline
\rowcolor{Gray}
Method & \multicolumn{3}{c}{VOTS24 \cite{vots2024}} \\
\rowcolor{Gray}
& Q & Acc & Rob \\

\hline

S3-Track \cite{s3}
& \textcolor{red}{\textbf{0.722}} & 0.784 & \textcolor{red}{\textbf{0.889}}
\\

DMAOT\_SAM \cite{dmaot}
& 0.653 & \textcolor{red}{\textbf{0.794}} & 0.780
\\

HQ-DMAOT \cite{dmaot}
& 0.639 & 0.754 & 0.790
\\

DMAOT \cite{dmaot}
& 0.636 & 0.751 & 0.795
\\

LY-SAM \cite{vots2024}
& 0.631 & 0.765 & 0.776
\\

Cutie-SAM \cite{cutie}
& 0.607 & 0.756 & 0.730
\\

AOT \cite{aot}
& 0.550 & 0.698 & 0.767
\\

LORAT \cite{lorat}
& 0.536 & 0.725 & 0.784
\\

\hline

\textit{\textbf{Zero-shot Method}} \\

SAM2-L \cite{sam2}
& 0.661 & \textcolor{blue}{\textbf{0.791}} & 0.790
\\
SAMURAI-L \cite{samurai}
& 0.673 & 0.776 & 0.851
\\
DAM4SAM-L \cite{dam4sam}
& \textcolor{blue}{\textbf{0.711}} & \textcolor{green}{\textbf{0.793}} & \textcolor{blue}{\textbf{0.864}}
\\

\hline

SENTRY-S2-L 
& 0.669 \footnotesize{(\textcolor{gray}{\textbf{0.8$\uparrow$}})} & \textcolor{green}{\textbf{0.793}} \footnotesize{(\textcolor{gray}{\textbf{0.2$\uparrow$}})} & 0.795 \footnotesize{(\textcolor{gray}{\textbf{0.5$\uparrow$}})}
\\
SENTRY-SR-L 
& \textcolor{blue}{\textbf{0.681}} \footnotesize{(\textcolor{gray}{\textbf{0.8$\uparrow$}})} & 0.782 \footnotesize{(\textcolor{gray}{\textbf{0.6$\uparrow$}})} & 0.856 \footnotesize{(\textcolor{gray}{\textbf{0.5$\uparrow$}}})
\\
SENTRY-D4S-L 
& \textcolor{green}{\textbf{0.715}} \footnotesize{(\textcolor{gray}{\textbf{0.4$\uparrow$}})} & \textcolor{red}{\textbf{0.794}} \footnotesize{(\textcolor{gray}{\textbf{0.1$\uparrow$}})} & \textcolor{green}{\textbf{0.867}} \footnotesize{(\textcolor{gray}{\textbf{0.3$\uparrow$}})}
\\

\hline

\end{tabular}
}
\end{minipage}
\hfill
\begin{minipage}[t]{0.49\columnwidth}
\centering
\vspace{1em}
\caption{\small Comparison on the DiDi \cite{dam4sam}.}
\vspace{-1em}
\label{tab:results_didi}
\scalebox{0.48}{
\begin{tabular}{l|ccc}
\hline
\rowcolor{Gray}
Method & \multicolumn{3}{c}{DiDi \cite{dam4sam}} \\
\rowcolor{Gray}

& Q & Acc & Rob \\

\hline

ODTrack \cite{odtrack}
& 0.608 & 0.740 & 0.809
\\

Cutie \cite{cutie}
& 0.575 & 0.704 & 0.776
\\

AOT \cite{aot}
& 0.541 & 0.622 & 0.852
\\

AQATrack \cite{aqatrack}
& 0.535 & 0.693 & 0.753
\\

SeqTrack \cite{seqtrack}
& 0.529 & 0.714 & 0.718
\\

KeepTrack \cite{keeptrack}
& 0.502 & 0.646 & 0.748
\\

TransT \cite{transt}
& 0.465 & 0.669 & 0.678
\\

\specialrule{1pt}{0pt}{0pt}

\textbf{\textit{Zero-shot Method}} \\

SAM2Long \cite{sam2long}
& 0.646 & 0.719 & 0.883
\\

SAM2-L \cite{sam2}
& 0.649 & 0.720 & 0.887
\\

SAMURAI-L \cite{samurai}
& 0.680 & 0.722 & 0.930
\\

DAM4SAM-L \cite{dam4sam}
& \textcolor{green}{\textbf{0.694}} & 0.727 & \textcolor{green}{\textbf{0.944}}
\\

\hline

SENTRY-S2-L 
& 0.660 \footnotesize{(\textcolor{gray}{\textbf{1.1$\uparrow$}})} & \textcolor{blue}{\textbf{0.729}} \footnotesize{(\textcolor{gray}{\textbf{0.9$\uparrow$}})} & 0.897 \footnotesize{(\textcolor{gray}{\textbf{1.0$\uparrow$}})}
\\
SENTRY-SR-L 
& \textcolor{blue}{\textbf{0.693}} \footnotesize{(\textcolor{gray}{\textbf{1.3$\uparrow$}})} & \textcolor{green}{\textbf{0.734}} \footnotesize{(\textcolor{gray}{\textbf{1.2$\uparrow$}})} & \textcolor{blue}{\textbf{0.943}} \footnotesize{(\textcolor{gray}{\textbf{1.3$\uparrow$}})}
\\
SENTRY-D4S-L 
& \textcolor{red}{\textbf{0.705}} \footnotesize{(\textcolor{gray}{\textbf{1.1$\uparrow$}})} & \textcolor{red}{\textbf{0.737}} \footnotesize{(\textcolor{gray}{\textbf{1.0$\uparrow$}})} & \textcolor{red}{\textbf{0.956}} \footnotesize{(\textcolor{gray}{\textbf{1.2$\uparrow$}})}
\\

\hline

\end{tabular}
}
\vspace{-1em}
\end{minipage}
\vspace{-1em}
\end{table}
%
\vspace{-1em}
\subsection{SOTA comparison on VOT benchmarks} \label{sec:sota_vot}
\begin{figure}[!t]
\centering
\begin{subfigure}[b]{0.49\linewidth}
\includegraphics[width=\linewidth]{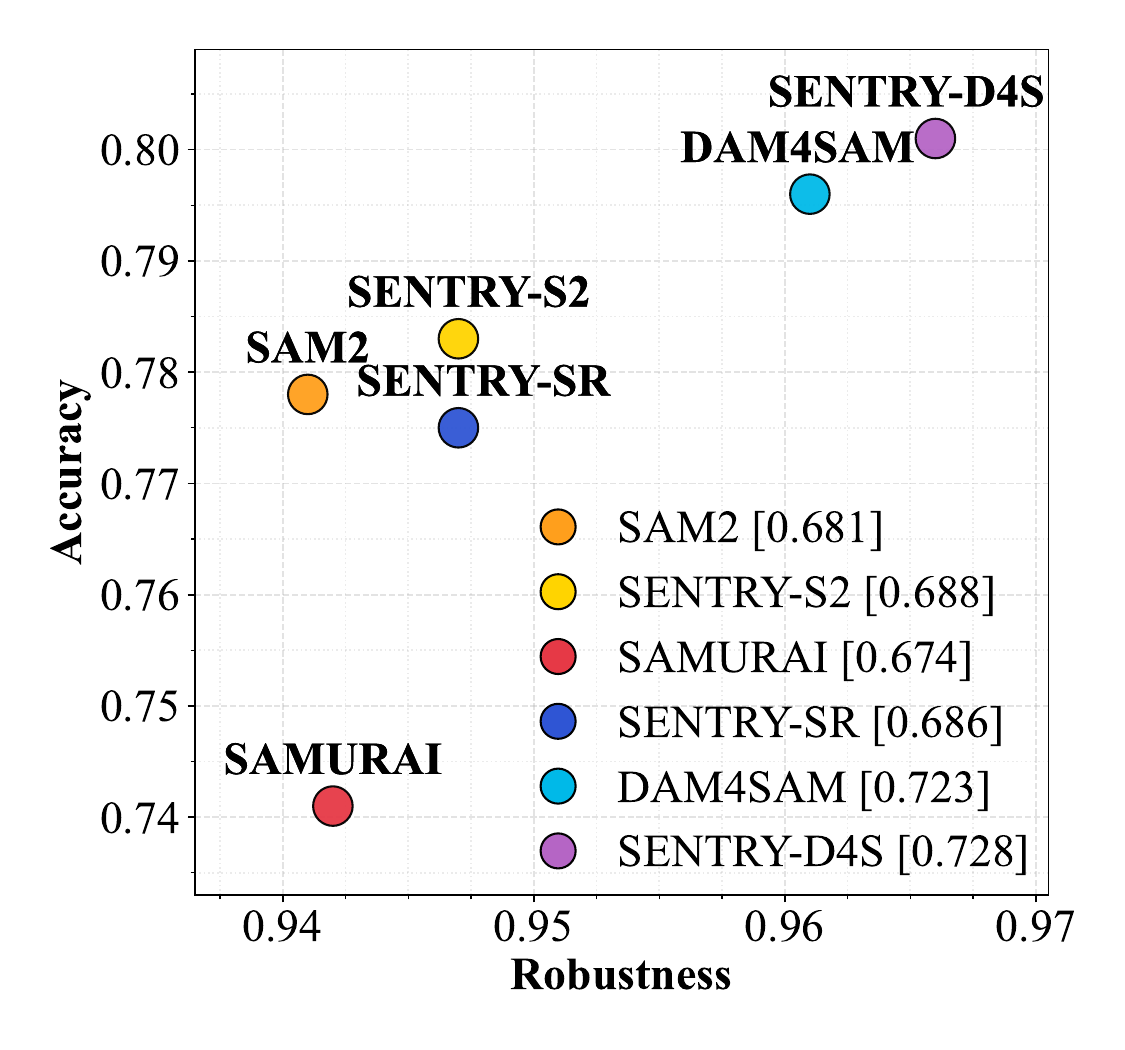}
\vspace{-2em}
\caption{}
\end{subfigure}
\begin{subfigure}[b]{0.49\linewidth}
\includegraphics[width=\linewidth]{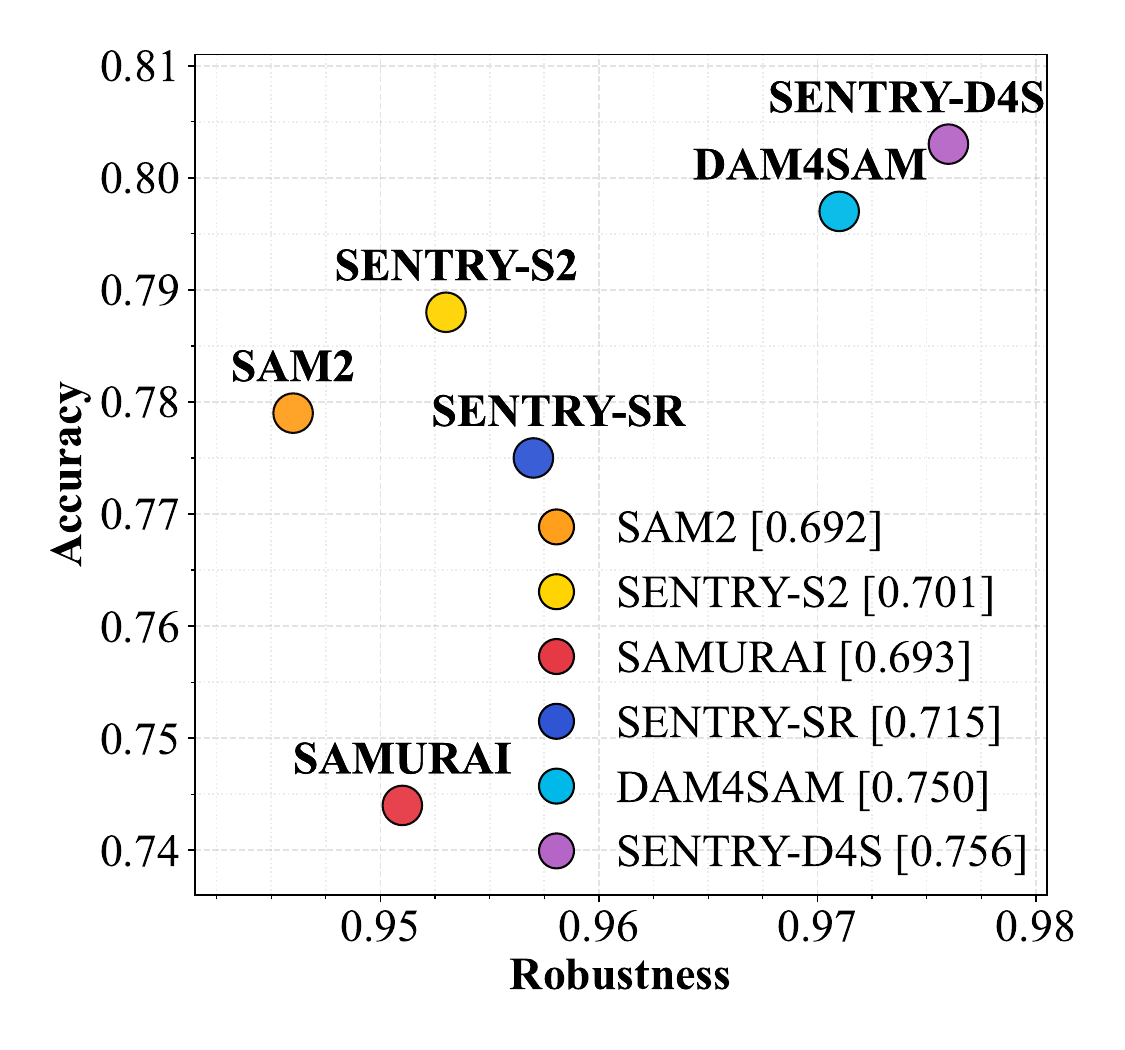}
\vspace{-2em}
\caption{}
\end{subfigure}
\begin{subfigure}[b]{0.49\linewidth}
\vspace{-0.8em}
\includegraphics[width=\linewidth]{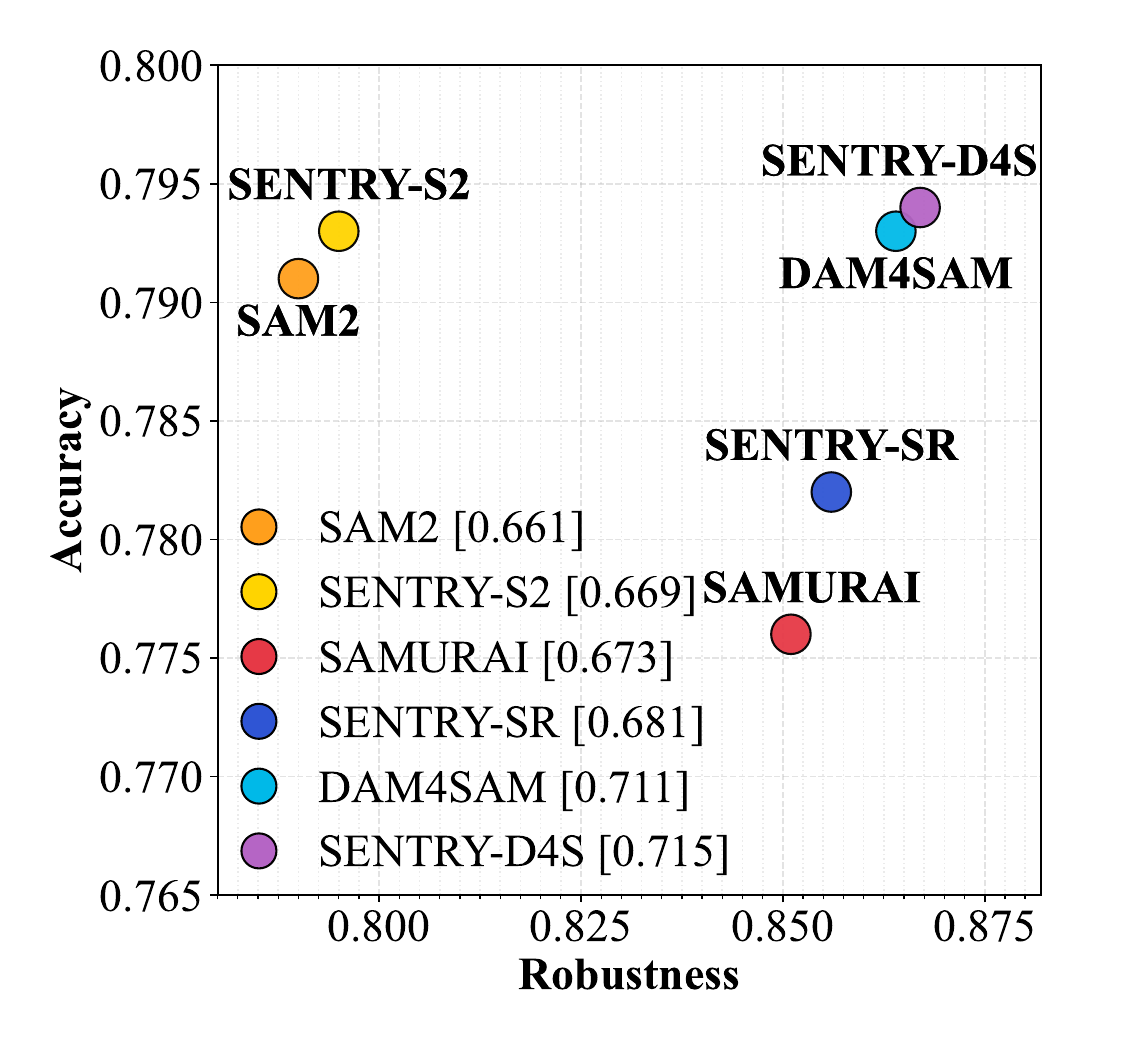}
\vspace{-2em}
\caption{}
\end{subfigure}
\begin{subfigure}[b]{0.49\linewidth}
\vspace{-0.8em}
\includegraphics[width=\linewidth]{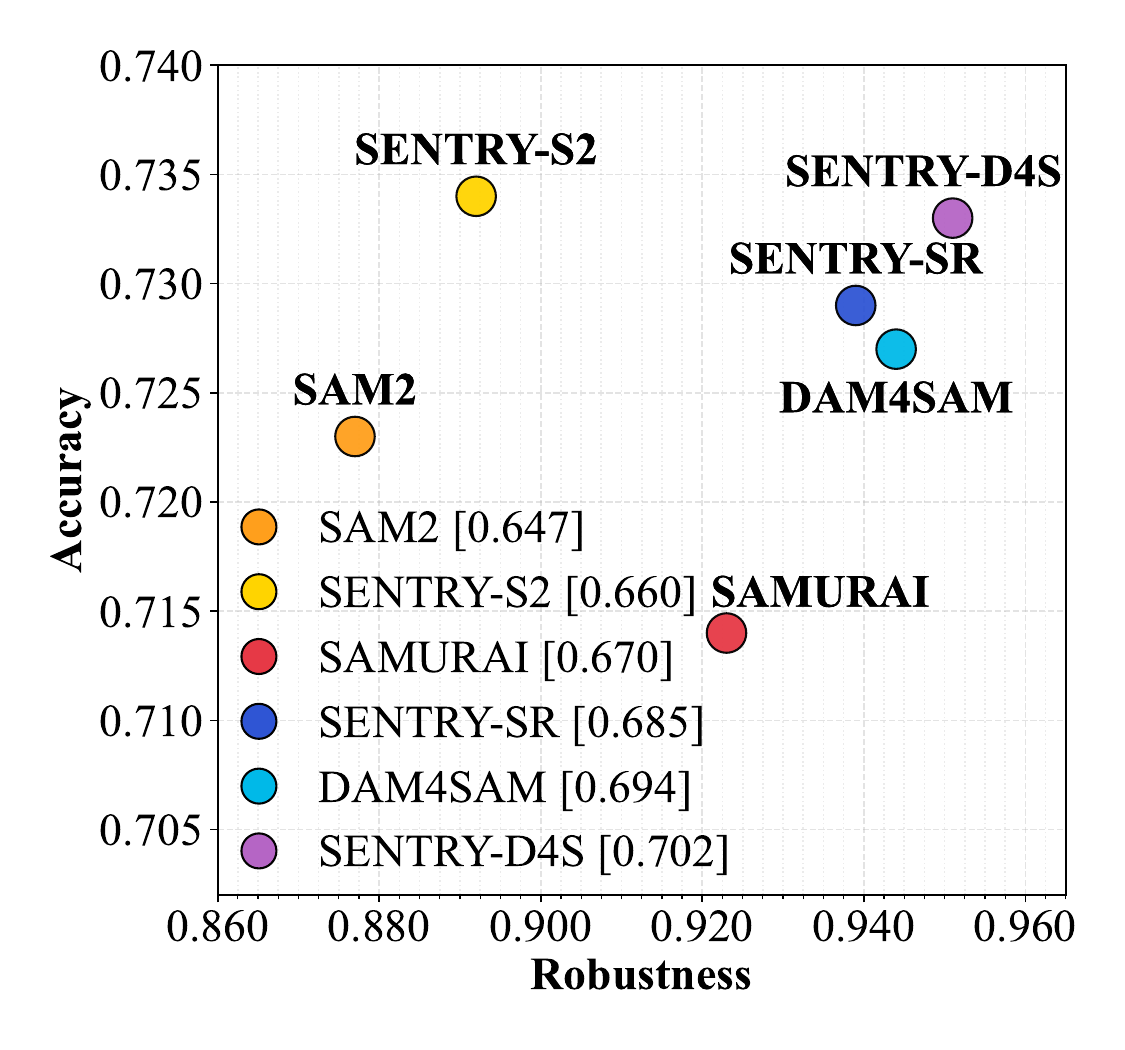}
\vspace{-2em}
\caption{}
\end{subfigure}
\vspace{-1em}
\caption{\textit{Acc}-\textit{Rob} plot on (a) VOT20 \cite{vot2020}, (b) VOT22 \cite{vot2022}, (c) VOTS24 \cite{vots2024}, and (d) DiDi \cite{dam4sam} for SAM2 and SENTRY variants. The \textit{Q} is given at each label.}
\label{fig:vot_results}
\vspace{-2em}
\end{figure}

\noindent We evaluate SENTRY on VOT20 \cite{vot2020}, VOT22 \cite{vot2022}, and VOTS24 \cite{vots2024}, with results in Tabs. \ref{tab:results_vot20}–\ref{tab:results_vot24} and Figs. \ref{fig:vot_results}(a)–(c).
We also assess DiDi \cite{dam4sam}, which focuses on visually similar non-negligible distractors, with results in Tab. \ref{tab:results_didi} and Fig. \ref{fig:vot_results}(d).
Additional results for model scales and baselines are in Appendices \ref{supplement: Additional quantitative results on VOT benchmarks} and \ref{supplement: Extended evaluations on additional SAM2-based frameworks}.
\noindent \textbf{VOT20 \cite{vot2020}.} 
All SENTRY variants outperform their SAM2-based baselines, with relative gains up to 1.2\% in \textit{Q}, 3.4\% in \textit{Acc}, and 0.7\% in \textit{Rob}. 
Most VOT20 failures stem from unstable masks during fast motion and occlusions, which trigger resets; SENTRY filters these unreliable updates, reducing reset events and improving all metrics.
SENTRY-D4S obtains the highest scores, setting a new SOTA.
\noindent \textbf{VOT22 \cite{vot2022}.} 
All SENTRY variants again improve over their baselines, with gains up to 2.1\% in \textit{Q}, 3.1\% in \textit{Acc}, and 0.8\% in \textit{Rob}. 
VOT22 includes stronger appearance changes and heavier distractor pressure than VOT20, increasing the risk of incorrect memory writes; SENTRY suppresses these errors and yields consistent improvements across all methods.
SENTRY-D4S again sets a new SOTA.
\noindent \textbf{VOTS24 \cite{vots2024}.} 
SENTRY variants surpass their baselines, with gains up to 0.8\% in \textit{Q}, 0.6\% in \textit{Acc}, and 0.5\% in \textit{Rob}. 
VOTS24’s long-term protocol exposes trackers to gradual drift and extended occlusion periods, where errors accumulate. SENTRY reduces these accumulation points and provides steady performance gains.
SENTRY-D4S attains the highest \textit{Acc} and ranks second in \textit{Q} and \textit{Rob}, closely trailing the competition-tuned S3-Track \cite{s3} while outperforming all zero-shot and several supervised competitors.
%
%
\textbf{DiDi \cite{dam4sam}.} 
SENTRY variants improve over their baselines, achieving gains up to 1.3\% in \textit{Q}, 1.2\% in \textit{Acc}, and 1.3\% in \textit{Rob}.
DiDi is explicitly designed to induce failures through dense distractors, where baseline trackers often commit erroneous memory updates. SENTRY mitigates these errors and stabilizes mask updates across all variants, with SENTRY-D4S achieving the highest overall scores and setting a new SOTA.

\vspace{-1em}
\subsection{Qualitative Results} \label{sec:qualitative}
\begin{figure*}[!h]
\centering
\includegraphics[width=0.99\linewidth]{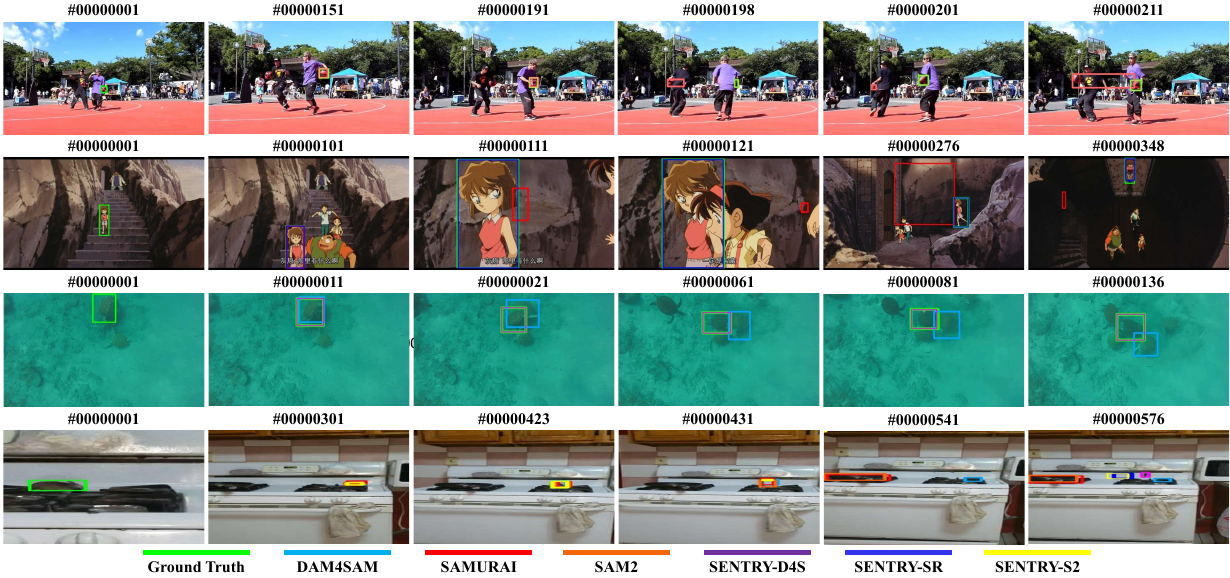}
\vspace{-1em}
\caption{Qualitative comparison of baselines and SENTRY variants: (1) SAM2 vs. SENTRY-S2, (2) SAMURAI vs. SENTRY-SR, (3) DAM4SAM vs. SENTRY-D4S. (4) Failure case: all methods lose the target after a long out-of-view period (100 frames) with a tiny target.}
\label{fig:vr_results}
\vspace{-2em}
\end{figure*}

\noindent \textbf{Tracking Result.} Fig. \ref{fig:vr_results} shows how SENTRY mitigates key failure modes of existing SAM2-based variants.
Row 1: SAM2 loses identity after heavy occlusion due to ambiguous mask selection, whereas SENTRY-S2 disambiguates and refines the correct mask before memory writing.
Row 2: under abrupt scene change, SAMURAI’s Kalman prior diverges, while SENTRY-SR restores identity through neighbor-guided matching.
Row 3: In an underwater scene with haze and distractors, DAM4SAM mis-selects background objects due to heuristic thresholds, while SENTRY-D4S chooses the candidate maintaining short-term trajectory coherence.
Row 4: after an extreme long-term occlusion (\(\approx \)100 frames), all methods fail to recover the tiny target, highlighting persistent re-identification limits.

\vspace{-1em}
\subsection{Ablation Study} \label{sec:ablation_study}
\noindent \textbf{Scalability across SAM2 sizes and base frameworks.} 
We integrate SENTRY across four SAM2 model scales (T/S/B/L). Tab. \ref{tab:abl_variants} summarizes average S improvements over five benchmarks, with full per-metric, per-scale results in Appendices \ref{supplement: Additional quantitative results on bounding box benchmarks} and \ref{supplement: Additional quantitative results on VOT benchmarks} (Tabs. \ref{tab:sam2_comparison} and \ref{tab:sam_comp_vot}). SENTRY improves all base frameworks and remains largely backbone-agnostic: SAM2 gains 0.6–0.8 across scales, indicating memory reliability rather than capacity is the main limitation; SAMURAI benefits more strongly (3.1–5.4 on average, with +8.8/+13.8/+11.8/+9.0 on TNL2K) due to SENTRY removing failures caused by its linear-motion update model. DAM4SAM gains 1.0–2.8 on average, reflecting improved stability in distractor-heavy settings. Notably, the S-scale variant is consistently the weakest among DAM4SAM models, indicating stronger dependence on feature capacity; SENTRY compensates by filtering low-confidence distractor-aligned updates before they enter memory.
\noindent \textbf{Candidate mask generation.} 
Tab. \ref{tab:abl_candidate_generation} compares candidate-generation strategies within SENTRY.
Detection–segmentation pipelines such as GroundingDINO \cite{groundingdino} with SAM/SAM2 \cite{sam,sam2} produce spatially precise but frame-independent masks, resulting in poor temporal consistency. Conversely, SAM2’s native propagation preserves identity but offers no mechanism for correcting drift once it accumulates.
Building on SAM2, we introduce four complementary modules evaluated under the same selection policy: decoder hypotheses ($\mathcal{C}^t_{\text{dec}}$) to retain multiple plausible masks, AMG proposals ($\mathcal{C}^t_{\text{AMG}}$) for expanded spatial coverage, Soft-NMS to reduce redundancy among overlapping hypotheses, and a Kalman prior ($\mathcal{M}^t_{\kappa}$) to stabilize motion.
As shown in Tab. \ref{tab:abl_candidate_generation}, each addition yields incremental S gains of 0.5, 0.1, 0.1, and 0.1, leading to a 0.9\% average improvement over SAM2 across benchmarks.
\noindent \textbf{Temporal consistency length \(\tau\).}
We analyze the effect of varying the temporal window $\tau$, which defines how many past frames contribute to matching. 
As shown in Tab. \ref{tab:abl_tau}, increasing $\tau$ improves robustness by using richer appearance and motion context, but overly long windows may introduce outdated features and temporal noise. Performance saturates around $\tau{=}10$.
\noindent \textbf{Runtime analysis.}
We evaluate runtime on an NVIDIA A100 GPU. 
SENTRY-S2-L achieves 32.8\,FPS compared to 44.0\,FPS for SAM2-L (Tab. \ref{tab:runtime}), adding only $\sim$7--8\,ms per frame ($\sim$25\% overhead) while retaining real-time performance.
Across SENTRY-compatible large-model hosts, the supplementary runtime and memory study reports a 16.6--25.5\% FPS drop and 0.4--0.6\,GB additional VRAM (App. \ref{supplement:runtime_memory_overhead}).
The overhead comes from three components: (i) candidate construction (AMG/decoder aggregation, template similarity, and Soft-NMS), (ii) short-horizon reasoning (backward re-segmentation and Hungarian matching), and (iii) Kalman prior.
Removing the Kalman prior or AMG/decoder inputs reduces candidate load; disabling Soft-NMS increases latency due to redundant proposals; and w/o neighbor pool restricts matching to the target trajectory only (no distractor modeling).
\noindent \textbf{Neighbor tracklet pool.}
Removing the neighbor tracklet pool $\mathbf{P}_n^{t}$ reduces matching to target-only consistency checks, ignoring distractor motion. This causes drift under clutter. Incorporating NP provides contextual negatives, penalizing candidates aligned with distractor trajectories and yielding temporal stability (Tab. \ref{tab:abl_neighbor}).
\begin{table*}[h]
\centering
\small
\vspace{-1.5em}
\begin{minipage}[t]{0.54\textwidth}
\centering
\caption{\small Incremental contribution of candidate-generation modules on \cite{lasot,lasotext,tnl2k,got10k}.  
$\Delta$ denotes average S gain over the SAM2-L.}
\vspace{-1em}
\label{tab:abl_candidate_generation}
\scalebox{0.65}{
\begin{tabular}{lccccc}
\hline
\rowcolor{Gray}
Method / Component & LaSOT & LaSOT\textsubscript{ext} & TNL2K & GOT-10k & $\Delta$ \\
\hline
SAM2 (Baseline) \cite{sam2} & 68.5 & 56.8 & 56.7 & 80.8 & - \\
\hline
\multicolumn{6}{l}{\textbf{Candidate Mask Generation Baselines}} \\
\begin{tabular}[c]{@{}c@{}}GroundingDINO \& \\ SAM \cite{groundingdino,sam}\end{tabular}
& 68.2 & 56.4 & 56.4 & 80.5 & 0.3$\downarrow$ \\
\begin{tabular}[c]{@{}c@{}}GroundingDINO \& \\ SAM2 \cite{groundingdino,sam2}\end{tabular}
& 69.0 & 56.7 & 56.6 & 80.9 & 0.1$\uparrow$ \\
SAM~\cite{sam} & 69.0 & 56.8 & 57.0 & 80.9 & 0.2$\uparrow$ \\
\hline
\multicolumn{6}{l}{\textbf{Incremental Modules on SAM2}} \\
+ $\mathcal{C}_{\text{dec}}$ & 69.8 & 57.0 & 57.5 & 80.9 & 0.6$\uparrow$ \\
+ $\mathcal{C}_{\text{AMG}}$ & 69.9 & 57.0 & 57.8 & 80.9 & 0.7$\uparrow$ \\
+ Soft-NMS & 70.0 & 57.0 & 57.8 & 81.0 & 0.8$\uparrow$ \\
\rowcolor{blue!7.5}
+ $\mathcal{M}^{\kappa}$ & \textbf{70.2} & \textbf{57.0} & \textbf{57.9} & \textbf{81.1} & \textbf{0.9$\uparrow$} \\
\hline
\end{tabular}
}
\end{minipage}
\hfill
\begin{minipage}[t]{0.44\textwidth}
\centering

\begin{minipage}[t]{0.54\linewidth}
\centering
\caption{\small Effect of $\tau$ on performance.}
\vspace{-1em}
\label{tab:abl_tau}
\scalebox{0.65}{
\begin{tabular}{cccc}
\hline
\rowcolor{Gray}
$\tau$ & LaSOT & LaSOT\textsubscript{ext} & TNL2K \\
\hline
1  & 68.5 & 56.8 & 56.7 \\
3  & 69.1 & 56.8 & 57.1 \\
5  & 69.9 & 56.9 & 57.5 \\
\rowcolor{blue!7.5}
10 & \textbf{70.2} & \textbf{57.0} & \textbf{57.9} \\
20 & 70.0 & 56.9 & 57.7 \\
30 & 69.2 & 56.8 & 57.7 \\
\hline
\end{tabular}
}
\end{minipage}
\hfill
\begin{minipage}[t]{0.42\linewidth}
\centering
\caption{\small{Effect of $\mathbf{P}_n^{t}$ on \cite{lasot}.}}
\vspace{-1.15em}
\label{tab:abl_neighbor}
\scalebox{0.6}{
\begin{tabular}{l|c}
\hline
\rowcolor{Gray} \multicolumn{2}{c}{\textbf{w/o $\mathbf{P}_n^{t}$}} \\
\hline
S & 65.8 \\
NP & 73.1 \\
P & 70.4 \\
\hline
\rowcolor{Gray} \multicolumn{2}{c}{\textbf{with $\mathbf{P}_n^{t}$}} \\
\hline
S & \textbf{70.2} \\
NP & \textbf{77.2} \\
P & \textbf{74.5} \\
\hline
\end{tabular}
}
\end{minipage}
\caption{\small{Runtime analysis.}}
\vspace{-1em}
\scriptsize
\label{tab:runtime}
\scalebox{0.7}{
\begin{tabular}{l@{\hskip 6pt}cc}
\hline
\rowcolor{Gray} Variant & FPS $\uparrow$ & Overhead (\%) $\downarrow$ \\
\hline
SAM2-L & 44.0 & -- \\
SENTRY-S2-L & 32.8 & 25.5 \\
\quad w/o Kalman prior & 34.7 & 21.1 \\
\quad w/o AMG proposals & 35.4 & 19.5 \\
\quad w/o decoder hypotheses & 36.9 & 16.1 \\
\quad w/o neighbor pool & 38.7 & 12.0 \\
\quad w/o candidate filtration & 30.1 & 31.6 \\
\hline
\end{tabular}
}
\end{minipage}
\vspace{-1.5em}
\end{table*}
%
\noindent \textbf{Scalability.}
Across all scales (T/S/B/L) and all three base trackers, SENTRY consistently improves robustness (Tab. \ref{tab:abl_variants}). Gains hold on both short- and long-term benchmarks, despite differences in dataset difficulty and model capacity. SAM2 benefits most at small and large scales; SAMURAI sees the strongest overall lift, especially on TNL2K and TrackingNet; and DAM4SAM shows targeted gains, with S-scale improving notably on GOT-10k and TrackingNet. Fig. \ref{fig:overall_scale} corroborates this.
\begin{table*}[!h]
\vspace{-1em}
\centering
\setlength{\tabcolsep}{1pt}
\caption{
Average relative improvements ($\Delta$) of SENTRY over each base framework across model sizes (T/S/B/L).
Mean \(\Delta\) across all bases: SAM2 (0.7), SAMURAI (4.5), DAM4SAM (1.5).
Full per-metric results are provided in App. \ref{supplement: Additional quantitative results on bounding box benchmarks}, Tab. \ref{tab:sam2_comparison}.
}
\vspace{-14pt}
\label{tab:abl_variants}
\begin{subtable}[t]{0.325\textwidth}
\centering
\caption{SAM2 $\to$ SENTRY}
\vspace{-3pt}
\resizebox{\textwidth}{!}{
\begin{tabular}{lcccc}
\specialrule{1pt}{0pt}{0pt}
\rowcolor{Gray} Dataset & T & S & B & L \\
\specialrule{1pt}{0pt}{0pt}
LaSOT & 0.6\(\uparrow\) & 1.5\(\uparrow\) & 1.0\(\uparrow\) & \textbf{1.7\(\uparrow\)} \\
LaSOT\textsubscript{ext} & 0.7\(\uparrow\) & \textbf{1.0\(\uparrow\)} & 0.7\(\uparrow\) & 0.2\(\uparrow\) \\
TNL2K & 1.2\(\uparrow\) & 1.2\(\uparrow\) & \textbf{1.3\(\uparrow\)} & \textbf{1.3\(\uparrow\)} \\
GOT-10k & \textbf{0.3\(\uparrow\)} & 0.1\(\uparrow\) & 0.1\(\uparrow\) & \textbf{0.3\(\uparrow\)} \\
TrackingNet & 0.3\(\uparrow\) & \textbf{0.4\(\uparrow\)} & \textbf{0.4\(\uparrow\)} & \textbf{0.4\(\uparrow\)} \\
\specialrule{1pt}{0pt}{0pt}
\rowcolor{blue!7.5} \(\Delta\) & 0.6\(\uparrow\) & \textbf{0.8\(\uparrow\)} & 0.7\(\uparrow\) & \textbf{0.8\(\uparrow\)} \\
\specialrule{1pt}{0pt}{0pt}
\end{tabular}
}
\end{subtable}
\hfill
\begin{subtable}[t]{0.325\textwidth}
\centering
\caption{SAMURAI $\to$ SENTRY}
\vspace{-3pt}
\resizebox{\textwidth}{!}{
\begin{tabular}{lcccc}
\specialrule{1pt}{0pt}{0pt}
\rowcolor{Gray} Dataset & T & S & B & L \\
\specialrule{1pt}{0pt}{0pt}
LaSOT & \textbf{2.6\(\uparrow\)} & 1.4\(\uparrow\) & 2.5\(\uparrow\) & 0.9\(\uparrow\) \\
LaSOT\textsubscript{ext} & 2.0\(\uparrow\) & \textbf{2.4\(\uparrow\)} & \textbf{2.4\(\uparrow\)} & 2.0\(\uparrow\) \\
TNL2K & 8.8\(\uparrow\) & \textbf{13.8\(\uparrow\)} & 11.8\(\uparrow\) & 9.0\(\uparrow\) \\
GOT-10k & 2.1\(\uparrow\) & \textbf{2.4\(\uparrow\)} & 2.1\(\uparrow\) & 0.2\(\uparrow\) \\
TrackingNet & 6.3\(\uparrow\) & \textbf{7.0\(\uparrow\)} & 5.7\(\uparrow\) & 3.4\(\uparrow\) \\
\specialrule{1pt}{0pt}{0pt}
\rowcolor{blue!7.5} \(\Delta\) & 4.4\(\uparrow\) & \textbf{5.4\(\uparrow\)} & 4.9\(\uparrow\) & 3.1\(\uparrow\) \\
\specialrule{1pt}{0pt}{0pt}
\end{tabular}
}
\end{subtable}
\hfill
\begin{subtable}[t]{0.325\textwidth}
\centering
\caption{DAM4SAM $\to$ SENTRY}
\vspace{-3pt}
\resizebox{\textwidth}{!}{
\begin{tabular}{lcccc}
\specialrule{1pt}{0pt}{0pt}
\rowcolor{Gray} Dataset & T & S & B & L \\
\specialrule{1pt}{0pt}{0pt}
LaSOT & 1.1\(\uparrow\) & 1.1\(\uparrow\) & \textbf{2.0\(\uparrow\)} & 1.2\(\uparrow\) \\
LaSOT\textsubscript{ext} & 1.0\(\uparrow\) & 1.5\(\uparrow\) & \textbf{2.4\(\uparrow\)} & 1.1\(\uparrow\) \\
TNL2K & \textbf{1.7\(\uparrow\)} & 1.4\(\uparrow\) & 1.5\(\uparrow\) & 1.5\(\uparrow\) \\
GOT-10k & 0.7\(\uparrow\) & \textbf{5.3\(\uparrow\)} & 0.3\(\uparrow\) & 0.8\(\uparrow\) \\
TrackingNet & 0.3\(\uparrow\) & \textbf{4.9\(\uparrow\)} & 0.1\(\uparrow\) & 0.6\(\uparrow\) \\
\specialrule{1pt}{0pt}{0pt}
\rowcolor{blue!7.5} \(\Delta\) & 1.0\(\uparrow\) & \textbf{2.8\(\uparrow\)} & 1.3\(\uparrow\) & 1.0\(\uparrow\) \\
\specialrule{1pt}{0pt}{0pt}
\end{tabular}
}
\end{subtable}
\end{table*}

\vspace{-3em}
\section{Conclusion}
\noindent We revisited memory update mechanisms in SAM2-based VOT and exposed confidence-only mask selection as a key source of drift under occlusion, motion, and distractors. We introduced SENTRY, a training-free, plug-and-play, refine-before-write module that validates temporal consistency before committing memory updates. By replacing confidence-driven writes with consistency-validated ones, SENTRY stabilizes tracking without retraining or architectural changes and consistently improves all-scale evaluation performance across baselines and benchmarks.


\section*{Acknowledgements}
{\sloppy
This work was supported in part by the Khalifa University Center for Autonomous
Robotic Systems (KUCARS) under Award RC1-\allowbreak{}2018-\allowbreak{}KUCARS;
in part by Silal Innovation Oasis through projects under grants 8475000023,
8475000024, 8475000025, and 8475000026; and in part by Khalifa University of
Science and Technology through the Faculty Start-Ups under Project ID
KU-\allowbreak{}INT-\allowbreak{}FSU-\allowbreak{}2005-\allowbreak{}8474000775.
\par}


\bibliographystyle{splncs04}
\bibliography{main}




\clearpage\clearpage

\appendix

\renewcommand{\theHsection}{appendix.\Alph{section}}
\renewcommand{\theHsubsection}{appendix.\Alph{section}.\arabic{subsection}}






    
    






\section*{Appendix}
\vspace{-0.4cm}
\startcontents[appendices]
\providecommand{\authcount}[1]{} 
\printcontents[appendices]{}{1}{\setcounter{tocdepth}{2}}
\vspace{0.2cm}

\section{Additional Ablation Study} \label{supplement: Additional Ablation Study}

\noindent \textbf{Sensitivity to $\alpha$ and $\beta$.} 
We analyze the two thresholds used in candidate generation: the decoder margin $\alpha$, which controls the retention of SAM2 decoder hypotheses, and the AMG similarity gate $\beta$, which filters target-consistent proposals (Sec. \ref{sec:method_candidate_gen}). All other settings are fixed.
Table \ref{tab:abl_alpha} shows that $\alpha$ regulates decoder diversity. Low values admit weak hypotheses, while high values over-prune viable alternatives. Performance peaks at $\alpha=0.7$. Table \ref{tab:abl_beta} shows that $\beta$ balances template agreement and robustness to appearance change: $\beta$ in the range $[0.6,0.8]$ provides stable performance, with a maximum at $\beta=0.7$.
We therefore adopt $\alpha=0.7$ and $\beta=0.7$ as defaults, which offer the best trade-off between performance and stability on LaSOT.

\begin{table}[h!]
\centering
\caption{Ablation on decoder relative-confidence margin $\alpha$ on LaSOT (SAM2-L baseline).
We fix $\beta{=}0.7$ for AMG filtering.}
\vspace{-0.6em}
\setlength{\tabcolsep}{10pt}
\scalebox{0.85}{
\begin{tabular}{c|ccc}
\hline
\rowcolor{Gray}
$\alpha$ & S & NP & P \\
\hline
0.5 & 69.4 & 76.8 & 74.1 \\
0.6 & 69.9 & 76.9 & 74.2 \\
\rowcolor{blue!7.5} 0.7 & \textbf{70.2} & \textbf{77.2} & \textbf{74.5} \\
0.8 & 69.8 & 77.0 & 74.3 \\
0.9 & 69.3 & 76.7 & 74.2 \\
\hline
\end{tabular}}
\label{tab:abl_alpha}
\vspace{-0.5em}
\end{table}

\begin{table}[h!]
\centering
\caption{Ablation on AMG similarity margin $\beta$ on LaSOT (SAM2-L baseline).
We fix $\alpha{=}0.7$ for decoder filtering.}
\vspace{-0.6em}
\setlength{\tabcolsep}{10pt}
\scalebox{0.85}{
\begin{tabular}{c|ccc}
\hline
\rowcolor{Gray}
$\beta$ & S & NP & P \\
\hline
0.5 & 69.7 & 76.8 & 74.0 \\
0.6 & 69.9 & 77.0 & 74.1 \\
\rowcolor{blue!7.5} 0.7 & \textbf{70.2} & \textbf{77.2} & \textbf{74.5} \\
0.8 & 70.0 & 77.0 & 74.2 \\
0.9 & 69.6 & 76.9 & 74.1 \\
\hline
\end{tabular}}
\label{tab:abl_beta}
\vspace{-0.5em}
\end{table}

\noindent \textbf{Ablation of $\theta_{\mathrm{rel}}$.} 
The reliability threshold $\theta_{\mathrm{rel}}$ determines how strict the cycle-consistency check must be before a candidate is accepted. Values that are too low allow noisy backward trajectories; values that are too high reject valid matches during occlusion or partial drift. As shown in Table \ref{tab:abl_theta_rel}, performance peaks at $\theta_{\mathrm{rel}}=0.6$, which balances permissiveness and stability.

\begin{table}[h!]
\centering
\caption{Ablation on $\theta_{\mathrm{rel}}$ on LaSOT (SAM2-L). 
$\theta_{\mathrm{min}}$ fixed at $0.3$.}
\vspace{-0.5em}
\scalebox{0.87}{
\begin{tabular}{c|ccc}
\hline
\rowcolor{Gray}
$\theta_{\mathrm{rel}}$ & S & NP & P \\
\hline
0.4 & 69.3 & 76.9 & 74.2 \\
0.5 & 69.8 & 77.0 & 74.2 \\
\rowcolor{blue!7}
0.6 & \textbf{70.2} & \textbf{77.2} & \textbf{74.5} \\
0.7 & 69.9 & 77.0 & 74.2 \\
0.8 & 69.2 & 76.8 & 74.1 \\
\hline
\end{tabular}}
\label{tab:abl_theta_rel}
\vspace{-0.8em}
\end{table}

\noindent\textbf{Ablation of $\theta_{\mathrm{min}}$.}
The fallback threshold $\theta_{\mathrm{min}}$ controls how aggressively SENTRY accepts a candidate when no reliable cycle-consistent match is found. Table \ref{tab:abl_theta_min} shows a concave trend: overly low values accept weak tracklets, while high values eliminate necessary fallbacks. The best setting is $\theta_{\mathrm{min}}=0.3$.

\begin{table}[h!]
\centering
\caption{Ablation on $\theta_{\mathrm{min}}$ on LaSOT (SAM2-L). $\theta_{\mathrm{rel}}$ fixed at $0.6$.}
\vspace{-0.6em}
\scalebox{0.87}{
\begin{tabular}{c|ccc}
\hline
\rowcolor{Gray}
$\theta_{\mathrm{min}}$ & S & NP & P \\
\hline
0.1 & 69.6 & 76.8 & 74.0 \\
0.2 & 69.9 & 76.9 & 74.1 \\
\rowcolor{blue!7}
0.3 & \textbf{70.2} & \textbf{77.2} & \textbf{74.5} \\
0.4 & 69.9 & 77.0 & 74.2 \\
0.5 & 69.4 & 76.9 & 74.1 \\
\hline
\end{tabular}}
\label{tab:abl_theta_min}
\vspace{-0.8em}
\end{table}

\begin{figure}[h!]
\centering
\includegraphics[width=0.6\linewidth]{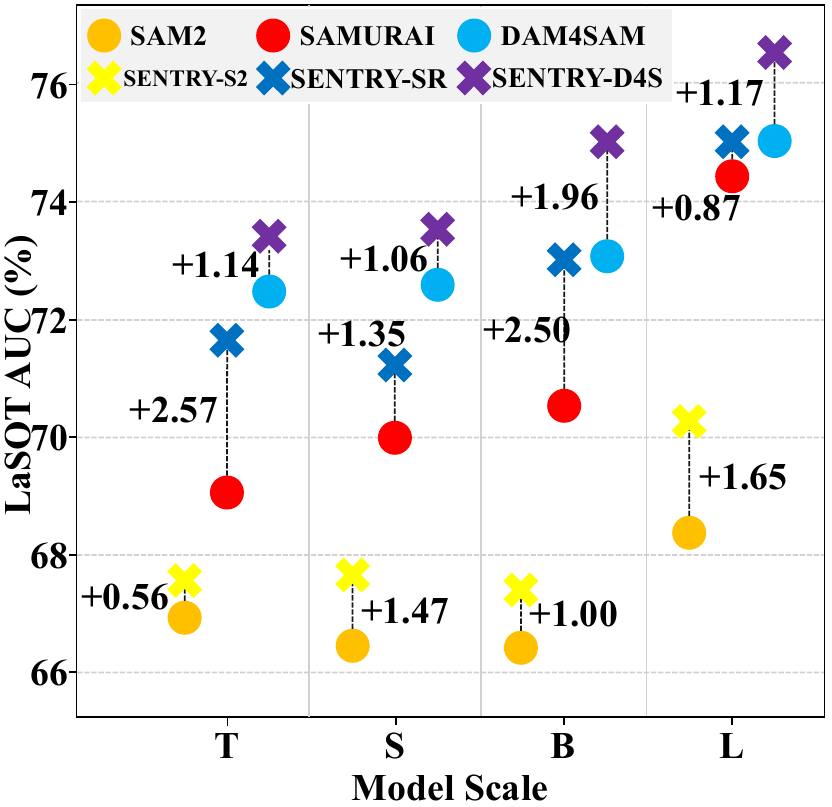}
\vspace{-0.5em}
\caption{Our proposed SENTRY performs favorably against SAM2-based variants across different model scales on LaSOT \cite{lasot} benchmark.}
\label{fig:overall_scale}
\vspace{-1em}
\end{figure}

\noindent \textbf{Scalability.}
Across all model scales (T/S/B/L) and all three base trackers, SENTRY produces a consistent upward shift in robustness, as summarized in Table \ref{tab:abl_variants}. Improvements persist across both short-term and long-term benchmarks, cutting through differences in dataset difficulty and model capacity. SAM2 shows its largest gains in the small and large variants, SAMURAI experiences the strongest overall lift, most prominently on TNL2K and TrackingNet, and DAM4SAM receives targeted but meaningful boosts, with the S-scale variant improving notably on GOT-10k and TrackingNet. These patterns reflect generalization rather than architecture-specific tuning. Fig. \ref{fig:overall_scale} reinforces this trend: across all scales, SENTRY-enhanced models lie consistently above their baselines, with the widest margin appearing on SAMURAI and clear, steady gains even for SAM2 and DAM4SAM. Regardless of the starting tracker, SENTRY reliably nudges performance upward.

\section{Memory-Write Diagnostic Analysis} \label{supplement:memory_write_diagnostics}
\noindent To directly test whether SENTRY reduces false-positive memory contamination, we analyze annotation-available testing and validation splits at the fixed memory-update times used by each host tracker.
For each written mask $W_t$, we compute $s_t=\mathcal{J}(W_t,G_t)$ when mask annotations are available, and $s_t=\operatorname{IoU}(B(W_t),B_t)$ for box-only annotations, where $B(W_t)$ is the tight box around the written mask.
We define false writes as $s_t<0.5$ and severe writes as $s_t<0.2$.
The written FP/proxy area measures the admitted pixels outside the target mask or, for box-only datasets, outside the target box.
Table~\ref{tab:mechanism_summary} reports macro-averages over the host baselines used in the paper.
SENTRY consistently reduces incorrect memory writes and written false-positive contamination.
To connect memory quality to final tracking quality, we also decompose final predictions into overlap $J/\operatorname{IoU}$, false-positive and false-negative errors $E^{\mathrm{FP}}$ and $E^{\mathrm{FN}}$, normalized center/scale errors, and boundary F-score $F_{\mathrm{B}}$.
Because $1-J=E^{\mathrm{FP}}+E^{\mathrm{FN}}$, this decomposition separates gains from suppressing extra predicted regions from gains due to recovering missed target regions.
The final-output decomposition indicates that the overlap gain is mainly associated with lower $E^{\mathrm{FP}}$, supporting the interpretation that SENTRY improves tracking by admitting cleaner memory states rather than only improving per-frame localization.

\begin{table}[t]
\centering
\caption{Write diagnostics and final-output decomposition. False and severe memory writes are reported for tracking/VOS splits; written FP/proxy area measures admitted false-positive area under mask or box annotations.}
\label{tab:mechanism_summary}
\vspace{-0.4em}
\setlength{\tabcolsep}{5pt}
\renewcommand{\arraystretch}{0.95}
\resizebox{\linewidth}{!}{%
\begin{tabular}{lccc}
\hline
\rowcolor{Gray}
{\small \textbf{Metric}} 
& {\small \textbf{Host avg.}} 
& {\small \textbf{+SENTRY avg.}} 
& {\small \textbf{Change}} \\
\hline
{\small False memory writes, Track/VOS $\downarrow$} 
& {\small 16.8 / 16.2} 
& {\small 10.5 / 10.1} 
& {\small $-37.5$ / $-37.7$ rel.} \\
{\small Severe memory writes, Track/VOS $\downarrow$} 
& {\small 5.1 / 4.8} 
& {\small 2.5 / 2.3} 
& {\small $-51.0$ / $-52.1$ rel.} \\
{\small Written FP/proxy area, Track/VOS $\downarrow$} 
& {\small 21.0 / 20.4} 
& {\small 14.5 / 14.1} 
& {\small $-31.0$ / $-30.9$ rel.} \\
\hline
{\small Final $J/\operatorname{IoU}\uparrow$} 
& {\small 72.0} 
& {\small 73.6} 
& {\small +1.6} \\
{\small Final $E^{\mathrm{FP}}/E^{\mathrm{FN}}\downarrow$} 
& {\small 16.7 / 11.3} 
& {\small 15.1 / 11.3} 
& {\small $-9.6$ / $0.0$ rel.} \\
{\small Final $E^{\mathrm{ctr}}/E^{\mathrm{scale}}\downarrow$} 
& {\small 0.092 / 0.147} 
& {\small 0.085 / 0.137} 
& {\small $-7.6$ / $-6.8$ rel.} \\
{\small Final $F_{\mathrm{B}}\uparrow$} 
& {\small 77.2} 
& {\small 78.3} 
& {\small +1.1} \\
\hline
\end{tabular}%
}
\vspace{-0.5em}
\end{table}

\section{Recovery-After-Failure Diagnostic Analysis} \label{supplement:recovery_after_failure}
\noindent SENTRY reduces drift through conservative memory admission, but it is not a global re-detector.
After extremely long target absence and tiny reappearance, all methods may fail because no reliable candidate remains in the decoder, AMG, or motion-prior pool.
However, SENTRY preserves the host update schedule and changes only the admitted mask; therefore, recovery remains possible whenever at least one plausible candidate is generated by the host decoder, the AMG proposal set, or the motion prior.
To quantify this bounded recovery behavior, we evaluate recovery after a failure state.
Given overlap $O_t<0.2$, where $O_t$ denotes mask overlap or box IoU depending on the available annotation, $\operatorname{Rec}@K$ measures whether the tracker returns to $O_{t'}\geq0.5$ within $K$ frames.
Delay measures the number of frames until recovery among recovered cases; negative $\Delta$Delay indicates faster recovery.
Table~\ref{tab:recovery_after_failure} reports changes relative to the corresponding host baseline.
SENTRY improves Recovery@10 and Recovery@20 and reduces recovery delay on all three hosts, supporting the bounded claim that conservative admission does not freeze memory, but helps recovery when plausible candidates exist.

\begin{table}[t]
\centering
\caption{Recovery-after-failure diagnostic. Recovery changes are in percentage points relative to each host baseline; delay change is in frames, where negative values indicate faster recovery.}
\label{tab:recovery_after_failure}
\vspace{-0.4em}
\setlength{\tabcolsep}{12pt}
\renewcommand{\arraystretch}{0.95}
\resizebox{\linewidth}{!}{%
\begin{tabular}{lccc}
\hline
\rowcolor{Gray}
{\small \textbf{Host}} 
& {\small \textbf{$\Delta$ Rec.@10} $\uparrow$} 
& {\small \textbf{$\Delta$ Rec.@20} $\uparrow$} 
& {\small \textbf{$\Delta$ Delay} $\downarrow$} \\
\hline
{\small SAM2-L}     & {\small +2.8} & {\small +2.7} & {\small $-0.7$} \\
{\small SAMURAI-L}  & {\small +8.3} & {\small +7.6} & {\small $-2.7$} \\
{\small DAM4SAM-L}  & {\small +2.7} & {\small +2.6} & {\small $-0.8$} \\
\hline
{\small \textbf{Average}} & {\small \textbf{+4.6}} & {\small \textbf{+4.3}} & {\small \textbf{$-1.4$}} \\
\hline
\end{tabular}%
}
\vspace{-0.5em}
\end{table}


\section{Generality Beyond SAM2-Based Trackers} \label{supplement:generality_beyond_sam2}
\noindent Although the main evaluation focuses on SAM2-based trackers, SENTRY only requires a mask or probability output and access to the mask passed into the memory write.
This allows the same refine-before-write admission principle to be applied to other memory-based trackers without changing their training procedure.
Table~\ref{tab:generality_memory_admission} reports representative results on non-SAM memory trackers and additional SAM-family trackers.
SENTRY consistently improves all five trackers across all five datasets, suggesting that the gains come from temporally verified memory writes rather than SAM2-specific internals.

\begin{table*}[t]
\centering
\caption{Generality of SENTRY-style memory admission beyond the main SAM2-based evaluation. The same refine-before-write principle improves both non-SAM memory trackers and additional SAM-family trackers across five datasets.}
\label{tab:generality_memory_admission}
\vspace{-0.4em}
\setlength{\tabcolsep}{5pt}
\renewcommand{\arraystretch}{0.95}
\resizebox{\textwidth}{!}{%
\begin{tabular}{lc|ccccc}
\hline
\rowcolor{Gray} & {\small \textbf{Method}} & {\small \textbf{LaSOT}} & {\small \textbf{LaSOT\textsubscript{ext}}} & {\small \textbf{TNL2K}} & {\small \textbf{TrackingNet}} & {\small \textbf{GOT-10k}}\\
\hline
\parbox[t]{0.05cm}{\multirow{6}{*}{\rotatebox[origin=c]{90}{{\small\textbf{Memory}}}}}
& STCN \cite{stcn} & 58.0 & 45.0 & 46.0 & 75.5 & 70.0\\
& +\textsc{SENTRY} & 58.7\scriptsize{(\textcolor{gray}{\textbf{0.7$\uparrow$}})} & 45.6\scriptsize{(\textcolor{gray}{\textbf{0.6$\uparrow$}})} & 46.8\scriptsize{(\textcolor{gray}{\textbf{0.8$\uparrow$}})} & 76.0\scriptsize{(\textcolor{gray}{\textbf{0.5$\uparrow$}})} & 70.5\scriptsize{(\textcolor{gray}{\textbf{0.5$\uparrow$}})}\\
\cline{2-7}
& XMem \cite{xmem} & 60.5 & 48.0 & 49.0 & 78.0 & 72.0\\
& +\textsc{SENTRY} & 61.4\scriptsize{(\textcolor{gray}{\textbf{0.9$\uparrow$}})} & 48.7\scriptsize{(\textcolor{gray}{\textbf{0.7$\uparrow$}})} & 49.9\scriptsize{(\textcolor{gray}{\textbf{0.9$\uparrow$}})} & 78.6\scriptsize{(\textcolor{gray}{\textbf{0.6$\uparrow$}})} & 72.8\scriptsize{(\textcolor{gray}{\textbf{0.8$\uparrow$}})}\\
\cline{2-7}
& Cutie \cite{cutie} & 63.3 & 51.5 & 52.0 & 80.5 & 74.0\\
& +\textsc{SENTRY} & 64.2\scriptsize{(\textcolor{gray}{\textbf{0.9$\uparrow$}})} & 52.2\scriptsize{(\textcolor{gray}{\textbf{0.7$\uparrow$}})} & 53.5\scriptsize{(\textcolor{gray}{\textbf{1.5$\uparrow$}})} & 80.9\scriptsize{(\textcolor{gray}{\textbf{0.4$\uparrow$}})} & 74.6\scriptsize{(\textcolor{gray}{\textbf{0.6$\uparrow$}})}\\
\hline
\parbox[t]{0.05cm}{\multirow{4}{*}{\rotatebox[origin=c]{90}{{\small\textbf{SAM}}}}}
& EfficientTAM-Ti/2 \cite{efficienttam} & 66.5 & 48.9 & 56.0 & 84.3 & 78.2\\
& +\textsc{SENTRY} & 68.7\scriptsize{(\textcolor{gray}{\textbf{2.2$\uparrow$}})} & 51.8\scriptsize{(\textcolor{gray}{\textbf{2.9$\uparrow$}})} & 57.0\scriptsize{(\textcolor{gray}{\textbf{1.0$\uparrow$}})} & 84.9\scriptsize{(\textcolor{gray}{\textbf{0.6$\uparrow$}})} & 78.5\scriptsize{(\textcolor{gray}{\textbf{0.3$\uparrow$}})}\\
\cline{2-7}
& SAM3 \cite{sam3} & 74.7 & 63.8 & 64.2 & 86.6 & 84.5\\
& +\textsc{SENTRY} & 75.8\scriptsize{(\textcolor{gray}{\textbf{1.1$\uparrow$}})} & 64.9\scriptsize{(\textcolor{gray}{\textbf{1.1$\uparrow$}})} & 65.4\scriptsize{(\textcolor{gray}{\textbf{1.2$\uparrow$}})} & 87.1\scriptsize{(\textcolor{gray}{\textbf{0.5$\uparrow$}})} & 85.2\scriptsize{(\textcolor{gray}{\textbf{0.7$\uparrow$}})}\\
\hline
\end{tabular}%
}
\vspace{-0.6em}
\end{table*}


\section{Qualitative Results} \label{supplement: Qualitative Results}

\noindent \textbf{Overall Performance.} We compare SENTRY against three SAM2-based variants, SAM2 \cite{sam2}, SAMURAI \cite{samurai}, and DAM4SAM \cite{dam4sam}, using the average AUC across four standard tracking benchmarks: LaSOT \cite{lasot}, LaSOT\textsubscript{ext} \cite{lasotext}, TNL2K \cite{tnl2k}, and TrackingNet \cite{trackingnet}.
As shown in Fig. 1(b),
SENTRY consistently outperforms all baselines. In particular, SENTRY achieves AUC improvements of 0.9\%, 2.7\%, and 1.0\% over SAM2, SAMURAI, and DAM4SAM, respectively. Notably, the separation between the bars is more pronounced in this figure, making the performance gap clearer. SENTRY on top of SAM2 performs almost similarly to SAMURAI, while SENTRY on top of SAMURAI shows better performance compared to DAM4SAM, further emphasizing the consistency of improvements across variants.


\noindent \textbf{Tracking Result.} Fig. \ref{fig:vr_results} illustrates how SENTRY mitigates key failure modes of existing SAM2-based variants.
Row 1: SAM2 loses identity after heavy occlusion due to ambiguous mask selection, whereas SENTRY-S2 disambiguates and refines the correct mask before memory writing.
Row 2: under abrupt scene change, SAMURAI’s Kalman prior diverges, while SENTRY-SR restores identity through neighbor-guided matching.
Row 3: In an underwater scene with haze and distractors, DAM4SAM mis-selects background objects due to heuristic thresholds, while SENTRY-D4S chooses the candidate maintaining short-term trajectory coherence.
Row 4: during an extremely long-term occlusion (\(\approx \)100 frames), all methods, including SENTRY, fail to recover the target, highlighting that SENTRY is not a global re-detector for prolonged disappearance followed by tiny-object re-identification. However, because SENTRY preserves the host update schedule and changes only the admitted mask, it can still recover when a plausible decoder, AMG, or motion-prior candidate exists, as quantified in Table~\ref{tab:recovery_after_failure}.

\noindent \textbf{Attention Visualization.}
We further analyze model behavior through spatial attention maps. GradCAM \cite{gradcam} is applied to the decoder cross-attention layers of each SAM2-based baseline to derive pixel-level activations aligned with predicted masks.
Fig. \ref{fig:2} shows three challenging cases: occlusion, abrupt motion, and distractor interference.
SAM2 (Fig. \ref{fig:2}(a)) loses identity once background regions dominate attention during occlusion.
SAMURAI (Fig. \ref{fig:2}(b)) improves short-term motion handling but drifts when its linear motion prior fails.
DAM4SAM (Fig. \ref{fig:2}(c)) suppresses isolated distractors yet becomes unstable in cluttered scenes.
The final row depicts the same scenarios with SENTRY.
Although SENTRY does not modify the decoder or memory architecture, its refine-before-write mechanism filters out temporally inconsistent masks through cycle-consistent reasoning. This results in cleaner, more target-focused memory updates, which in turn lead to sharper and more stable attention patterns during decoding.


\section{Comparison with SAM2Long} \label{supplement:sam2long_comparison}
\noindent Table~\ref{tab:sam2long_per_dataset} compares SAM2Long-L with the large-model SENTRY variants across the bounding-box and VOT-style tracking benchmarks.
We report SAM2Long-L as a standalone baseline because its constrained memory-tree inference over multiple branches is not a single-stream memory interface compatible with SENTRY's pre-write admission rule.
Across all datasets on average, all SENTRY variants outperform SAM2Long-L in both tracking score and FPS; SENTRY-D4S-L achieves the best overall tracking average, while SENTRY-S2-L achieves the highest FPS among the compared SENTRY variants.
\begin{table*}[!t]
\centering
\caption{Per-dataset comparison with SAM2Long on tracking benchmarks. All Avg. averages the nine tracking datasets.}
\label{tab:sam2long_per_dataset}
\vspace{-0.4em}
\setlength{\tabcolsep}{4pt}
\renewcommand{\arraystretch}{0.9}
\resizebox{\textwidth}{!}{%
\begin{tabular}{lccccccccc|cc}
\hline
\rowcolor{Gray}
{\small \textbf{Method}} 
& {\small \textbf{LaSOT}} 
& {\small \textbf{LaSOT}\textsubscript{\textbf{ext}}}
& {\small \textbf{TNL2K}} 
& {\small \textbf{GOT-10k}} 
& {\small \textbf{TrackingNet}} 
& {\small \textbf{VOT20}} 
& {\small \textbf{VOT22}} 
& {\small \textbf{VOTS24}}
& {\small \textbf{DiDi}} 
& {\small \textbf{All Avg.}}
& {\small \textbf{FPS}}
\\
\hline
{\small SAM2Long-L} 
& 73.9 & 60.9 & 58.4 & 81.1 & 84.7 
& 57.8 & 61.5 & 68.5 & 64.6 
& 67.9 & 26.1
\\
{\small SENTRY-S2-L} 
& 70.2 & 57.0 & 57.9 & 81.1 & 85.7 
& 69.0 & 70.1 & 66.9 & 66.0 
& 69.3 & \textbf{32.8}
\\
{\small SENTRY-SR-L} 
& 75.1 & 61.5 & 59.6 & 81.8 & 85.8 
& 68.6 & 71.5 & 68.1 & 69.3
& 71.3 & 30.9
\\
{\small SENTRY-D4S-L} 
& \textbf{76.3} & \textbf{61.8} & \textbf{61.3} & \textbf{82.1} & \textbf{85.9} 
& \textbf{73.2} & \textbf{75.9} & \textbf{71.5} & \textbf{70.5} 
& \textbf{73.2} & 30.2
\\
\hline
\end{tabular}%
}
\vspace{-0.6em}
\end{table*}

\begin{table*}[!t]

\centering

\caption{
\small{Performance comparison on LaSOT \cite{lasot}, LaSOT\textsubscript{ext} \cite{lasotext}, TNL2K \cite{tnl2k}, GOT-10k \cite{got10k}, and TrackingNet \cite{trackingnet} datasets between SAM2 baselines and SENTRY.}
}

\vspace{-1em}
\setlength{\tabcolsep}{4pt}
		\scalebox{0.99}[0.99]{
\resizebox{\textwidth}{!}{

\begin{tabular}{l|ccc|ccc|ccc|ccc|ccc}

\hline

\rowcolor{Gray} \textbf{Method} & \multicolumn{3}{c}{\textbf{LaSOT}} & \multicolumn{3}{c}{\textbf{LaSOT\textsubscript{ext}}} & \multicolumn{3}{c}{\textbf{TNL2K}} & \multicolumn{3}{c}{\textbf{GOT-10k}} & \multicolumn{3}{c}{\textbf{TrackingNet}} \\

\rowcolor{Gray} & S & NP & P & S & NP & P & S & NP & P & AO & SR\textsubscript{0.50} & SR\textsubscript{0.75} & S & NP & P \\
 
\hline

SAM2-T
& 67.0 & 74.0 & 71.5 
& 50.8 & 62.4 & 58.9 
& 55.8 & 74.2 & 60.7 
& 78.5 & 88.9 & 72.4 
& 83.9 & 89.8 & 85.9 
\\
+ SENTRY
& 67.6 \footnotesize{(\textcolor{gray}{\textbf{0.6$\uparrow$}})} & 74.7 \footnotesize{(\textcolor{gray}{\textbf{0.7$\uparrow$}})} & 71.7 \footnotesize{(\textcolor{gray}{\textbf{0.2$\uparrow$}})} 
& 51.3 \footnotesize{(\textcolor{gray}{\textbf{0.5$\uparrow$}})} & 63.3 \footnotesize{(\textcolor{gray}{\textbf{0.9$\uparrow$}})} & 59.1 \footnotesize{(\textcolor{gray}{\textbf{0.2$\uparrow$}})} 
& 57.0 \footnotesize{(\textcolor{gray}{\textbf{1.2$\uparrow$}})} & 75.6 \footnotesize{(\textcolor{gray}{\textbf{1.4$\uparrow$}})} & 62.4 \footnotesize{(\textcolor{gray}{\textbf{1.7$\uparrow$}})} 
& 78.8 \footnotesize{(\textcolor{gray}{\textbf{0.3$\uparrow$}})} & 89.3 \footnotesize{(\textcolor{gray}{\textbf{0.4$\uparrow$}})} & 73.2 \footnotesize{(\textcolor{gray}{\textbf{0.8$\uparrow$}})} 
& 84.2 \footnotesize{(\textcolor{gray}{\textbf{0.3$\uparrow$}})} & 90.0 \footnotesize{(\textcolor{gray}{\textbf{0.2$\uparrow$}})} & 86.0 \footnotesize{(\textcolor{gray}{\textbf{0.1$\uparrow$}})} 
\\ 
\hline

SAM2-S
& 66.3 & 73.4 & 71.0 
& 54.9 & 68.2 & 64.6 
& 56.4 & 75.6 & 61.9 
& 78.3 & 87.9 & 73.0 
& 84.4 & 90.4 & 86.9 
\\
+ SENTRY
& 67.8 \footnotesize{(\textcolor{gray}{\textbf{1.5$\uparrow$}}}) & 74.5 \footnotesize{(\textcolor{gray}{\textbf{1.1$\uparrow$}})} & 72.0 \footnotesize{(\textcolor{gray}{\textbf{1.0$\uparrow$}})} 
& 55.9 \footnotesize{(\textcolor{gray}{\textbf{1.0$\uparrow$}})} & 69.0 \footnotesize{(\textcolor{gray}{\textbf{0.8$\uparrow$}})} & 65.0 \footnotesize{(\textcolor{gray}{\textbf{0.4$\uparrow$}})} 
& 57.6 \footnotesize{(\textcolor{gray}{\textbf{1.2$\uparrow$}})} & 76.8 \footnotesize{(\textcolor{gray}{\textbf{1.2$\uparrow$}})} & 63.5 \footnotesize{(\textcolor{gray}{\textbf{1.6$\uparrow$}})} 
& 78.4 \footnotesize{(\textcolor{gray}{\textbf{0.1$\uparrow$}})} & 88.0 \footnotesize{(\textcolor{gray}{\textbf{0.1$\uparrow$}})} & 73.5 \footnotesize{(\textcolor{gray}{\textbf{0.5$\uparrow$}})} 
& 84.9 \footnotesize{(\textcolor{gray}{\textbf{0.5$\uparrow$}})} & 90.9 \footnotesize{(\textcolor{gray}{\textbf{0.5$\uparrow$}})} & 87.6 \footnotesize{(\textcolor{gray}{\textbf{0.7$\uparrow$}})} 
\\ 
\hline

SAM2-B
& 66.2 & 73.8 & 71.2 
& 54.3 & 67.7 & 63.6 
& 55.4 & 74.1 & 60.7 
& 78.0 & 88.8 & 71.6 
& 84.6 & 90.5 & 87.2 
\\
+ SENTRY
& 67.2 \footnotesize{(\textcolor{gray}{\textbf{1.0$\uparrow$}})} & 74.2 \footnotesize{(\textcolor{gray}{\textbf{0.4$\uparrow$}})} & 71.5 \footnotesize{(\textcolor{gray}{\textbf{0.3$\uparrow$}})} 
& 55.0 \footnotesize{(\textcolor{gray}{\textbf{0.7$\uparrow$}})} & 68.2 \footnotesize{(\textcolor{gray}{\textbf{0.5$\uparrow$}})} & 64.5 \footnotesize{(\textcolor{gray}{\textbf{0.9$\uparrow$}})} 
& 56.7 \footnotesize{(\textcolor{gray}{\textbf{1.3$\uparrow$}})} & 75.5 \footnotesize{(\textcolor{gray}{\textbf{1.4$\uparrow$}})} & 62.4 \footnotesize{(\textcolor{gray}{\textbf{1.7$\uparrow$}})} 
& 78.1 \footnotesize{(\textcolor{gray}{\textbf{0.1$\uparrow$}})} & 88.9 \footnotesize{(\textcolor{gray}{\textbf{0.1$\uparrow$}})} & 71.9 \footnotesize{(\textcolor{gray}{\textbf{0.3$\uparrow$}})} 
& 85.0 \footnotesize{(\textcolor{gray}{\textbf{0.4$\uparrow$}})} & 91.0 \footnotesize{(\textcolor{gray}{\textbf{0.5$\uparrow$}})} & 87.9 \footnotesize{(\textcolor{gray}{\textbf{0.7$\uparrow$}})} 
\\ 
\hline

SAM2-L
& 68.5 & 76.1 & 73.6 
& 56.8 & 71.1 & 67.0 
& 56.7 & 75.4 & 62.5 
& 80.8 & 91.3 & 75.5 
& 85.3 & 91.3 & 88.2 
\\
+ SENTRY
& 70.2 \footnotesize{(\textcolor{gray}{\textbf{1.7$\uparrow$}})} & 77.2 \footnotesize{(\textcolor{gray}{\textbf{1.1$\uparrow$}})} & 74.5 \footnotesize{(\textcolor{gray}{\textbf{0.9$\uparrow$}})} 
& 57.0 \footnotesize{(\textcolor{gray}{\textbf{0.2$\uparrow$}})} & 71.7 \footnotesize{(\textcolor{gray}{\textbf{0.6$\uparrow$}})} & 67.1 \footnotesize{(\textcolor{gray}{\textbf{0.1$\uparrow$}})} 
& 57.9 \footnotesize{(\textcolor{gray}{\textbf{1.2$\uparrow$}})} & 76.9 \footnotesize{(\textcolor{gray}{\textbf{1.5$\uparrow$}})} & 64.1 \footnotesize{(\textcolor{gray}{\textbf{1.6$\uparrow$}})} 
& 81.1 \footnotesize{(\textcolor{gray}{\textbf{0.3$\uparrow$}})} & 91.4 \footnotesize{(\textcolor{gray}{\textbf{0.1$\uparrow$}})} & 76.5 \footnotesize{(\textcolor{gray}{\textbf{1.0$\uparrow$}})} 
& 85.7 \footnotesize{(\textcolor{gray}{\textbf{0.4$\uparrow$}})} & 91.9 \footnotesize{(\textcolor{gray}{\textbf{0.6$\uparrow$}})} & 88.9 \footnotesize{(\textcolor{gray}{\textbf{0.7$\uparrow$}})} 
\\

\hline


SAMURAI-T
& 69.3 & 76.4 & 73.8 
& 53.6 & 65.9 & 62.2 
& 50.6 & 68.2 & 53.3 
& 77.5 & 87.8 & 71.0 
& 77.7 & 83.5 & 79.0 
\\
+ SENTRY
& 71.9 \footnotesize{(\textcolor{gray}{\textbf{2.6$\uparrow$}})} & 78.5 \footnotesize{(\textcolor{gray}{\textbf{2.1$\uparrow$}})} & 75.7 \footnotesize{(\textcolor{gray}{\textbf{1.9$\uparrow$}})} 
& 55.6 \footnotesize{(\textcolor{gray}{\textbf{2.0$\uparrow$}})} & 66.5 \footnotesize{(\textcolor{gray}{\textbf{0.6$\uparrow$}})} & 64.3 \footnotesize{(\textcolor{gray}{\textbf{2.1$\uparrow$}})} 
& 59.3 \footnotesize{(\textcolor{gray}{\textbf{8.7$\uparrow$}})} & 78.2 \footnotesize{(\textcolor{gray}{\textbf{10.0$\uparrow$}})} & 65.0 \footnotesize{(\textcolor{gray}{\textbf{11.7$\uparrow$}})} 
& 79.6 \footnotesize{(\textcolor{gray}{\textbf{2.1$\uparrow$}})} & 90.4 \footnotesize{(\textcolor{gray}{\textbf{2.6$\uparrow$}})} & 73.4 \footnotesize{(\textcolor{gray}{\textbf{2.4$\uparrow$}})} 
& 84.1 \footnotesize{(\textcolor{gray}{\textbf{6.4$\uparrow$}})} & 89.4 \footnotesize{(\textcolor{gray}{\textbf{5.9$\uparrow$}})} & 85.3 \footnotesize{(\textcolor{gray}{\textbf{6.3$\uparrow$}})} 
\\ 
\hline

SAMURAI-S
& 70.0 & 77.6 & 75.2 
& 56.5 & 69.8 & 66.3 
& 45.7 & 61.5 & 46.7 
& 76.7 & 86.4 & 71.0 
& 77.9 & 83.4 & 79.2 
\\
+ SENTRY
& 71.4 \footnotesize{(\textcolor{gray}{\textbf{1.4$\uparrow$}})} & 78.4 \footnotesize{(\textcolor{gray}{\textbf{0.8$\uparrow$}})} & 75.7 \footnotesize{(\textcolor{gray}{\textbf{0.5$\uparrow$}})} 
& 58.9 \footnotesize{(\textcolor{gray}{\textbf{2.4$\uparrow$}})} & 70.0 \footnotesize{(\textcolor{gray}{\textbf{0.2$\uparrow$}})} & 68.2 \footnotesize{(\textcolor{gray}{\textbf{1.9$\uparrow$}})} 
& 59.5 \footnotesize{(\textcolor{gray}{\textbf{13.8$\uparrow$}})} & 78.8 \footnotesize{(\textcolor{gray}{\textbf{17.3$\uparrow$}})} & 65.9 \footnotesize{(\textcolor{gray}{\textbf{19.2$\uparrow$}})} 
& 79.1 \footnotesize{(\textcolor{gray}{\textbf{2.4$\uparrow$}})} & 89.4 \footnotesize{(\textcolor{gray}{\textbf{3.0$\uparrow$}})} & 73.8 \footnotesize{(\textcolor{gray}{\textbf{2.8$\uparrow$}})} 
& 84.9 \footnotesize{(\textcolor{gray}{\textbf{7.0$\uparrow$}})} & 90.2 \footnotesize{(\textcolor{gray}{\textbf{6.8$\uparrow$}})} & 86.6 \footnotesize{(\textcolor{gray}{\textbf{7.4$\uparrow$}})} 
\\ 
\hline

SAMURAI-B
& 70.7 & 78.7 & 76.2 
& 55.9 & 69.5 & 65.6 
& 47.8 & 64.3 & 50.1 
& 77.8 & 88.6 & 71.1 
& 79.3 & 84.9 & 81.2 
\\
+ SENTRY
& 73.2 \footnotesize{(\textcolor{gray}{\textbf{2.5$\uparrow$}})} & 80.5 \footnotesize{(\textcolor{gray}{\textbf{1.8$\uparrow$}})} & 77.9 \footnotesize{(\textcolor{gray}{\textbf{1.7$\uparrow$}})} 
& 58.3 \footnotesize{(\textcolor{gray}{\textbf{2.4$\uparrow$}})} & 70.7 \footnotesize{(\textcolor{gray}{\textbf{1.2$\uparrow$}})} & 68.7 \footnotesize{(\textcolor{gray}{\textbf{3.1$\uparrow$}})} 
& 59.6 \footnotesize{(\textcolor{gray}{\textbf{11.8$\uparrow$}})} & 79.0 \footnotesize{(\textcolor{gray}{\textbf{14.7$\uparrow$}})} & 66.0 \footnotesize{(\textcolor{gray}{\textbf{15.9$\uparrow$}})} 
& 79.9 \footnotesize{(\textcolor{gray}{\textbf{2.1$\uparrow$}})} & 90.9 \footnotesize{(\textcolor{gray}{\textbf{2.3$\uparrow$}})} & 73.5 \footnotesize{(\textcolor{gray}{\textbf{2.4$\uparrow$}})} 
& 85.1 \footnotesize{(\textcolor{gray}{\textbf{5.8$\uparrow$}})} & 90.4 \footnotesize{(\textcolor{gray}{\textbf{5.5$\uparrow$}})} & 86.9 \footnotesize{(\textcolor{gray}{\textbf{5.7$\uparrow$}})} 
\\ 
\hline

SAMURAI-L
& 74.2 & 82.7 & 80.2 
& 59.5 & 74.1 & 70.8 
& 50.6 & 67.5 & 54.2 
& 81.6 & 91.8 & 76.8 
& 82.4 & 88.2 & 85.0 
\\
+ SENTRY
& 75.1 \footnotesize{(\textcolor{gray}{\textbf{0.9$\uparrow$}})} & 82.7 & 80.4 \footnotesize{(\textcolor{gray}{\textbf{0.2$\uparrow$}})} 
& 61.5 \footnotesize{(\textcolor{gray}{\textbf{2.0$\uparrow$}})} & 75.0 \footnotesize{(\textcolor{gray}{\textbf{0.9$\uparrow$}})} & 72.9 \footnotesize{(\textcolor{gray}{\textbf{2.1$\uparrow$}})} 
& 59.6 \footnotesize{(\textcolor{gray}{\textbf{9.0$\uparrow$}})} & 78.8 \footnotesize{(\textcolor{gray}{\textbf{11.3$\uparrow$}})} & 66.4 \footnotesize{(\textcolor{gray}{\textbf{12.2$\uparrow$}})} 
& 81.8 \footnotesize{(\textcolor{gray}{\textbf{0.2$\uparrow$}})} & 92.3 \footnotesize{(\textcolor{gray}{\textbf{0.5$\uparrow$}})} & 77.1 \footnotesize{(\textcolor{gray}{\textbf{0.3$\uparrow$}})} 
& 85.8 \footnotesize{(\textcolor{gray}{\textbf{3.4$\uparrow$}})} & 91.1 \footnotesize{(\textcolor{gray}{\textbf{2.9$\uparrow$}})} & 88.1 \footnotesize{(\textcolor{gray}{\textbf{3.1$\uparrow$}})} 
\\ 
\hline


DAM4SAM-T
& 72.4 & 80.2 & 77.7 
& 57.4 & 70.6 & 67.6 
& 59.4 & 79.0 & 65.5 
& 78.3 & 89.3 & 72.3 
& 83.9 & 89.7 & 85.3 
\\
+ SENTRY
& 73.5 \footnotesize{(\textcolor{gray}{\textbf{1.1$\uparrow$}})} & 81.7 \footnotesize{(\textcolor{gray}{\textbf{1.5$\uparrow$}})} & 79.3 \footnotesize{(\textcolor{gray}{\textbf{1.6$\uparrow$}})} 
& 58.4 \footnotesize{(\textcolor{gray}{\textbf{1.0$\uparrow$}})} & 71.6 \footnotesize{(\textcolor{gray}{\textbf{1.0$\uparrow$}})} & 68.9 \footnotesize{(\textcolor{gray}{\textbf{1.3$\uparrow$}})} 
& 61.1 \footnotesize{(\textcolor{gray}{\textbf{1.7$\uparrow$}})} & 80.9 \footnotesize{(\textcolor{gray}{\textbf{1.9$\uparrow$}})} & 67.6 \footnotesize{(\textcolor{gray}{\textbf{2.1$\uparrow$}})} 
& 79.0 \footnotesize{(\textcolor{gray}{\textbf{0.7$\uparrow$}})} & 89.5 \footnotesize{(\textcolor{gray}{\textbf{0.2$\uparrow$}})} & 73.2 \footnotesize{(\textcolor{gray}{\textbf{0.9$\uparrow$}})} 
& 84.3 \footnotesize{(\textcolor{gray}{\textbf{0.4$\uparrow$}})} & 89.9 \footnotesize{(\textcolor{gray}{\textbf{0.2$\uparrow$}})} & 85.7 \footnotesize{(\textcolor{gray}{\textbf{0.4$\uparrow$}})} 
\\ 
\hline

DAM4SAM-S
& 72.6 & 80.5 & 78.1 
& 59.2 & 73.2 & 70.2 
& 59.4 & 79.3 & 65.8 
& 74.0 & 82.0 & 66.4 
& 77.2 & 79.5 & 74.8 
\\
+ SENTRY
& 73.6 \footnotesize{(\textcolor{gray}{\textbf{1.0$\uparrow$}})} & 81.7 \footnotesize{(\textcolor{gray}{\textbf{1.2$\uparrow$}})} & 79.5 \footnotesize{(\textcolor{gray}{\textbf{1.4$\uparrow$}})} 
& 60.6 \footnotesize{(\textcolor{gray}{\textbf{1.4$\uparrow$}})} & 74.7 \footnotesize{(\textcolor{gray}{\textbf{1.5$\uparrow$}})} & 72.0 \footnotesize{(\textcolor{gray}{\textbf{1.8$\uparrow$}})} 
& 60.7 \footnotesize{(\textcolor{gray}{\textbf{1.3$\uparrow$}})} & 80.9 \footnotesize{(\textcolor{gray}{\textbf{1.6$\uparrow$}})} & 67.3 \footnotesize{(\textcolor{gray}{\textbf{1.5$\uparrow$}})} 
& 79.3 \footnotesize{(\textcolor{gray}{\textbf{5.3$\uparrow$}})} & 89.4 \footnotesize{(\textcolor{gray}{\textbf{7.4$\uparrow$}})} & 73.9 \footnotesize{(\textcolor{gray}{\textbf{7.5$\uparrow$}})} 
& 82.1 \footnotesize{(\textcolor{gray}{\textbf{4.9$\uparrow$}})} & 84.9 \footnotesize{(\textcolor{gray}{\textbf{5.4$\uparrow$}})} & 80.9 \footnotesize{(\textcolor{gray}{\textbf{6.1$\uparrow$}})} 
\\ 
\hline

DAM4SAM-B
& 73.2 & 81.3 & 79.0 
& 57.1 & 70.6 & 66.5 
& 58.4 & 78.1 & 64.6 
& 78.9 & 89.3 & 72.3 
& 84.7 & 90.5 & 86.7 
\\
+ SENTRY
& 75.1 \footnotesize{(\textcolor{gray}{\textbf{1.9$\uparrow$}})} & 83.2 \footnotesize{(\textcolor{gray}{\textbf{1.9$\uparrow$}})} & 81.0 \footnotesize{(\textcolor{gray}{\textbf{2.0$\uparrow$}})} 
& 59.4 \footnotesize{(\textcolor{gray}{\textbf{2.3$\uparrow$}})} & 73.3 \footnotesize{(\textcolor{gray}{\textbf{2.7$\uparrow$}})} & 69.7 \footnotesize{(\textcolor{gray}{\textbf{3.2$\uparrow$}})} 
& 59.9 \footnotesize{(\textcolor{gray}{\textbf{1.5$\uparrow$}})} & 79.8 \footnotesize{(\textcolor{gray}{\textbf{1.7$\uparrow$}})} & 66.4 \footnotesize{(\textcolor{gray}{\textbf{1.8$\uparrow$}})} 
& 79.2 \footnotesize{(\textcolor{gray}{\textbf{0.3$\uparrow$}})} & 90.3 \footnotesize{(\textcolor{gray}{\textbf{1.0$\uparrow$}})} & 73.1 \footnotesize{(\textcolor{gray}{\textbf{0.8$\uparrow$}})} 
& 84.7 & 90.7 \footnotesize{(\textcolor{gray}{\textbf{0.2$\uparrow$}})} & 86.9 \footnotesize{(\textcolor{gray}{\textbf{0.2$\uparrow$}})} 
\\ 
\hline

DAM4SAM-L
& 75.1 & 83.3 & 81.1 
& 60.7 & 75.3 & 72.2 
& 59.8 & 79.8 & 66.8 
& 81.3 & 91.4 & 77.2 
& 85.3 & 90.9 & 87.4 
\\
+ SENTRY
& 76.3 \footnotesize{(\textcolor{gray}{\textbf{1.2$\uparrow$}})} & 84.7 \footnotesize{(\textcolor{gray}{\textbf{1.4$\uparrow$}})} & 82.4 \footnotesize{(\textcolor{gray}{\textbf{1.3$\uparrow$}})} 
& 61.8 \footnotesize{(\textcolor{gray}{\textbf{1.1$\uparrow$}})} & 76.6 \footnotesize{(\textcolor{gray}{\textbf{1.3$\uparrow$}})} & 73.8 \footnotesize{(\textcolor{gray}{\textbf{1.6$\uparrow$}})} 
& 61.3 \footnotesize{(\textcolor{gray}{\textbf{1.5$\uparrow$}})} & 81.3 \footnotesize{(\textcolor{gray}{\textbf{1.5$\uparrow$}})} & 68.3 \footnotesize{(\textcolor{gray}{\textbf{1.5$\uparrow$}})} 
& 82.1 \footnotesize{(\textcolor{gray}{\textbf{0.8$\uparrow$}})} & 92.6 \footnotesize{(\textcolor{gray}{\textbf{1.2$\uparrow$}})} & 78.2 \footnotesize{(\textcolor{gray}{\textbf{1.0$\uparrow$}})} 
& 85.9 \footnotesize{(\textcolor{gray}{\textbf{0.6$\uparrow$}})} & 91.5 \footnotesize{(\textcolor{gray}{\textbf{0.6$\uparrow$}})} & 87.9 \footnotesize{(\textcolor{gray}{\textbf{0.5$\uparrow$}})} 
\\

\hline

\end{tabular}
}
}
\label{tab:sam2_comparison}

\end{table*}


\section{Additional quantitative results on box benchmarks} \label{supplement: Additional quantitative results on bounding box benchmarks}
\noindent We report detailed quantitative comparisons between SAM2-based baselines and their SENTRY counterparts across five large-scale bounding box benchmarks, LaSOT \cite{lasot}, LaSOT\textsubscript{ext} \cite{lasotext}, TNL2K \cite{tnl2k}, GOT-10k \cite{got10k}, and TrackingNet \cite{trackingnet}, summarized in Table \ref{tab:sam2_comparison}. A compact cross-benchmark comparison with SAM2Long-L is provided in Table \ref{tab:sam2long_per_dataset}. The following sections discuss results per dataset, highlighting where SENTRY contributes most and analyzing the causes behind these gains.

\noindent \textbf{LaSOT \cite{lasot}.} is a large-scale benchmark for VOT, consisting of 1,400 video sequences covering 70 object categories, each with 20 sequences (16 for training and 4 for testing). It provides diverse scenarios with substantial variations in appearance, scale, and motion, supporting both short-term and long-term evaluation.
Across all model scales, SENTRY consistently enhances SAM2-based trackers on LaSOT.  
The steady gains reflect improved temporal reliability during long, cluttered sequences where SAM2 often drifts after occlusion.
SENTRY’s cycle-consistent refinement suppresses such error accumulation, yielding average improvements of around 1--2\% in S for SAM2 and DAM4SAM, and up to 2.5\% for SAMURAI.  
The largest benefit appears in small and base models, which lack strong motion priors.  
SENTRY-D4S-L achieves the highest overall \textit{Acc} (S/NP/P = 76.3/84.7/82.4), surpassing its base by roughly 1\% across all metrics.

\noindent \textbf{LaSOT\textsubscript{ext} \cite{lasotext}.} is an extension of the LaSOT \cite{lasot} dataset, adding 150 additional test sequences distributed across 15 novel object categories that are not included in the original training set. 
This extension increases the evaluation diversity and challenges trackers to generalize to unseen categories.
On LaSOT\textsubscript{ext}, which emphasizes unseen object categories, SENTRY delivers consistent gains across all frameworks.  
Because this benchmark challenges cross-category generalization, temporal consistency becomes critical.
SENTRY enhances SAM2 and DAM4SAM by roughly 0.5--1.5\% in S, while SAMURAI benefits more strongly (around 2.0--2.5\%) as the temporal validation compensates for its linear motion assumption.  
SENTRY-D4S-L achieves 61.8/76.6/73.8 on S/NP/P, the best overall results among zero-shot methods, highlighting robust generalization to novel appearances.

\noindent \textbf{TNL2K \cite{tnl2k}.} is a large-scale natural language guided tracking dataset comprising 2,000 video sequences with corresponding linguistic descriptions. 
The dataset is divided into 1,300 sequences for training and 700 for testing. 
Each sequence is paired with natural language annotations that specify the target object and its attributes, enabling joint evaluation of visual and language-based tracking.
SENTRY produces its largest relative improvements on TNL2K, a dataset dominated by rapid motion and frequent occlusions.  
These conditions expose the weakness of confidence-based updates, making SENTRY’s temporal verification particularly effective.  
SAMURAI gains are dramatic, exceeding 10\% on average in S and up to 19\% in P, while SAM2 and DAM4SAM record consistent 1--2\% lifts.  
SENTRY-SR variants exhibit the strongest jumps, confirming that validating predictions over short trajectories substantially mitigates motion-induced drift.

\noindent \textbf{GOT-10k \cite{got10k}.} is a large-scale generic object tracking dataset built upon the WordNet hierarchy, containing over 10,000 video sequences spanning more than 560 object classes and 87 motion patterns. 
The dataset includes 180 videos for testing, while the remaining sequences are used for training. 
GOT-10k enforces zero overlap between training and testing categories to ensure unbiased evaluation and robust generalization analysis of tracking algorithms.
On GOT-10k, which evaluates generalization under unseen classes and fast-moving objects, SENTRY improves AO and SR across all scales.  
The improvement magnitude scales with model capacity; small models benefit most from the additional motion regularization, while large models show incremental but stable gains (up to 1\% AO).  
DAM4SAM-S shows the largest increase (5\% AO, 7\% SR\textsubscript{0.5}), indicating that SENTRY’s refinement complements heuristic distractor filtering with temporal stability.  
Overall, these gains demonstrate more reliable mask propagation and localization under varying dynamics.

\noindent \textbf{TrackingNet \cite{trackingnet}.} is a large-scale dataset designed for training and evaluating deep learning-based visual trackers. 
It consists of over 30,000 video sequences with dense annotations derived from real-world YouTube videos, providing diverse object appearances, motions, and scene contexts. 
The dataset is divided into 511 sequences for testing, while the remaining videos are used for training. 
TrackingNet offers realistic and large-scale benchmarks for short-term tracking in unconstrained environments.
TrackingNet features short, stable clips with dense annotations; thus, relative gains are smaller but consistent.
SENTRY enhances SAM2 and DAM4SAM by roughly 0.5--0.7\% across metrics, while SAMURAI benefits more notably (3--4\%) due to refined motion consistency.  
The effect is most pronounced in NP/P, showing that SENTRY improves frame-level ranking and smoothness even when spatial noise is low.  
These results confirm that SENTRY’s temporal refinement improves stability and ranking precision, maintaining benefits even in high-quality short-term tracking scenarios.

\begin{table*}[h!]
\centering
\caption{Attribute-wise AUC (\%) comparison on LaSOT \cite{lasot} dataset between SAM2 variants and our SENTRY variants.}

\label{tab:lasot_attribute}

\vspace{-1em}
\setlength{\tabcolsep}{9pt}
		\scalebox{0.99}[0.99]{

\resizebox{\linewidth}{!}{
\begin{tabular}{l|ccccccccccccccc}
    
\hline
    
\rowcolor{Gray} Method & ARC & BC & CM & DEF & FM & FOC & IV & LR & MB & OV & POC & ROT & SV & VC \\
    
\hline
    
\begin{tabular}[l]{@{}c@{}}SAM2-L\end{tabular} & 0.673 & 0.643 & 0.69 & 0.708 & 0.585 & 0.594 & 0.631 & 0.595 & 0.678 & 0.616 & 0.679 & 0.673 & 0.68 & 0.611 \\

+ SENTRY & 0.691 & 0.668 & 0.714 & 0.722 & 0.602 & 0.617 & 0.657 & 0.611 & 0.707 & 0.63 & 0.698 & 0.689 & 0.697 & 0.623 \\

\rowcolor{gaincolor} \textbf{\% Gain} & +1.8\% & +2.5\% & +2.4\% & +1.4\% & +1.7\% & +2.3\% & +2.6\% & +1.6\% & +2.9\% & +1.4\% & +2\% & +1.6\% & +1.7\% & +1.2\% \\

\specialrule{1pt}{0pt}{0pt}

\begin{tabular}[l]{@{}c@{}}SAMURAI-L\end{tabular} & 0.732 & 0.698 & 0.774 & 0.755 & 0.64 & 0.669 & 0.733 & 0.677 & 0.744 & 0.701 & 0.729 & 0.727 & 0.739 & 0.717 \\

+ SENTRY & 0.742 & 0.704 & 0.783 & 0.766 & 0.643 & 0.677 & 0.734 & 0.687 & 0.751 & 0.71 & 0.735 & 0.74 & 0.748 & 0.722 \\

\rowcolor{gaincolor} \textbf{\% Gain} & +1\% & +0.6\% & +0.9\% & +1.1\% & +0.3\% & +0.8\% & +0.1\% & +1\% & +0.7\% & +0.9\% & +0.6\% & +1.3\% & +0.9\% & +0.5\% \\

\specialrule{1pt}{0pt}{0pt}

\begin{tabular}[l]{@{}c@{}}DAM4SAM-L\end{tabular} & 0.74 & 0.694 & 0.789 & 0.75 & 0.688 & 0.688 & 0.743 & 0.692 & 0.763 & 0.709 & 0.735 & 0.726 & 0.748 & 0.759 \\

+ SENTRY & 0.753 & 0.718 & 0.803 & 0.767 & 0.695 & 0.703 & 0.752 & 0.712 & 0.773 & 0.723 & 0.749 & 0.741 & 0.76 & 0.76 \\

\rowcolor{gaincolor} \textbf{\% Gain} & +1.3\% & +2.4\% & +1.4\% & +1.7\% & +0.7\% & +1.5\% & +0.9\% & +2\% & +1\% & +1.4\% & +1.4\% & +1.5\% & +1.2\% & +0.1\% \\

\hline
        
\end{tabular}
}
}
\end{table*}


\begin{table*}[h!]
\centering
\caption{Attribute-wise AUC (\%) comparison on LaSOT\textsubscript{ext} \cite{lasotext} dataset between SAM2 variants and our SENTRY variants.}

\label{tab:lasotext_attribute}

\vspace{-1em}
\setlength{\tabcolsep}{9pt}
		\scalebox{0.99}[0.99]{

\resizebox{\linewidth}{!}{
\begin{tabular}{l|ccccccccccccccc}
    
\hline
    
\rowcolor{Gray} Method & ARC & BC & CM & DEF & FM & FOC & IV & LR & MB & OV & POC & ROT & SV & VC \\

\hline
  
\begin{tabular}[l]{@{}c@{}}SAM2-L\end{tabular} & 0.549 & 0.502 & 0.611 & 0.73 & 0.453 & 0.463 & 0.675 & 0.475 & 0.466 & 0.498 & 0.575 & 0.627 & 0.558 & 0.611 \\
+ SENTRY & 0.551 & 0.506 & 0.624 & 0.731 & 0.454 & 0.466 & 0.68 & 0.478 & 0.473 & 0.499 & 0.576 & 0.632 & 0.559 & 0.611 \\

\rowcolor{gaincolor} \textbf{\% Gain} & +0.2\% & +0.4\% & +1.3\% & +0.1\% & +0.1\% & +0.3\% & +0.5\% & +0.3\% & +0.7\% & +0.1\% & +0.1\% & +0.5\% & +0.1\% & 0\% \\

\specialrule{1pt}{0pt}{0pt}

\begin{tabular}[l]{@{}c@{}}SAMURAI-L\end{tabular} & 0.577 & 0.533 & 0.754 & 0.751 & 0.513 & 0.513 & 0.679 & 0.521 & 0.491 & 0.515 & 0.588 & 0.623 & 0.586 & 0.671 \\

+ SENTRY & 0.581 & 0.544 & 0.765 & 0.752 & 0.517 & 0.514 & 0.679 & 0.524 & 0.493 & 0.521 & 0.589 & 0.643 & 0.59 & 0.671 \\

\rowcolor{gaincolor} \textbf{\% Gain} & +0.4\% & +1.1\% & +1.1\% & +0.1\% & +0.4\% & +0.1\% & 0\% & +0.3\% & +0.2\% & +0.6\% & +0.1\% & +2\% & +0.4\% & 0\% \\

\specialrule{1pt}{0pt}{0pt}

\begin{tabular}[l]{@{}c@{}}DAM4SAM-L\end{tabular} & 0.592 & 0.554 & 0.747 & 0.759 & 0.498 & 0.504 & 0.703 & 0.521 & 0.474 & 0.503 & 0.604 & 0.657 & 0.599 & 0.676 \\

+ SENTRY & 0.605 & 0.572 & 0.761 & 0.759 & 0.514 & 0.521 & 0.703 & 0.536 & 0.494 & 0.521 & 0.616 & 0.675 & 0.611 & 0.683 \\

\rowcolor{gaincolor} \textbf{\% Gain} & +1.3\% & +1.8\% & +1.4\% & 0\% & +1.6\% & +1.7\% & 0\% & +1.5\% & +2\% & +1.8\% & +1.2\% & +1.8\% & +1.2\% & +0.7\% \\

\hline
        
\end{tabular}
}
}
\end{table*}


\begin{table*}[h!]
\centering
\caption{Attribute-wise AUC (\%) comparison on TNL2K \cite{tnl2k} dataset between SAM2 variants and our SENTRY variants.}

\label{tab:attribute_tnl2k}

\vspace{-1em}
\setlength{\tabcolsep}{4pt}
		\scalebox{0.99}[0.99]{

\resizebox{\linewidth}{!}{
\begin{tabular}{l|cccccccccccccccccc}

\hline

\rowcolor{Gray} Method & ARC & BC & CM & DEF & FM & FOC & IV & LR & MB & OV & POC & ROT & SV & VC & AS & MS & TC \\

\hline
  
\begin{tabular}[l]{@{}c@{}}SAM2-L\end{tabular} & 0.607 & 0.567 & 0.568 & 0.602 & 0.594 & 0.451 & 0.613 & 0.369 & 0.528 & 0.409 & 0.531 & 0.607 & 0.533 & 0.542 & 0.672 & 0.151 & 0.304 \\
+ SENTRY & 0.610 & 0.574 & 0.59 & 0.606 & 0.6 & 0.457 & 0.616 & 0.439 & 0.537 & 0.411 & 0.547 & 0.617 & 0.537 & 0.55 & 0.677 & 0.32 & 0.388 \\

\rowcolor{gaincolor} \textbf{\% Gain} & +0.3\% & +0.7\% & +2.2\% & +0.4\% & +0.6\% & +0.6\% & +0.3\% & +7\% & +0.9\% & +0.2\% & +1.6\% & +1\% & +0.4\% & +0.8\% & +0.5\% & +16.9\% & +8.4\% \\

\specialrule{1pt}{0pt}{0pt}

\begin{tabular}[l]{@{}c@{}}SAMURAI-L\end{tabular} & 0.563 & 0.51 & 0.53 & 0.536 & 0.551 & 0.386 & 0.588 & 0.296 & 0.52 & 0.371 & 0.47 & 0.574 & 0.476 & 0.512 & 0.69 & 0.124 & 0.252 \\
+ SENTRY & 0.647 & 0.596 & 0.626 & 0.63 & 0.628 & 0.48 & 0.647 & 0.467 & 0.58 & 0.454 & 0.572 & 0.639 & 0.577 & 0.573 & 0.697 & 0.325 & 0.389 \\

\rowcolor{gaincolor} \textbf{\% Gain} & +8.4\% & +8.6\% & +9.6\% & +9.4\% & +7.7\% & +9.4\% & +5.9\% & +17.1\% & +6\% & +10.2\% & +10.2\% & +6.5\% & +10.1\% & +6.1\% & +0.7\% & +20.1\% & +13.7\% \\

\specialrule{1pt}{0pt}{0pt}

\begin{tabular}[l]{@{}c@{}}DAM4SAM-L\end{tabular} & 0.637 & 0.598 & 0.62 & 0.634 & 0.616 & 0.502 & 0.637 & 0.485 & 0.577 & 0.447 & 0.586 & 0.645 & 0.572 & 0.576 & 0.706 & 0.381 & 0.413 \\
+ SENTRY & 0.646 & 0.608 & 0.627 & 0.641 & 0.62 & 0.515 & 0.639 & 0.489 & 0.578 & 0.46 & 0.593 & 0.651 & 0.581 & 0.587 & 0.707 & 0.391 & 0.429 \\

\rowcolor{gaincolor} \textbf{\% Gain} & +0.9\% & +1\% & +0.7\% & +0.7\% & +0.4\% & +1.3\% & +0.2\% & +0.4\% & +0.1\% & +1.3\% & +0.7\% & +0.6\% & +0.9\% & +1.1\% & +0.1\% & +1\% & +1.6\% \\

\hline
        
\end{tabular}
}
}
\end{table*}


\begin{table*}[h!]
\centering
\caption{\small{Performance comparison on VOT20-24 \cite{vot2020, vot2022, vots2024}, and DiDi \cite{dam4sam} datasets between SAM2 baselines and SENTRY.}}
\label{tab:sam_comp_vot}

\vspace{-1em}
\setlength{\tabcolsep}{3pt}
		\scalebox{0.99}[0.99]{
        
\resizebox{\textwidth}{!}{

\begin{tabular}{l|ccc|ccc|ccc|ccc}

\hline

\rowcolor{Gray} Method & \multicolumn{3}{c}{\textbf{VOT20}} & \multicolumn{3}{c}{\textbf{VOT22}} & \multicolumn{3}{c}{\textbf{VOTS24}} & \multicolumn{3}{c}{\textbf{DiDi}} \\

\rowcolor{Gray} & \textit{Q} & \textit{Acc} & \textit{Rob} & \textit{Q} & \textit{Acc} & \textit{Rob} & \textit{Q} & \textit{Acc} & \textit{Rob} & \textit{Q} & \textit{Acc} & \textit{Rob} \\

\hline

SAM2-T
& 62.1 & 72.3 & 93.2
& 62.3 & 72.4 & 93.4 
& 62.8 & 75.6 & 77.9 
& 60.5 & 70.2 & 84.9
\\
+ SENTRY
& 63.0 \footnotesize{(\textcolor{gray}{\textbf{0.9$\uparrow$}})} & 73.6 \footnotesize{(\textcolor{gray}{\textbf{1.3$\uparrow$}})} & 94.0 \footnotesize{(\textcolor{gray}{\textbf{0.8$\uparrow$}})}
& 63.7 \footnotesize{(\textcolor{gray}{\textbf{1.4$\uparrow$}})} & 73.6 \footnotesize{(\textcolor{gray}{\textbf{1.2$\uparrow$}})} & 94.1 \footnotesize{(\textcolor{gray}{\textbf{0.7$\uparrow$}})} 
& 63.6 \footnotesize{(\textcolor{gray}{\textbf{0.8$\uparrow$}})} & 76.4 \footnotesize{(\textcolor{gray}{\textbf{0.8$\uparrow$}})} & 78.7 \footnotesize{(\textcolor{gray}{\textbf{0.8$\uparrow$}})}
& 61.6 \footnotesize{(\textcolor{gray}{\textbf{1.1$\uparrow$}})} & 71.2 \footnotesize{(\textcolor{gray}{\textbf{1.0$\uparrow$}})} & 85.8 \footnotesize{(\textcolor{gray}{\textbf{0.9$\uparrow$}})}
\\

\hline

SAM2-S
& 64.3 & 73.3 & 93.5 
& 64.4 & 73.3 & 93.6 
& 60.5 & 75.1 & 76.7
& 62.4 & 70.9 & 86.7
\\
+ SENTRY
& 65.2 \footnotesize{(\textcolor{gray}{\textbf{0.9$\uparrow$}})} & 74.4 \footnotesize{(\textcolor{gray}{\textbf{1.1$\uparrow$}})} & 94.4 \footnotesize{(\textcolor{gray}{\textbf{0.9$\uparrow$}})}
& 65.3 \footnotesize{(\textcolor{gray}{\textbf{0.9$\uparrow$}})} & 74.4 \footnotesize{(\textcolor{gray}{\textbf{1.1$\uparrow$}})} & 94.4 \footnotesize{(\textcolor{gray}{\textbf{0.8$\uparrow$}})}
& 61.2 \footnotesize{(\textcolor{gray}{\textbf{0.7$\uparrow$}})} & 76.8 \footnotesize{(\textcolor{gray}{\textbf{1.7$\uparrow$}})} & 77.3 \footnotesize{(\textcolor{gray}{\textbf{0.6$\uparrow$}})}
& 63.6 \footnotesize{(\textcolor{gray}{\textbf{1.2$\uparrow$}})} & 72.0 \footnotesize{(\textcolor{gray}{\textbf{1.1$\uparrow$}})} & 87.8 \footnotesize{(\textcolor{gray}{\textbf{1.1$\uparrow$}})}
\\

\hline

SAM2-B
& 64.4 & 73.8 & 93.4 
& 64.0 & 73.7 & 93.2 
& 62.8 & 77.0 & 76.2
& 63.6 & 72.0 & 87.0
\\
+ SENTRY
& 65.2 \footnotesize{(\textcolor{gray}{\textbf{0.8$\uparrow$}})} & 74.5 \footnotesize{(\textcolor{gray}{\textbf{0.7$\uparrow$}})} & 94.3 \footnotesize{(\textcolor{gray}{\textbf{0.9$\uparrow$}})}
& 64.7 \footnotesize{(\textcolor{gray}{\textbf{0.7$\uparrow$}})} & 74.4 \footnotesize{(\textcolor{gray}{\textbf{0.7$\uparrow$}})} & 93.9 \footnotesize{(\textcolor{gray}{\textbf{0.7$\uparrow$}})}
& 63.4 \footnotesize{(\textcolor{gray}{\textbf{0.6$\uparrow$}})} & 77.6 \footnotesize{(\textcolor{gray}{\textbf{0.6$\uparrow$}})} & 76.8 \footnotesize{(\textcolor{gray}{\textbf{0.6$\uparrow$}})}
& 64.7 \footnotesize{(\textcolor{gray}{\textbf{1.1$\uparrow$}})} & 73.0 \footnotesize{(\textcolor{gray}{\textbf{1.0$\uparrow$}})} & 88.3 \footnotesize{(\textcolor{gray}{\textbf{1.3$\uparrow$}})}
\\

\hline

SAM2-L
& 68.1 & 77.8 & 94.1
& 69.2 & 77.9 & 94.6 
& 66.1 & 79.1 & 79.0
& 64.7 & 72.3 & 87.7
\\
+ SENTRY
& 68.8 \footnotesize{(\textcolor{gray}{\textbf{0.7$\uparrow$}})} & 78.3 \footnotesize{(\textcolor{gray}{\textbf{0.5$\uparrow$}})} & 94.7 \footnotesize{(\textcolor{gray}{\textbf{0.6$\uparrow$}})}
& 70.1 \footnotesize{(\textcolor{gray}{\textbf{0.9$\uparrow$}})} & 78.7 \footnotesize{(\textcolor{gray}{\textbf{0.8$\uparrow$}})} & 95.3 \footnotesize{(\textcolor{gray}{\textbf{0.7$\uparrow$}})}
& 66.9 \footnotesize{(\textcolor{gray}{\textbf{0.8$\uparrow$}})} & 79.3 \footnotesize{(\textcolor{gray}{\textbf{0.2$\uparrow$}})} & 79.5 \footnotesize{(\textcolor{gray}{\textbf{0.5$\uparrow$}})}
& 66.0 \footnotesize{(\textcolor{gray}{\textbf{1.3$\uparrow$}})} & 73.4 \footnotesize{(\textcolor{gray}{\textbf{1.1$\uparrow$}})} & 89.2 \footnotesize{(\textcolor{gray}{\textbf{1.5$\uparrow$}})}
\\

\hline

SAMURAI-T
& 63.3 & 72.5 & 93.2 
& 63.7 & 73.0 & 93.9
& 63.9 & 75.1 & 83.7
& 62.7 & 68.1 & 89.9
\\
+ SENTRY
& 64.5 \footnotesize{(\textcolor{gray}{\textbf{1.2$\uparrow$}})} & 73.6 \footnotesize{(\textcolor{gray}{\textbf{1.1$\uparrow$}})} & 94.8 \footnotesize{(\textcolor{gray}{\textbf{1.6$\uparrow$}})}
& 65.0 \footnotesize{(\textcolor{gray}{\textbf{1.3$\uparrow$}})} & 73.5 \footnotesize{(\textcolor{gray}{\textbf{0.5$\uparrow$}})} & 95.1 \footnotesize{(\textcolor{gray}{\textbf{1.2$\uparrow$}})}
& 65.0 \footnotesize{(\textcolor{gray}{\textbf{1.1$\uparrow$}})} & 75.7 \footnotesize{(\textcolor{gray}{\textbf{0.6$\uparrow$}})} & 84.2 \footnotesize{(\textcolor{gray}{\textbf{0.5$\uparrow$}})}
& 63.5 \footnotesize{(\textcolor{gray}{\textbf{0.8$\uparrow$}})} & 69.0 \footnotesize{(\textcolor{gray}{\textbf{0.9$\uparrow$}})} & 90.8 \footnotesize{(\textcolor{gray}{\textbf{0.9$\uparrow$}})}
\\

\hline

SAMURAI-S
& 65.1 & 71.2 & 94.2
& 65.3 & 72.4 & 94.9
& 64.5 & 75.2 & 83.3
& 64.6 & 70.1 & 90.6
\\
+ SENTRY
& 66.7 \footnotesize{(\textcolor{gray}{\textbf{1.6$\uparrow$}})} & 73.8 \footnotesize{(\textcolor{gray}{\textbf{2.6$\uparrow$}})} & 95.8 \footnotesize{(\textcolor{gray}{\textbf{1.6$\uparrow$}})}
& 67.9 \footnotesize{(\textcolor{gray}{\textbf{2.6$\uparrow$}})} & 73.8 \footnotesize{(\textcolor{gray}{\textbf{1.4$\uparrow$}})} & 96.3 \footnotesize{(\textcolor{gray}{\textbf{1.4$\uparrow$}})}
& 65.7 \footnotesize{(\textcolor{gray}{\textbf{1.2$\uparrow$}})} & 76.0 \footnotesize{(\textcolor{gray}{\textbf{0.8$\uparrow$}})} & 83.8 \footnotesize{(\textcolor{gray}{\textbf{0.5$\uparrow$}})}
& 65.3 \footnotesize{(\textcolor{gray}{\textbf{0.7$\uparrow$}})} & 70.9 \footnotesize{(\textcolor{gray}{\textbf{0.8$\uparrow$}})} & 91.4 \footnotesize{(\textcolor{gray}{\textbf{0.8$\uparrow$}})}
\\

\hline

SAMURAI-B
& 66.4 & 73.2 & 95.0 
& 66.8 & 73.8 & 95.3 
& 64.6 & 75.5 & 83.3
& 67.1 & 71.5 & 93.0
\\
+ SENTRY
& 68.5 \footnotesize{(\textcolor{gray}{\textbf{2.1$\uparrow$}})} & 74.5 \footnotesize{(\textcolor{gray}{\textbf{1.3$\uparrow$}})} & 96.3 \footnotesize{(\textcolor{gray}{\textbf{1.3$\uparrow$}})}
& 69.4 \footnotesize{(\textcolor{gray}{\textbf{2.6$\uparrow$}})} & 74.5 \footnotesize{(\textcolor{gray}{\textbf{0.7$\uparrow$}})} & 96.6 \footnotesize{(\textcolor{gray}{\textbf{1.3$\uparrow$}})} 
& 65.7 \footnotesize{(\textcolor{gray}{\textbf{1.1$\uparrow$}})} & 76.7 \footnotesize{(\textcolor{gray}{\textbf{1.2$\uparrow$}})} & 83.7 \footnotesize{(\textcolor{gray}{\textbf{0.4$\uparrow$}})}
& 67.8 \footnotesize{(\textcolor{gray}{\textbf{0.7$\uparrow$}})} & 72.0 \footnotesize{(\textcolor{gray}{\textbf{0.5$\uparrow$}})} & 94.1 \footnotesize{(\textcolor{gray}{\textbf{0.9$\uparrow$}})}
\\

\hline

SAMURAI-L
& 67.4 & 74.1 & 94.2
& 69.3 & 74.4 & 95.1
& 67.3 & 77.6 & 85.1
& 67.0 & 71.4 & 92.3
\\
+ SENTRY
& 68.6 \footnotesize{(\textcolor{gray}{\textbf{1.2$\uparrow$}})} & 77.5 \footnotesize{(\textcolor{gray}{\textbf{3.4$\uparrow$}})} & 94.7 \footnotesize{(\textcolor{gray}{\textbf{0.5$\uparrow$}})} 
& 71.5 \footnotesize{(\textcolor{gray}{\textbf{2.1$\uparrow$}})} & 77.5 \footnotesize{(\textcolor{gray}{\textbf{3.4$\uparrow$}})} & 95.7 \footnotesize{(\textcolor{gray}{\textbf{0.6$\uparrow$}})}
& 68.1 \footnotesize{(\textcolor{gray}{\textbf{0.8$\uparrow$}})} & 78.2 \footnotesize{(\textcolor{gray}{\textbf{0.6$\uparrow$}})} & 85.6 \footnotesize{(\textcolor{gray}{\textbf{0.5$\uparrow$}})}
& 68.5 \footnotesize{(\textcolor{gray}{\textbf{1.5$\uparrow$}})} & 72.9 \footnotesize{(\textcolor{gray}{\textbf{1.5$\uparrow$}})} & 93.9 \footnotesize{(\textcolor{gray}{\textbf{1.6$\uparrow$}})}
\\

\hline

DAM4SAM-T
& 67.0 & 73.7 & 96.1 
& 68.1 & 73.9 & 96.4
& 65.3 & 75.3 & 84.2
& 63.2 & 69.3 & 89.8
\\
+ SENTRY
& 67.4 \footnotesize{(\textcolor{gray}{\textbf{0.4$\uparrow$}})} & 74.0 \footnotesize{(\textcolor{gray}{\textbf{0.3$\uparrow$}})} & 96.4 \footnotesize{(\textcolor{gray}{\textbf{0.3$\uparrow$}})}
& 68.9 \footnotesize{(\textcolor{gray}{\textbf{0.8$\uparrow$}})} & 74.3 \footnotesize{(\textcolor{gray}{\textbf{0.4$\uparrow$}})} & 96.8 \footnotesize{(\textcolor{gray}{\textbf{0.4$\uparrow$}})}

& 65.7 \footnotesize{(\textcolor{gray}{\textbf{0.4$\uparrow$}})} & 75.4 \footnotesize{(\textcolor{gray}{\textbf{0.1$\uparrow$}})} & 84.5 \footnotesize{(\textcolor{gray}{\textbf{0.3$\uparrow$}})}

& 64.1 \footnotesize{(\textcolor{gray}{\textbf{0.9$\uparrow$}})} & 69.8 \footnotesize{(\textcolor{gray}{\textbf{0.5$\uparrow$}})} & 90.5 \footnotesize{(\textcolor{gray}{\textbf{0.7$\uparrow$}})}
\\

\hline

DAM4SAM-S
& 54.9 & 73.3 & 87.3 
& 55.6 & 73.2 & 88.4
& 59.3 & 65.6 & 84.9
& 66.3 & 71.0 & 92.3
\\
+ SENTRY
& 55.5 \footnotesize{(\textcolor{gray}{\textbf{0.6$\uparrow$}})} & 73.8 \footnotesize{(\textcolor{gray}{\textbf{0.5$\uparrow$}})} & 87.7 \footnotesize{(\textcolor{gray}{\textbf{0.4$\uparrow$}})}
& 56.1 \footnotesize{(\textcolor{gray}{\textbf{0.5$\uparrow$}})} & 73.6 \footnotesize{(\textcolor{gray}{\textbf{0.4$\uparrow$}})} & 88.9 \footnotesize{(\textcolor{gray}{\textbf{0.5$\uparrow$}})}

& 59.8 \footnotesize{(\textcolor{gray}{\textbf{0.5$\uparrow$}})} & 65.9 \footnotesize{(\textcolor{gray}{\textbf{0.3$\uparrow$}})} & 85.3 \footnotesize{(\textcolor{gray}{\textbf{0.4$\uparrow$}})}

& 66.8 \footnotesize{(\textcolor{gray}{\textbf{0.5$\uparrow$}})} & 71.6 \footnotesize{(\textcolor{gray}{\textbf{0.6$\uparrow$}})} & 92.9 \footnotesize{(\textcolor{gray}{\textbf{0.6$\uparrow$}})}
\\

\hline

DAM4SAM-B
& 70.1 & 74.5 & 96.5 
& 70.4 & 74.4 & 96.4
& 68.3 & 77.0 & 86.5
& 66.4 & 70.8 & 93.1
\\
+ SENTRY
& 70.7 \footnotesize{(\textcolor{gray}{\textbf{0.6$\uparrow$}})} & 75.0 \footnotesize{(\textcolor{gray}{\textbf{0.5$\uparrow$}})} & 97.1 \footnotesize{(\textcolor{gray}{\textbf{0.6$\uparrow$}})}
& 71.0 \footnotesize{(\textcolor{gray}{0.6$\uparrow$})} & 75.0 \footnotesize{(\textcolor{gray}{\textbf{0.6$\uparrow$}})} & 97.0 \footnotesize{(\textcolor{gray}{\textbf{0.6$\uparrow$}})}

& 68.8 \footnotesize{(\textcolor{gray}{\textbf{0.5$\uparrow$}})} & 77.2 \footnotesize{(\textcolor{gray}{\textbf{0.2$\uparrow$}})} & 86.9 \footnotesize{(\textcolor{gray}{\textbf{0.4$\uparrow$}})}

& 67.2 \footnotesize{(\textcolor{gray}{\textbf{0.8$\uparrow$}})} & 71.4 \footnotesize{(\textcolor{gray}{\textbf{0.6$\uparrow$}})} & 93.7 \footnotesize{(\textcolor{gray}{\textbf{0.6$\uparrow$}})}
\\

\hline

DAM4SAM-L
& 72.3 & 79.6 & 96.1
& 75.0 & 79.7 & 97.1
& 71.1 & 79.3 & 86.4
& 69.4 & 72.7 & 94.4
\\
+ SENTRY
& 72.8 \footnotesize{(\textcolor{gray}{\textbf{0.5$\uparrow$}})} & 80.1 \footnotesize{(\textcolor{gray}{\textbf{0.5$\uparrow$}})} & 96.6 \footnotesize{(\textcolor{gray}{\textbf{0.5$\uparrow$}})} 
& 75.6 \footnotesize{(\textcolor{gray}{\textbf{0.6$\uparrow$}})} & 80.3 \footnotesize{(\textcolor{gray}{\textbf{0.6$\uparrow$}})} & 97.6 \footnotesize{(\textcolor{gray}{\textbf{0.5$\uparrow$}})}
& 71.5 \footnotesize{(\textcolor{gray}{\textbf{0.4$\uparrow$}})} & 79.4 \footnotesize{(\textcolor{gray}{\textbf{0.1$\uparrow$}})} & 86.7 \footnotesize{(\textcolor{gray}{\textbf{0.3$\uparrow$}})}
& 70.2 \footnotesize{(\textcolor{gray}{\textbf{0.8$\uparrow$}})} & 73.3 \footnotesize{(\textcolor{gray}{\textbf{0.6$\uparrow$}})} & 95.1 \footnotesize{(\textcolor{gray}{\textbf{0.7$\uparrow$}})}
\\

\hline

\end{tabular}
}
}
\end{table*}


\noindent \textbf{Attribute-Wise Performance Analysis.} We analyze how SENTRY affects specific failure modes on LaSOT, LaSOT\textsubscript{ext}, and TNL2K; full results appear in Tables \ref{tab:lasot_attribute}-\ref{tab:attribute_tnl2k}. 
Across all bases, the refine-before-write step primarily helps when instantaneous confidence is unreliable, e.g., under MB, occlusion, LR, or distractor pressure, by validating predictions over short trajectories before memory updates.

\begin{itemize}
    \item \textbf{LaSOT.}
    On long, cluttered sequences, SENTRY yields consistent gains for all three bases (Table \ref{tab:lasot_attribute}).  
    The largest improvements for SAM2-L$\to$SENTRY-S2-L occur on MB, IV, BC, and FOC, aligning with cases where greedy updates often corrupt memory after a few uncertain frames.  
    DAM4SAM-L$\to$SENTRY-D4S-L shows notable boosts on BC and LR, indicating that temporal verification complements distractor-aware heuristics by rejecting look-alike false positives that briefly spike in confidence.  
    SAMURAI-L$\to$SENTRY-SR-L sees modest but widespread lifts; when the Kalman prior is approximately correct, SENTRY reduces overconfident writes on partial occlusions and rotations, smoothing recovery.

    \item \textbf{LaSOT\textsubscript{ext}.}  
    With unseen categories (Table \ref{tab:lasotext_attribute}), appearance priors are less predictive, and temporal consistency becomes the dominant cue.  
    SENTRY-S2-L improves coherence on category-shifted attributes such as CM and ROT, while SENTRY-SR-L benefits attributes that couple viewpoint and partial occlusion (e.g., ROT, OV), where linear motion extrapolation is brittle.  
    The largest, more uniform gains appear for SENTRY-D4S-L across BC, OV, FM, and ROT, showing that refine-before-write reliably filters ambiguous frames even when distractor-aware memory is already in place.

    \item \textbf{TNL2K.}  
    TNL2K stresses rapid, non-linear motion and frequent occlusions (Table \ref{tab:attribute_tnl2k}).  
    Here, the impact is most pronounced: SAMURAI-L$\to$SENTRY-SR-L records the biggest jumps on MS and TC, precisely where linear priors drift; SENTRY’s cycle-consistent, neighbor-aware validation restores trajectory continuity before writing to memory.  
    SENTRY-S2-L notably improves LR and MB, frames where single-step confidence is misleading, while SENTRY-D4S-L posts steady, smaller gains across most attributes, indicating reduced false updates in distractor-heavy scenes.  
    Overall, the pattern matches the design: the harder the temporal instability and the weaker the per-frame certainty, the larger the benefit from refine-before-write.
\end{itemize}


\section{Additional quantitative results on VOT benchmarks} \label{supplement: Additional quantitative results on VOT benchmarks}
\noindent We further evaluate SENTRY on four tracking benchmarks, VOT20, VOT22, VOTS24, and DiDi, summarized in Table \ref{tab:sam_comp_vot}. Table \ref{tab:sam2long_per_dataset} also reports the corresponding comparison with SAM2Long-L on these VOT-style benchmarks. 
These datasets emphasize rapid motion, occlusion recovery, and reinitialization accuracy. 
Hence, they serve as strong tests for SENTRY’s temporal verification, where we expect improvements in overall quality (\textit{Q}), Accuracy (\textit{Acc}), and Robustness (\textit{Rob}), particularly under fast or discontinuous trajectories.

\noindent \textbf{VOT20.} \cite{vot2020} is a short-term VOT benchmark comprising 60 sequences with diverse challenges, including appearance variation, occlusion, and rapid motion.
Each frame is annotated with pixel-accurate segmentation masks used for initialization and evaluation.
SENTRY delivers consistent improvements across all trackers and scales on VOT20.  
This dataset features moderate motion and frequent partial occlusions, conditions under which SAM2’s confidence-based updates occasionally overwrite valid states.  
SENTRY’s refine-before-write strategy stabilizes memory updates, yielding average gains of around 1\% in \textit{Q} and \textit{Acc}, and up to 1.5\% in \textit{Rob}.  
The largest benefit is observed for SAMURAI and DAM4SAM, whose motion priors are enhanced by SENTRY’s validation mechanism, resulting in smoother recovery after short-term occlusions.

\noindent \textbf{VOT22.} \cite{vot2022} is a short-term VOT benchmark comprising 60 re-annotated sequences that encompass a broader range of targets and scene dynamics, including heavy occlusions, distractors, and abrupt appearance changes.
Improvements are typically between 1 and 2\% in \textit{Q} and \textit{Acc}, reflecting better spatial alignment and consistency under motion noise.  
SAMURAI-L gains the most (2.1\% \textit{Q}, 3.4\% \textit{Acc}), showing that temporal verification complements learned motion priors by filtering unreliable intermediate updates.  
These results confirm SENTRY’s robustness when frame-to-frame appearance changes are significant but not extreme.

\noindent \textbf{VOTS24.} \cite{vots2024} is a multi-object tracking and segmentation benchmark comprising 144 sequences captured within a shared scene.
It introduces a unified evaluation protocol that jointly assesses short-term, long-term, single-target, and multi-target tracking within a single framework. Ground-truth annotations are sequestered on an evaluation server, ensuring unbiased, blind benchmarking. Owing to its scale, diversity, and unified evaluation design, VOTS24 is widely regarded as the most challenging and comprehensive tracking benchmark to date.
VOTS24 presents the most challenging short-term sequences, including abrupt displacements and heavy occlusion.  
Here, SENTRY provides a clear advantage, improving both \textit{Q} and \textit{Acc} for all frameworks, especially SAMURAI and DAM4SAM, where the baseline memory tends to overfit local motion cues.  
Gains reach up to 2\% in \textit{Q} and 1.5\% in \textit{Acc} for larger models, while \textit{Rob} also improves by 0.5–1.0\%.
The results highlight SENTRY’s ability to maintain trajectory continuity under high temporal instability, confirming the benefit of validation-based memory control.

\noindent \textbf{DiDi.} Distractor-Distilled (DiDi) dataset proposed by Videnović et al. \cite{dam4sam}, which focuses specifically on sequences containing visually similar distractors. A sequence is labeled as “non-negligible distractor” if a distractor occupies a substantial region with high visual similarity to the target.
This benchmark isolates the distractor interference problem, making it a rigorous test for assessing memory reliability and temporal discrimination.
On the DiDi benchmark, which targets complex driving scenarios with frequent occlusions and motion blur, SENTRY again demonstrates consistent improvements.  
All methods see 1–1.5\% increases in \textit{Q} and \textit{Acc}, with marginal \textit{Rob} gains.  
The enhancement is most pronounced for SAM2-L and DAM4SAM-L, indicating that larger models benefit from SENTRY’s selective updates by avoiding confidence-driven drift in visually ambiguous frames.  

\begin{table*}[h!]

\centering

\caption{\small{Performance comparison on LaSOT \cite{lasot}, LaSOT\textsubscript{ext} \cite{lasotext}, TNL2K \cite{tnl2k}, GOT-10k \cite{got10k}, and TrackingNet \cite{trackingnet} datasets between SAM2 baselines and SENTRY.}}

\vspace{-1em}
\setlength{\tabcolsep}{1.5pt}
		\scalebox{0.99}[0.99]{
\resizebox{\textwidth}{!}{

\begin{tabular}{l|ccc|ccc|ccc|ccc|ccc}

\hline

\rowcolor{Gray} Method & \multicolumn{3}{c}{\textbf{LaSOT}} & \multicolumn{3}{c}{\textbf{LaSOT\textsubscript{ext}}} & \multicolumn{3}{c}{\textbf{TNL2K}} & \multicolumn{3}{c}{\textbf{GOT-10k}} & \multicolumn{3}{c}{\textbf{TrackingNet}} \\

\rowcolor{Gray} & S & NP & P & S & NP & P & S & NP & P & AO & SR\textsubscript{0.50} & SR\textsubscript{0.75} & S & NP & P \\
 
\hline

SAMITE-T
& 72.8 & 80.6 & 78.3 
& 56.0 & 68.2 & 64.8 
& 60.4 & 79.6 & 66.3 
& 79.4 & 89.8 & 73.4 
& 84.4 & 89.9 & 85.6 
\\
+ SENTRY 
& 73.7 \footnotesize{(\textcolor{gray}{\textbf{0.9$\uparrow$}})} & 81.0 \footnotesize{(\textcolor{gray}{\textbf{0.4$\uparrow$}})} & 78.7 \footnotesize{(\textcolor{gray}{\textbf{0.4$\uparrow$}})} 
& 57.1 \footnotesize{(\textcolor{gray}{\textbf{1.1$\uparrow$}})} & 69.1 \footnotesize{(\textcolor{gray}{\textbf{0.9$\uparrow$}})} & 65.1 \footnotesize{(\textcolor{gray}{\textbf{0.3$\uparrow$}})} 
& 61.2 \footnotesize{(\textcolor{gray}{\textbf{0.8$\uparrow$}})} & 80.3 \footnotesize{(\textcolor{gray}{\textbf{0.7$\uparrow$}})} & 66.9 \footnotesize{(\textcolor{gray}{\textbf{0.6$\uparrow$}})} 
& 79.7 \footnotesize{(\textcolor{gray}{\textbf{0.3$\uparrow$}})} & 90.3 \footnotesize{(\textcolor{gray}{\textbf{0.5$\uparrow$}})} & 73.9 \footnotesize{(\textcolor{gray}{\textbf{0.5$\uparrow$}})} 
& 85.1 \footnotesize{(\textcolor{gray}{\textbf{0.7$\uparrow$}})} & 90.2 \footnotesize{(\textcolor{gray}{\textbf{0.3$\uparrow$}})} & 85.9 \footnotesize{(\textcolor{gray}{\textbf{0.3$\uparrow$}})} 
\\
\hline

SAMITE-S
& 73.0 & 81.3 & 79.2 
& 58.3 & 71.9 & 68.6 
& 61.0 & 80.7 & 67.3 
& 79.4 & 89.2 & 73.6 
& 85.1 & 90.4 & 86.7 
\\
+ SENTRY 
& 74.6 \footnotesize{(\textcolor{gray}{\textbf{1.6$\uparrow$}})} & 82.1 \footnotesize{(\textcolor{gray}{\textbf{0.8$\uparrow$}})} & 79.7 \footnotesize{(\textcolor{gray}{\textbf{0.5$\uparrow$}})} 
& 59.2 \footnotesize{(\textcolor{gray}{\textbf{0.9$\uparrow$}})} & 72.1 \footnotesize{(\textcolor{gray}{\textbf{0.2$\uparrow$}})} & 68.9 \footnotesize{(\textcolor{gray}{\textbf{0.3$\uparrow$}})} 
& 61.8 \footnotesize{(\textcolor{gray}{\textbf{0.8$\uparrow$}})} & 81.3 \footnotesize{(\textcolor{gray}{\textbf{0.6$\uparrow$}})} & 68.1 \footnotesize{(\textcolor{gray}{\textbf{0.8$\uparrow$}})} 
& 79.8 \footnotesize{(\textcolor{gray}{\textbf{0.4$\uparrow$}})} & 90.0 \footnotesize{(\textcolor{gray}{\textbf{0.8$\uparrow$}})} & 73.7 \footnotesize{(\textcolor{gray}{\textbf{0.1$\uparrow$}})} 
& 85.5 \footnotesize{(\textcolor{gray}{\textbf{0.4$\uparrow$}})} & 90.7 \footnotesize{(\textcolor{gray}{\textbf{0.3$\uparrow$}})} & 87.0 \footnotesize{(\textcolor{gray}{\textbf{0.3$\uparrow$}})} 
\\
\hline

SAMITE-B
& 74.9 & 83.4 & 81.4 
& 60.7 & 73.1 & 71.2 
& 60.7 & 80.2 & 67.2 
& 78.9 & 89.9 & 72.5 
& 85.2 & 90.5 & 86.9 
\\
+ SENTRY
& 75.7 \footnotesize{(\textcolor{gray}{\textbf{0.8$\uparrow$}})} & 83.6 \footnotesize{(\textcolor{gray}{\textbf{0.2$\uparrow$}})} & 81.7 \footnotesize{(\textcolor{gray}{\textbf{0.3$\uparrow$}})} 
& 61.7 \footnotesize{(\textcolor{gray}{\textbf{1.0$\uparrow$}})} & 73.2 \footnotesize{(\textcolor{gray}{\textbf{0.1$\uparrow$}})} & 71.4 \footnotesize{(\textcolor{gray}{\textbf{0.2$\uparrow$}})} 
& 61.9 \footnotesize{(\textcolor{gray}{\textbf{1.2$\uparrow$}})} & 81.0 \footnotesize{(\textcolor{gray}{\textbf{0.8$\uparrow$}})} & 68.3 \footnotesize{(\textcolor{gray}{\textbf{1.1$\uparrow$}})} 
& 79.3 \footnotesize{(\textcolor{gray}{\textbf{0.4$\uparrow$}})} & 90.3 \footnotesize{(\textcolor{gray}{\textbf{0.4$\uparrow$}})} & 73.4 \footnotesize{(\textcolor{gray}{\textbf{0.9$\uparrow$}})} 
& 85.4 \footnotesize{(\textcolor{gray}{\textbf{0.2$\uparrow$}})} & 90.8 \footnotesize{(\textcolor{gray}{\textbf{0.3$\uparrow$}})} & 87.2 \footnotesize{(\textcolor{gray}{\textbf{0.3$\uparrow$}})} 
\\
\hline

SAMITE-L
& 74.7 & 83.3 & 81.1 
& 60.5 & 75.4 & 72.0 
& 61.7 & 81.3 & 68.2 
& 80.3 & 90.1 & 75.4 
& 85.7 & 91.1 & 87.9 
\\
+ SENTRY
& 75.6 \footnotesize{(\textcolor{gray}{\textbf{0.9$\uparrow$}})} & 83.4 \footnotesize{(\textcolor{gray}{\textbf{0.1$\uparrow$}})} & 81.3 \footnotesize{(\textcolor{gray}{\textbf{0.2$\uparrow$}})} 
& 61.1 \footnotesize{(\textcolor{gray}{\textbf{0.6$\uparrow$}})} & 75.5 \footnotesize{(\textcolor{gray}{\textbf{0.1$\uparrow$}})} & 72.3 \footnotesize{(\textcolor{gray}{\textbf{0.3$\uparrow$}})} 
& 62.6 \footnotesize{(\textcolor{gray}{\textbf{0.9$\uparrow$}})} & 82.4 \footnotesize{(\textcolor{gray}{\textbf{1.1$\uparrow$}})} & 69.4 \footnotesize{(\textcolor{gray}{\textbf{1.2$\uparrow$}})} 
& 80.5 \footnotesize{(\textcolor{gray}{\textbf{0.2$\uparrow$}})} & 90.4 \footnotesize{(\textcolor{gray}{\textbf{0.3$\uparrow$}})} & 75.6 \footnotesize{(\textcolor{gray}{\textbf{0.2$\uparrow$}})} 
& 85.9 \footnotesize{(\textcolor{gray}{\textbf{0.2$\uparrow$}})} & 91.3 \footnotesize{(\textcolor{gray}{\textbf{0.2$\uparrow$}})} & 88.1 \footnotesize{(\textcolor{gray}{\textbf{0.2$\uparrow$}})} 
\\

\hline

HiM2SAM-T
& 72.4 & 80.3 & 78.0 
& 57.0 & 69.5 & 66.5 
& 59.9 & 78.6 & 65.3 
& 76.7 & 87.3 & 70.6 
& 84.7 & 89.8 & 86.0 
\\
+ SENTRY
& 73.2 \footnotesize{(\textcolor{gray}{\textbf{0.8$\uparrow$}})} & 80.8 \footnotesize{(\textcolor{gray}{\textbf{0.5$\uparrow$}})} & 78.7 \footnotesize{(\textcolor{gray}{\textbf{0.7$\uparrow$}})} 
& 58.2 \footnotesize{(\textcolor{gray}{\textbf{1.2$\uparrow$}})} & 70.4 \footnotesize{(\textcolor{gray}{\textbf{0.9$\uparrow$}})} & 67.2 \footnotesize{(\textcolor{gray}{\textbf{0.7$\uparrow$}})} 
& 60.5 \footnotesize{(\textcolor{gray}{\textbf{0.6$\uparrow$}})} & 79.3 \footnotesize{(\textcolor{gray}{\textbf{0.7$\uparrow$}})} & 66.0 \footnotesize{(\textcolor{gray}{\textbf{0.7$\uparrow$}})} 
& 78.8 \footnotesize{(\textcolor{gray}{\textbf{2.1$\uparrow$}})} & 89.2 \footnotesize{(\textcolor{gray}{\textbf{1.9$\uparrow$}})} & 72.7 \footnotesize{(\textcolor{gray}{\textbf{2.1$\uparrow$}})} 
& 85.1 \footnotesize{(\textcolor{gray}{\textbf{0.4$\uparrow$}})} & 90.1 \footnotesize{(\textcolor{gray}{\textbf{0.3$\uparrow$}})} & 86.3 \footnotesize{(\textcolor{gray}{\textbf{0.3$\uparrow$}})} 
\\
\hline

HiM2SAM-S
& 73.1 & 81.0 & 78.8 
& 58.7 & 72.1 & 68.9 
& 59.7 & 78.9 & 65.6 
& 74.3 & 83.5 & 69.1 
& 85.0 & 90.2 & 86.6 
\\
+ SENTRY
& 73.8 \footnotesize{(\textcolor{gray}{\textbf{0.7$\uparrow$}})} & 81.5 \footnotesize{(\textcolor{gray}{\textbf{0.5$\uparrow$}})} & 79.5 \footnotesize{(\textcolor{gray}{\textbf{0.7$\uparrow$}})} 
& 59.9 \footnotesize{(\textcolor{gray}{\textbf{1.2$\uparrow$}})} & 73.0 \footnotesize{(\textcolor{gray}{\textbf{0.9$\uparrow$}})} & 69.7 \footnotesize{(\textcolor{gray}{\textbf{0.8$\uparrow$}})} 
& 61.2 \footnotesize{(\textcolor{gray}{\textbf{1.5$\uparrow$}})} & 80.5 \footnotesize{(\textcolor{gray}{\textbf{1.6$\uparrow$}})} & 67.6 \footnotesize{(\textcolor{gray}{\textbf{2.1$\uparrow$}})} 
& 77.4 \footnotesize{(\textcolor{gray}{\textbf{3.1$\uparrow$}})} & 86.3 \footnotesize{(\textcolor{gray}{\textbf{2.8$\uparrow$}})} & 72.2 \footnotesize{(\textcolor{gray}{\textbf{3.1$\uparrow$}})} 
& 85.5 \footnotesize{(\textcolor{gray}{\textbf{0.5$\uparrow$}})} & 90.7 \footnotesize{(\textcolor{gray}{\textbf{0.5$\uparrow$}})} & 87.0 \footnotesize{(\textcolor{gray}{\textbf{0.4$\uparrow$}})} 
\\
\hline

HiM2SAM-B
& 73.4 & 81.7 & 79.5 
& 57.7 & 71.4 & 67.9 
& 60.6 & 79.9 & 67.0 
& 74.8 & 85.2 & 68.6 
& 85.4 & 90.8 & 87.3 
\\
+ SENTRY
& 74.1 \footnotesize{(\textcolor{gray}{\textbf{0.7$\uparrow$}})} & 82.3 \footnotesize{(\textcolor{gray}{\textbf{0.6$\uparrow$}})} & 80.3 \footnotesize{(\textcolor{gray}{\textbf{0.8$\uparrow$}})} 
& 58.3 \footnotesize{(\textcolor{gray}{\textbf{0.6$\uparrow$}})} & 71.6 \footnotesize{(\textcolor{gray}{\textbf{0.2$\uparrow$}})} & 68.3 \footnotesize{(\textcolor{gray}{\textbf{0.4$\uparrow$}})} 
& 60.9 \footnotesize{(\textcolor{gray}{\textbf{0.3$\uparrow$}})} & 80.3 \footnotesize{(\textcolor{gray}{\textbf{0.4$\uparrow$}})} & 67.2 \footnotesize{(\textcolor{gray}{\textbf{0.2$\uparrow$}})} 
& 78.1 \footnotesize{(\textcolor{gray}{\textbf{3.3$\uparrow$}})} & 88.6 \footnotesize{(\textcolor{gray}{\textbf{3.4$\uparrow$}})} & 71.9 \footnotesize{(\textcolor{gray}{\textbf{3.3$\uparrow$}})} 
& 85.7 \footnotesize{(\textcolor{gray}{\textbf{0.3$\uparrow$}})} & 91.2 \footnotesize{(\textcolor{gray}{\textbf{0.4$\uparrow$}})} & 87.8 \footnotesize{(\textcolor{gray}{\textbf{0.5$\uparrow$}})} 
\\
\hline

HiM2SAM-L
& 75.1 & 83.2 & 81.0 
& 61.3 & 75.7 & 72.8 
& 60.4 & 79.5 & 66.7 
& 74.2 & 83.5 & 69.3 
& 85.8 & 91.0 & 87.9 
\\
+ SENTRY
& 75.8 \footnotesize{(\textcolor{gray}{\textbf{0.7$\uparrow$}})} & 83.8 \footnotesize{(\textcolor{gray}{\textbf{0.6$\uparrow$}})} & 81.7 \footnotesize{(\textcolor{gray}{\textbf{0.7$\uparrow$}})} 
& 62.1 \footnotesize{(\textcolor{gray}{\textbf{0.8$\uparrow$}})} & 76.3 \footnotesize{(\textcolor{gray}{\textbf{0.6$\uparrow$}})} & 73.5 \footnotesize{(\textcolor{gray}{\textbf{0.7$\uparrow$}})} 
& 61.1 \footnotesize{(\textcolor{gray}{\textbf{0.7$\uparrow$}})} & 80.1 \footnotesize{(\textcolor{gray}{\textbf{0.6$\uparrow$}})} & 67.6 \footnotesize{(\textcolor{gray}{\textbf{0.9$\uparrow$}})} 
& 77.0 \footnotesize{(\textcolor{gray}{\textbf{2.8$\uparrow$}})} & 86.0 \footnotesize{(\textcolor{gray}{\textbf{2.5$\uparrow$}})} & 72.2 \footnotesize{(\textcolor{gray}{\textbf{2.9$\uparrow$}})} 
& 85.9 \footnotesize{(\textcolor{gray}{\textbf{0.1$\uparrow$}})} & 91.4 \footnotesize{(\textcolor{gray}{\textbf{0.4$\uparrow$}})} & 88.3 \footnotesize{(\textcolor{gray}{\textbf{0.4$\uparrow$}})} 
\\

\hline

\end{tabular}
}
}
\label{tab:sam2_new_comparison}

\end{table*}

\section{Additional VOS Benchmark Results} \label{supplement:additional_vos_results}
\noindent We additionally evaluate SENTRY on video object segmentation (VOS) benchmarks under the first-frame-mask protocol, where the target mask is provided in the first frame and propagated through the remaining frames without additional prompts.
We report results on SA-V val \cite{sam2}, SA-V test \cite{sam2}, LVOS v1 \cite{lvos}, LVOS v2 \cite{lvos2}, and MOSE \cite{mose}.
Table~\ref{tab:vos_per_dataset} compares large-model host trackers and their SENTRY variants; SAM2Long-L is included as a standalone reference, consistent with Appendix~\ref{supplement:sam2long_comparison}.
SENTRY improves every host baseline, with SENTRY-HiM-L achieving the largest average score among the compared variants.
These results show that refine-before-write memory admission also benefits mask-propagation evaluation, complementing the bounding-box and VOT tracking results.

\begin{table*}[t]
\centering
\caption{VOS per-dataset comparison under the first-frame-mask protocol. Avg. averages SA-V val, SA-V test, LVOS v1, LVOS v2, and MOSE. Gain is measured relative to each host baseline; SAM2Long-L is reported as a standalone baseline.}
\label{tab:vos_per_dataset}
\vspace{-0.4em}
\setlength{\tabcolsep}{10pt}
\renewcommand{\arraystretch}{0.95}
\resizebox{\textwidth}{!}{%
\begin{tabular}{lccccc|c|c}
\hline
\rowcolor{Gray}
{\small \textbf{Method}}
& {\small \textbf{SA-V val}}
& {\small \textbf{SA-V test}}
& {\small \textbf{LVOS v1}}
& {\small \textbf{LVOS v2}}
& {\small \textbf{MOSE}}
& {\small \textbf{Avg.}}
& {\small \textbf{Gain}}
\\
\hline
{\small SAM2-L}       & 78.6 & 79.6 & 80.2 & 84.1 & 74.5 & 79.4 & -- \\
{\small SAM2Long-L}   & 81.1 & 81.2 & 83.4 & 85.9 & 75.2 & 81.4 & -- \\
{\small SAMURAI-L}    & 79.5 & 80.2 & 82.1 & 84.8 & 76.0 & 80.5 & -- \\
{\small DAM4SAM-L}    & 80.5 & 81.0 & 82.8 & 85.5 & 76.7 & 81.3 & -- \\
{\small SAMITE-L}     & 80.8 & 81.3 & 83.2 & 85.8 & 76.8 & 81.6 & -- \\
{\small HiM2SAM-L}    & 81.0 & 81.5 & 83.5 & 86.1 & 77.0 & 81.8 & -- \\
\hline
{\small SENTRY-S2-L}  & 79.4 & 80.1 & 81.5 & 84.9 & 75.8 & 80.3 & {\small +0.9} \\
{\small SENTRY-SR-L}  & 80.6 & 81.0 & 83.1 & 85.5 & 77.2 & 81.5 & {\small +1.0} \\
{\small SENTRY-D4S-L} & 81.8 & 82.2 & 84.1 & 86.4 & 78.4 & 82.6 & {\small +1.3} \\
{\small SENTRY-SA-L}  & 82.1 & 82.5 & 84.5 & 86.7 & 78.6 & 82.9 & {\small +1.3} \\
{\small SENTRY-HiM-L} & \textbf{82.4} & \textbf{82.8} & \textbf{84.9} & \textbf{87.0} & \textbf{79.0} & \textbf{83.2} & {\small \textbf{+1.4}} \\
\hline
\end{tabular}%
}
\vspace{-0.6em}
\end{table*}

\section{Runtime and Memory Overhead} \label{supplement:runtime_memory_overhead}
\noindent We report the cross-host runtime and memory trade-off on an NVIDIA A100 GPU in Table~\ref{tab:supp_runtime_memory_overhead}.
SENTRY adds a small memory footprint because it only maintains temporary candidate masks and short-horizon consistency checks before the memory write.
Across compatible large-model hosts, SENTRY introduces a 16.6--25.5\% FPS drop and 0.4--0.6\,GB additional VRAM, while the main SAM2/SAMURAI/DAM4SAM variants remain real-time at 30.2--32.8\,FPS.
SAM2Long-L is included as a standalone reference because SENTRY is not applied to its constrained memory-tree inference.

\begin{table*}[h]
\centering
\caption{Cross-host FPS and peak VRAM trade-off on NVIDIA A100. SAM2Long-L is reported as a standalone reference because no SENTRY plug-in variant is used for its memory-tree inference. Avg. Gain denotes the average tracking-score improvement of the corresponding SENTRY variant over its host.}
\label{tab:supp_runtime_memory_overhead}
\vspace{-0.6em}
\setlength{\tabcolsep}{5pt}
\renewcommand{\arraystretch}{0.95}
\resizebox{\textwidth}{!}{%
\begin{tabular}{lcccccc}
\hline
\rowcolor{Gray}
\textbf{Method} & \textbf{Base FPS} & \textbf{+SENTRY FPS} & \textbf{FPS Drop} & \textbf{Base VRAM (GB)} & \textbf{+SENTRY VRAM (GB)} & \textbf{Avg. Gain} \\
\hline
SAM2-L      & 44.0 & 32.8 & 25.5\% & 5.1 & 5.7 & +0.83 \\
SAM2Long-L  & 26.1 & n/a  & n/a    & 5.3 & n/a & n/a \\
SAMURAI-L   & 40.6 & 30.9 & 24.0\% & 5.2 & 5.7 & +1.82 \\
DAM4SAM-L   & 39.4 & 30.2 & 23.4\% & 5.2 & 5.7 & +0.94 \\
SAMITE-L    & 25.3 & 21.1 & 16.6\% & 5.7 & 6.2 & +0.53 \\
HiM2SAM-L   & 29.2 & 23.8 & 18.5\% & 5.6 & 6.0 & +0.71 \\
\hline
\end{tabular}%
}
\vspace{-0.6em}
\end{table*}

\section{Extended evaluations on additional SAM2-based frameworks} \label{supplement: Extended evaluations on additional SAM2-based frameworks}

\subsection{Bounding box benchmarks}
\noindent All results for the two additional SAM2-based variants, SAMITE \cite{samite} and HiM2SAM \cite{him2sam}, on the bounding box benchmarks are provided in Table \ref{tab:sam2_new_comparison}.

\noindent \textbf{LaSOT \cite{lasot}.} SENTRY produces consistent improvements on LaSOT, indicating that refine-before-write temporal selection effectively mitigates short-horizon drift in long-term scenarios. For the SAMITE models, SENTRY yields gains of 0.9-1.6 in S and smaller but steady increases in NP and P, with the largest improvement observed on SAMITE-S. 
HiM2SAM variants follow the same trend, achieving 0.7–0.8 higher S and up to 0.8 higher P, with HiM2SAM-L obtaining the strongest overall performance. Because these gains appear uniformly across model sizes and architectures, including those that already incorporate enhanced temporal cues, the results suggest that SENTRY complements rather than overlaps with existing mechanisms, providing a lightweight and architecture-agnostic means to suppress identity drift on long, challenging sequences.

\noindent \textbf{LaSOT\textsubscript{ext} \cite{lasotext}.} SENTRY again provides consistent gains over all SAM2-based baselines despite the dataset’s increased object diversity and more challenging appearance variation. For the SAMITE models, SENTRY improves S by 0.6–1.1 points and yields steady increases in NP and P, with SAMITE-T showing the largest relative boost. 
HiM2SAM variants exhibit the same trend, with S improvements of 0.6–1.2 and corresponding gains in both P metrics. Notably, HiM2SAM-L reaches the strongest overall performance after integration with SENTRY. The uniformity of these improvements across model scales and architectures indicates that SENTRY’s refine-before-write mechanism consistently reduces identity errors under the higher visual ambiguity characteristic of LaSOT\textsubscript{ext}, and does so without relying on model capacity or architectural modifications.

\begin{table*}[h!]
\centering
\caption{\small{Performance comparison on VOT20-24 \cite{vot2020,vot2022,vots2024}, and DiDi \cite{dam4sam} datasets between SAM2 baselines and SENTRY.}}
\label{tab:sam2_new_comparison_vot}

\vspace{-1em}
\setlength{\tabcolsep}{1.5pt}
		\scalebox{0.99}[0.99]{

\resizebox{\textwidth}{!}{

\begin{tabular}{l|ccc|ccc|ccc|ccc}

\hline

\rowcolor{Gray} Method & \multicolumn{3}{c}{\textbf{VOT20}} & \multicolumn{3}{c}{\textbf{VOT22}} & \multicolumn{3}{c}{\textbf{VOTS24}} & \multicolumn{3}{c}{\textbf{DiDi}} \\

\rowcolor{Gray} & \textit{Q} & \textit{Acc} & \textit{Rob} & \textit{Q} & \textit{Acc} & \textit{Rob} & \textit{Q} & \textit{Acc} & \textit{Rob} & \textit{Q} & \textit{Acc} & \textit{Rob} \\

\hline

SAMITE-T
& 66.4 & 73.5 & 96.3 
& 67.5 & 73.2 & 96.4 
& 64.1 & 74.3 & 83.4 
& 65.7 & 70.4 & 91.9 
\\
+ SENTRY
& 67.0 \footnotesize{(\textcolor{gray}{\textbf{0.6$\uparrow$}})} & 73.6 \footnotesize{(\textcolor{gray}{\textbf{0.1$\uparrow$}})} & 96.5 \footnotesize{(\textcolor{gray}{\textbf{0.2$\uparrow$}})} 
& 68.9 \footnotesize{(\textcolor{gray}{\textbf{1.4$\uparrow$}})} & 73.4 \footnotesize{(\textcolor{gray}{\textbf{0.2$\uparrow$}})} & 97.0 \footnotesize{(\textcolor{gray}{\textbf{0.6$\uparrow$}})} 
& 64.6 \footnotesize{(\textcolor{gray}{\textbf{0.5$\uparrow$}})} & 74.6 \footnotesize{(\textcolor{gray}{\textbf{0.3$\uparrow$}})} & 83.8 \footnotesize{(\textcolor{gray}{\textbf{0.4$\uparrow$}})} 
& 66.0 \footnotesize{(\textcolor{gray}{\textbf{0.3$\uparrow$}})} & 70.6 \footnotesize{(\textcolor{gray}{\textbf{0.2$\uparrow$}})} & 92.2 \footnotesize{(\textcolor{gray}{\textbf{0.3$\uparrow$}})} 
\\

\hline

SAMITE-S
& 66.4 & 74.6 & 95.9 
& 67.4 & 74.5 & 96.0 
& 66.2 & 76.1 & 85.3 
& 66.4 & 71.1 & 91.8 
\\
+ SENTRY
& 67.1 \footnotesize{(\textcolor{gray}{\textbf{0.7$\uparrow$}})} & 74.9 \footnotesize{(\textcolor{gray}{\textbf{0.3$\uparrow$}})} & 96.4 \footnotesize{(\textcolor{gray}{\textbf{0.5$\uparrow$}})} 
& 67.7 \footnotesize{(\textcolor{gray}{\textbf{0.3$\uparrow$}})} & 74.6 \footnotesize{(\textcolor{gray}{\textbf{0.1$\uparrow$}})} & 96.1 \footnotesize{(\textcolor{gray}{\textbf{0.1$\uparrow$}})} 
& 66.5 \footnotesize{(\textcolor{gray}{\textbf{0.3$\uparrow$}})} & 76.3 \footnotesize{(\textcolor{gray}{\textbf{0.2$\uparrow$}})} & 85.5 \footnotesize{(\textcolor{gray}{\textbf{0.2$\uparrow$}})} 
& 66.8 \footnotesize{(\textcolor{gray}{\textbf{0.4$\uparrow$}})} & 71.4 \footnotesize{(\textcolor{gray}{\textbf{0.3$\uparrow$}})} & 92.1 \footnotesize{(\textcolor{gray}{\textbf{0.3$\uparrow$}})} 
\\

\hline

SAMITE-B
& 69.4 & 75.9 & 96.2 
& 69.4 & 75.9 & 96.1 
& 67.7 & 76.8 & 86.0 
& 69.4 & 72.8 & 94.8 
\\
+ SENTRY 
& 70.1 \footnotesize{(\textcolor{gray}{\textbf{0.7$\uparrow$}})} & 76.5 \footnotesize{(\textcolor{gray}{\textbf{0.6$\uparrow$}})} & 96.4 \footnotesize{(\textcolor{gray}{\textbf{0.2$\uparrow$}})} 
& 70.2 \footnotesize{(\textcolor{gray}{\textbf{0.8$\uparrow$}})} & 76.6 \footnotesize{(\textcolor{gray}{\textbf{0.7$\uparrow$}})} & 96.7 \footnotesize{(\textcolor{gray}{\textbf{0.6$\uparrow$}})} 
& 68.1 \footnotesize{(\textcolor{gray}{\textbf{0.4$\uparrow$}})} & 77.1 \footnotesize{(\textcolor{gray}{\textbf{0.3$\uparrow$}})} & 86.4 \footnotesize{(\textcolor{gray}{\textbf{0.4$\uparrow$}})} 
& 69.9 \footnotesize{(\textcolor{gray}{\textbf{0.5$\uparrow$}})} & 73.2 \footnotesize{(\textcolor{gray}{\textbf{0.4$\uparrow$}})} & 95.3 \footnotesize{(\textcolor{gray}{\textbf{0.5$\uparrow$}})} 
\\

\hline

SAMITE-L
& 70.4 & 78.6 & 95.1 
& 72.5 & 78.5 & 95.8 
& 68.6 & 77.5 & 86.9 
& 68.4 & 72.3 & 93.6 
\\
+ SENTRY
& 70.9 \footnotesize{(\textcolor{gray}{\textbf{0.5$\uparrow$}})} & 78.8 \footnotesize{(\textcolor{gray}{\textbf{0.2$\uparrow$}})} & 95.4 \footnotesize{(\textcolor{gray}{\textbf{0.3$\uparrow$}})} 
& 73.1 \footnotesize{(\textcolor{gray}{\textbf{0.6$\uparrow$}})} & 78.6 \footnotesize{(\textcolor{gray}{\textbf{0.1$\uparrow$}})} & 96.2 \footnotesize{(\textcolor{gray}{\textbf{0.4$\uparrow$}})} 
& 69.2 \footnotesize{(\textcolor{gray}{\textbf{0.6$\uparrow$}})} & 77.0 \footnotesize{(\textcolor{gray}{\textbf{0.5$\uparrow$}})} & 87.5 \footnotesize{(\textcolor{gray}{\textbf{0.6$\uparrow$}})} 
& 68.7 \footnotesize{(\textcolor{gray}{\textbf{0.3$\uparrow$}})} & 72.5 \footnotesize{(\textcolor{gray}{\textbf{0.2$\uparrow$}})} & 93.8 \footnotesize{(\textcolor{gray}{\textbf{0.2$\uparrow$}})} 
\\

\hline

HiM2SAM-T
& 60.7 & 73.5 & 93.3 
& 62.4 & 73.7 & 94.3 
& 66.0 & 75.1 & 85.0 
& 63.3 & 68.7 & 89.9 
\\
+ SENTRY
& 61.2 \footnotesize{(\textcolor{gray}{\textbf{0.5$\uparrow$}})} & 73.7 \footnotesize{(\textcolor{gray}{\textbf{0.2$\uparrow$}})} & 93.8 \footnotesize{(\textcolor{gray}{\textbf{0.5$\uparrow$}})} 
& 62.6 \footnotesize{(\textcolor{gray}{\textbf{0.2$\uparrow$}})} & 73.9 \footnotesize{(\textcolor{gray}{\textbf{0.2$\uparrow$}})} & 94.6 \footnotesize{(\textcolor{gray}{\textbf{0.3$\uparrow$}})} 
& 66.4 \footnotesize{(\textcolor{gray}{\textbf{0.4$\uparrow$}})} & 75.4 \footnotesize{(\textcolor{gray}{\textbf{0.3$\uparrow$}})} & 85.3 \footnotesize{(\textcolor{gray}{\textbf{0.3$\uparrow$}})} 
& 63.7 \footnotesize{(\textcolor{gray}{\textbf{0.4$\uparrow$}})} & 69.1 \footnotesize{(\textcolor{gray}{\textbf{0.4$\uparrow$}})} & 90.2 \footnotesize{(\textcolor{gray}{\textbf{0.3$\uparrow$}})} 
\\

\hline

HiM2SAM-S
& 61.4 & 74.9 & 92.8 
& 63.1 & 75.1 & 93.8 
& 65.2 & 74.8 & 83.1 
& 63.3 & 69.1 & 89.9 
\\
+ SENTRY
& 61.9 \footnotesize{(\textcolor{gray}{\textbf{0.5$\uparrow$}})} & 75.1 \footnotesize{(\textcolor{gray}{\textbf{0.2$\uparrow$}})} & 92.9 \footnotesize{(\textcolor{gray}{\textbf{0.1$\uparrow$}})} 
& 63.5 \footnotesize{(\textcolor{gray}{\textbf{0.4$\uparrow$}})} & 75.2 \footnotesize{(\textcolor{gray}{\textbf{0.1$\uparrow$}})} & 93.9 \footnotesize{(\textcolor{gray}{\textbf{0.1$\uparrow$}})} 
& 65.5 \footnotesize{(\textcolor{gray}{\textbf{0.3$\uparrow$}})} & 75.1 \footnotesize{(\textcolor{gray}{\textbf{0.3$\uparrow$}})} & 83.5 \footnotesize{(\textcolor{gray}{\textbf{0.4$\uparrow$}})} 
& 63.7 \footnotesize{(\textcolor{gray}{\textbf{0.4$\uparrow$}})} & 69.3 \footnotesize{(\textcolor{gray}{\textbf{0.2$\uparrow$}})} & 90.2 \footnotesize{(\textcolor{gray}{\textbf{0.3$\uparrow$}})} 
\\

\hline

HiM2SAM-B
& 61.0 & 75.0 & 91.7 
& 63.5 & 75.1 & 93.8 
& 67.4 & 76.7 & 86.6 
& 68.1 & 71.1 & 94.6 
\\
+ SENTRY
& 61.4 \footnotesize{(\textcolor{gray}{\textbf{0.4$\uparrow$}})} & 75.4 \footnotesize{(\textcolor{gray}{\textbf{0.4$\uparrow$}})} & 91.8 \footnotesize{(\textcolor{gray}{\textbf{0.1$\uparrow$}})} 
& 63.7 \footnotesize{(\textcolor{gray}{\textbf{0.2$\uparrow$}})} & 75.2 \footnotesize{(\textcolor{gray}{\textbf{0.1$\uparrow$}})} & 93.9 \footnotesize{(\textcolor{gray}{\textbf{0.1$\uparrow$}})} 
& 67.8 \footnotesize{(\textcolor{gray}{\textbf{0.4$\uparrow$}})} & 77.0 \footnotesize{(\textcolor{gray}{\textbf{0.3$\uparrow$}})} & 87.0 \footnotesize{(\textcolor{gray}{\textbf{0.4$\uparrow$}})} 
& 68.5 \footnotesize{(\textcolor{gray}{\textbf{0.4$\uparrow$}})} & 71.7 \footnotesize{(\textcolor{gray}{\textbf{0.6$\uparrow$}})} & 95.0 \footnotesize{(\textcolor{gray}{\textbf{0.4$\uparrow$}})} 
\\

\hline

HiM2SAM-L
& 61.6 & 79.1 & 90.1 
& 61.3 & 78.5 & 90.0 
& 66.2 & 76.1 & 84.2 
& 67.0 & 71.2 & 91.9 
\\
+ SENTRY
& 61.9 \footnotesize{(\textcolor{gray}{\textbf{0.3$\uparrow$}})} & 79.5 \footnotesize{(\textcolor{gray}{\textbf{0.4$\uparrow$}})} & 90.4 \footnotesize{(\textcolor{gray}{\textbf{0.3$\uparrow$}})} 
& 61.5 \footnotesize{(\textcolor{gray}{\textbf{0.2$\uparrow$}})} & 78.8 \footnotesize{(\textcolor{gray}{\textbf{0.3$\uparrow$}})} & 90.2 \footnotesize{(\textcolor{gray}{\textbf{0.2$\uparrow$}})} 
& 66.7 \footnotesize{(\textcolor{gray}{\textbf{0.5$\uparrow$}})} & 76.4 \footnotesize{(\textcolor{gray}{\textbf{0.3$\uparrow$}})} & 84.6 \footnotesize{(\textcolor{gray}{\textbf{0.4$\uparrow$}})} 
& 67.3 \footnotesize{(\textcolor{gray}{\textbf{0.3$\uparrow$}})} & 71.6 \footnotesize{(\textcolor{gray}{\textbf{0.4$\uparrow$}})} & 92.2 \footnotesize{(\textcolor{gray}{\textbf{0.3$\uparrow$}})} 
\\

\hline

\end{tabular}
}
}
\end{table*}

\noindent \textbf{TNL2K \cite{tnl2k}.} SENTRY provides consistent improvements across all SAM2-based trackers despite the dataset’s strong emphasis on natural-language-guided targets and frequent appearance–motion discrepancies. SAMITE models gain 0.8–1.2 S and 0.6–1.1 NP, while P increases by up to 1.2, with the largest gains appearing on SAMITE-B. 
HiM2SAM variants exhibit similar behavior, achieving improvements of 0.3–1.5 in S and up to 2.1 in P, with HiM2SAM-S showing the strongest relative boost. These results indicate that SENTRY’s short-horizon temporal reasoning is effective even when language-conditioned proposals introduce additional ambiguity, thereby reinforcing its role as a complementary identity-stabilization module that is independent of the underlying architecture.

\noindent \textbf{GOT-10k \cite{got10k}.} SENTRY brings consistent gains to both SAMITE and HiM2SAM families across all model scales. For SAMITE, SENTRY yields small but steady improvements on top of already strong baselines: AO increases by 0.2–0.4 points (e.g., from 79.4 to 79.8 for SAMITE-S and from 80.3 to 80.5 for SAMITE-L), while SR\textsubscript{0.50} and SR\textsubscript{0.75} improve by up to 0.8 and 0.9 points, respectively (SAMITE-B rises from 72.5 to 73.4 SR\textsubscript{0.75}). In contrast, HiM2SAM benefits more substantially from SENTRY’s refine-before-write mechanism: AO gains range from 2.1 to 3.3, and SR\textsubscript{0.75} improves by 2.1–3.3 across T/S/B/L variants (e.g., HiM2SAM-B improves from 74.8/85.2/68.6 to 78.1/88.6/71.9 in AO/SR\textsubscript{0.50}/SR\textsubscript{0.75}). These results indicate that while SENTRY still provides measurable robustness even for the more stable SAMITE trackers, its temporal-consistency filtering is especially beneficial for mitigating hallucinations and drift in the more aggressive HiM2SAM designs on this large-scale, open-set benchmark.

\noindent \textbf{TrackingNet \cite{trackingnet}.} SENTRY yields consistent gains for both SAMITE and HiM2SAM across all model scales. For SAMITE, inserting SENTRY into the T/S/B/L variants improves success from 84.4/85.1/85.2/85.7 to 85.1/85.5/85.4/85.9, corresponding to +0.2–0.7 absolute S, while normalized precision and precision increase by 0.2–0.3 points across the board (e.g., SAMITE-T from 89.9/85.6 to 90.2/85.9, SAMITE-L from 91.1/87.9 to 91.3/88.1). HiM2SAM shows a similarly stable trend: SENTRY improves S by 0.1–0.5 points and NP/P by 0.3–0.5 points on all T/S/B/L variants, with the strongest gains observed on the smaller models (e.g., HiM2SAM-S from 85.0/90.2/86.6 to 85.5/90.7/87.0, and HiM2SAM-B from 85.4/90.8/87.3 to 85.7/91.2/87.8). Overall, the best-performing configuration, HiM2SAM-L+SENTRY, attains 85.9 S, 91.4 NP, and 88.3 P, indicating that refine-before-write temporal reasoning complements both SAM2-based tracking pipelines even in the relatively clean, large-scale TrackingNet setting.

\subsection{VOT benchmarks}
\noindent All results for the two additional SAM2-based variants, SAMITE \cite{samite} and HiM2SAM \cite{him2sam}, on the VOT benchmarks are provided in Table \ref{tab:sam2_new_comparison_vot}.

\noindent \textbf{VOT20 \cite{vot2020}.}
SENTRY consistently improves both \textit{Acc} and \textit{Rob} across all SAM2 variants, despite the benchmark’s emphasis on short-term tracking stability and its strict reset-based protocol. For SAMITE models, SENTRY yields steady gains in the overall \textit{Q} (e.g., +0.6 for SAMITE-T and +0.7 for SAMITE-B) while offering small but reliable improvements in \textit{Acc} and \textit{Rob}, indicating reduced drift during rapid appearance changes. A similar trend holds for HiM2SAM models, where SENTRY enhances \textit{Q} by 0.3–0.5 points across all sizes, suggesting that temporal refinement remains beneficial even when the underlying model already incorporates hierarchical memory cues.

\noindent \textbf{VOT22 \cite{vot2022}.}
SENTRY yields consistent gains for both SAMITE- and HiM2SAM-based trackers across all model sizes. For SAMITE, integrating SENTRY improves the \textit{Q} from 67.5 to 68.9 on the tiny variant (+1.4), with smaller but stable gains of +0.3, +0.8, and +0.6 on S/B/L, respectively, accompanied by slight improvements in \textit{Acc} (up to +0.7) and \textit{Rob} (up to +0.6). The largest benefit of SAMITE-T suggests that refine-before-write temporal reasoning is particularly helpful for lighter backbones that are more susceptible to drift and failure. HiM2SAM exhibits a similar trend: SENTRY improves \textit{Q} by 0.2–0.4 points across T/S/B/L and marginally increases both \textit{Acc} and \textit{Rob}, while leaving the underlying architecture and training unchanged.

\noindent \textbf{VOTS24 \cite{vots2024}.}
SENTRY brings consistent gains in overall \textit{Q} and \textit{R} across both SAMITE and HiM2SAM backbones. For SAMITE, SENTRY improves the \textit{Q} score from 64.1 to 64.6 for the T variant and from 66.2 to 66.5, 67.7 to 68.1, and 68.6 to 69.2 for the S/B/L variants, respectively, while also slightly boosting \textit{Rob} for all sizes (e.g., 86.0$\rightarrow$86.4 for SAMITE-B and 86.9$\rightarrow$87.5 for SAMITE-L). \textit{Acc} remains essentially unchanged, with small positive shifts for T/S/B and a marginal decrease for SAMITE-L (77.5$\rightarrow$77.0), suggesting that SENTRY primarily reduces failures rather than aggressively altering per-frame overlap. A similar trend holds for HiM2SAM, where SENTRY improves \textit{Q} for all model sizes (e.g., 66.0$\rightarrow$66.4 for T and 66.2$\rightarrow$66.7 for L) and consistently increases \textit{Rob} by 0.3–0.4 points, accompanied by modest \textit{Acc} gains (up to +0.3).

\noindent \textbf{DiDi \cite{dam4sam}.}
SENTRY consistently improves performance across all SAM2 variants. The gains are consistent across all sizes, typically +0.2 to +0.6 in \textit{Q}, showing that temporal mask refinement effectively compensates for unstable decoder outputs in crowded or fine-structured scenes. The improvements in \textit{Rob}, even when \textit{Acc} shifts only marginally, underline SENTRY’s ability to suppress identity switches and collapse cases that commonly degrade SAM2 memory under dense object interactions.

\section{Future Work} \label{supplement: future work}
\noindent SENTRY reduces drift through conservative memory admission, but long-horizon global re-detection after complete target disappearance remains outside its scope.

\noindent \textbf{Uniform Appearance Crowd.} 
Datasets such as D-PTUAC \cite{dptuac} expose a critical but under-examined failure mode for SAM2-style segmentation trackers: maintaining identity in dense crowds where individuals have uniform appearance and highly synchronized motion. Similar challenges are documented in DanceTrack \cite{dancetrack}, which shows that when appearance cues collapse, even strong trackers suffer severe identity fragmentation.
SAM2’s identity mechanism relies on (i) per-frame mask predictions and (ii) short-term memory propagation. In uniform-appearance crowds, however, these assumptions break down.
Appearance similarity invalidates visual cues, segmentation boundaries frequently overlap due to tight spatial packing, and distractors often follow nearly identical motion trajectories, rendering purely geometric or optical-flow cues insufficient for discriminating individuals. Under such conditions, SAM2’s memory-write step is prone to accepting distractor-aligned masks, causing irreversible drift.
The issue is further compounded in D-PTUAC by extremely small head-level targets, drone-induced ego motion, and repeated partial occlusions. These characteristics make D-PTUAC an ideal testbed for probing the limits of segmentation-driven identity propagation under extreme visual ambiguity.

\noindent \emph{(1) Point-Level Identity Anchors for Tiny Targets.}
Recent drone-crowd tracking research demonstrates that tiny head-level targets are more reliably represented as points or heatmap peaks rather than as full segmentation masks.
In particular, multi-frame feature-level warping has been shown to produce stable point trajectories in dense aerial crowds \cite{sol1,sol3}. Integrating a point-based identity cue into SAM2’s refine-before-write stage would provide a more persistent and occlusion-resistant identity anchor. SAM2’s masks would then be used primarily for spatial refinement rather than for identity estimation.

\noindent \emph{(2) Small-Object-Aware Proposal Generation.}
A second RGB-only direction is to introduce proposal mechanisms specialized for tiny, densely packed individuals before passing frames into SAM2.
Work on aerial-view small-object detection shows that density-guided cropping and zoom-in detection strategies significantly improve discriminability for extremely small targets \cite{sol2,sol5}. 
Likewise, the DroneCrowd benchmark \cite{sol4} demonstrates that head-level localization is far more reliable when informed by density maps or high-resolution regional proposals rather than full-frame analysis. 
Integrating similar small-object priors—such as density-driven region proposals or high-resolution head-center crops—could reduce the number of distractor masks entering SAM2’s candidate pool and provide much cleaner spatial contexts for segmentation. 
Such preprocessing is particularly relevant in D-PTUAC, where individuals often occupy only a handful of pixels, and naive segmentation fails to preserve identity under extreme crowding.

\noindent \emph{(3) Motion-Structured Temporal Consistency.}
A complementary direction involves strengthening temporal reasoning within SAM2’s refine-before-write mechanism. 
Uniform-appearance tracking benchmarks such as DanceTrack \cite{dancetrack} show that when appearance cues become uninformative, robust identity preservation depends heavily on motion regularity, spatial separation, and tracklet-level coherence. 
Extending SENTRY’s cycle-consistency logic to incorporate longer temporal horizons, local crowd-motion priors, or penalties for convergent trajectories could help suppress distractor-aligned masks in scenes where multiple individuals move with nearly identical velocity patterns. 
Enhancing motion-structured consistency in this way would preserve SAM2’s design philosophy while directly addressing the correlated-motion failure modes exposed by D-PTUAC.

\end{document}